\documentclass{article}

\PassOptionsToPackage{numbers, compress}{natbib}

\usepackage[final]{neurips_2025}

\usepackage[utf8]{inputenc} 
\usepackage[T1]{fontenc}    
\usepackage[hyperfootnotes=false]{hyperref}      
\usepackage{url}            
\usepackage{booktabs}       
\usepackage{amsfonts}       
\usepackage{mathtools}
\usepackage{nicefrac}       
\usepackage{microtype}      
\usepackage{ragged2e}
\usepackage{bbding}  
\usepackage{pifont}  
\usepackage{multirow}
\usepackage{graphicx}
\usepackage{hyperref}
\usepackage{amsfonts, amsmath, amssymb, amsthm}
\usepackage{enumitem}
\usepackage{wrapfig}
\usepackage{float}
\usepackage{array}
\usepackage{booktabs}
\usepackage{bm}
\usepackage{makecell} 
\usepackage{caption}
\usepackage[table]{xcolor}
\usepackage{subcaption}
\usepackage[misc]{ifsym}

\newcommand{\SE}{\mathrm{SE}}
\newcommand{\SO}{\mathrm{SO}}
\newcommand{\Ss}{\mathrm{S^2}}
\newcommand{\ac}{\mathbf{a}}
\newcommand{\OO}{\mathcal{O}}
\newcommand{\GL}{\mathrm{GL}}
\newcommand{\ob}{\mathbf{o}}

\newtheorem{definition}{Definition}
\newtheorem{proposition}{Proposition}

\definecolor{MossGreen}{RGB}{130, 180, 110}
\definecolor{SoftAmber}{RGB}{225, 177, 60}
\definecolor{Salmon}{RGB}{250, 128, 114}
\definecolor{DodgerBlue}{RGB}{30, 130, 255}
\definecolor{gred}{rgb}{0.859,0.267,0.216}
\definecolor{ggreen}{rgb}{0.059,0.616,0.345}
\definecolor{gpink}{RGB}{255,102,160}
\definecolor{gblue}{RGB}{51,51,255}
\definecolor{deepblue}{HTML}{27a2c3}
\definecolor{deepred}{HTML}{fe7b5b}

\hypersetup{
    colorlinks=true,
    linkcolor=DodgerBlue,
    filecolor=DodgerBlue,      
    urlcolor=DodgerBlue,
    citecolor=DodgerBlue,
    }

\newcolumntype{P}[1]{>{\centering\arraybackslash}p{#1}}

\title{3D Equivariant Visuomotor Policy Learning via Spherical Projection}

\newcommand\blfootnote[1]{%
  \begingroup
  \renewcommand\thefootnote{}\footnote{#1}%
  \addtocounter{footnote}{-1}%
  \endgroup
}
\author{%
 Boce Hu$^1$ \quad Dian Wang\textsuperscript{\Letter}$^{,2}$ \quad David Klee$^1$ \quad Heng Tian$^1$ \quad Xupeng Zhu$^1$ \quad Haojie Huang$^1$ \\
 \textbf{Robert Platt$^{\dagger,1}$} \quad \textbf{Robin Walters$^{\dagger,1}$}\\
$^1$Northeastern University \quad $^2$Stanford University \quad $^\dagger$Equal Advising\\[0.8em]
\texttt{\url{https://isp-3d.github.io/}}
}

\begin{document}

\maketitle

\begin{abstract}
Equivariant models have recently been shown to improve the data efficiency of diffusion policy by a significant margin. However, prior work that explored this direction focused primarily on point cloud inputs generated by multiple cameras fixed in the workspace. This type of point cloud input is not compatible with the now-common setting where the primary input modality is an eye-in-hand RGB camera like a GoPro. This paper closes this gap by incorporating into the diffusion policy model a process that projects features from the 2D RGB camera image onto a sphere. This enables us to reason about symmetries in $\mathrm{SO}(3)$ without explicitly reconstructing a point cloud. We perform extensive experiments in both simulation and the real world that demonstrate that our method consistently outperforms strong baselines in terms of both performance and sample efficiency. Our work, Image-to-Sphere Policy (\textbf{ISP}), is the first $\mathrm{SO}(3)$-equivariant policy learning framework for robotic manipulation that works using only monocular RGB inputs.
\end{abstract}

\section{Introduction}
\begin{wrapfigure}[13]{r}{0.51\textwidth}
    \centering
    \vspace{-1.1em}
    \includegraphics[width=\linewidth]{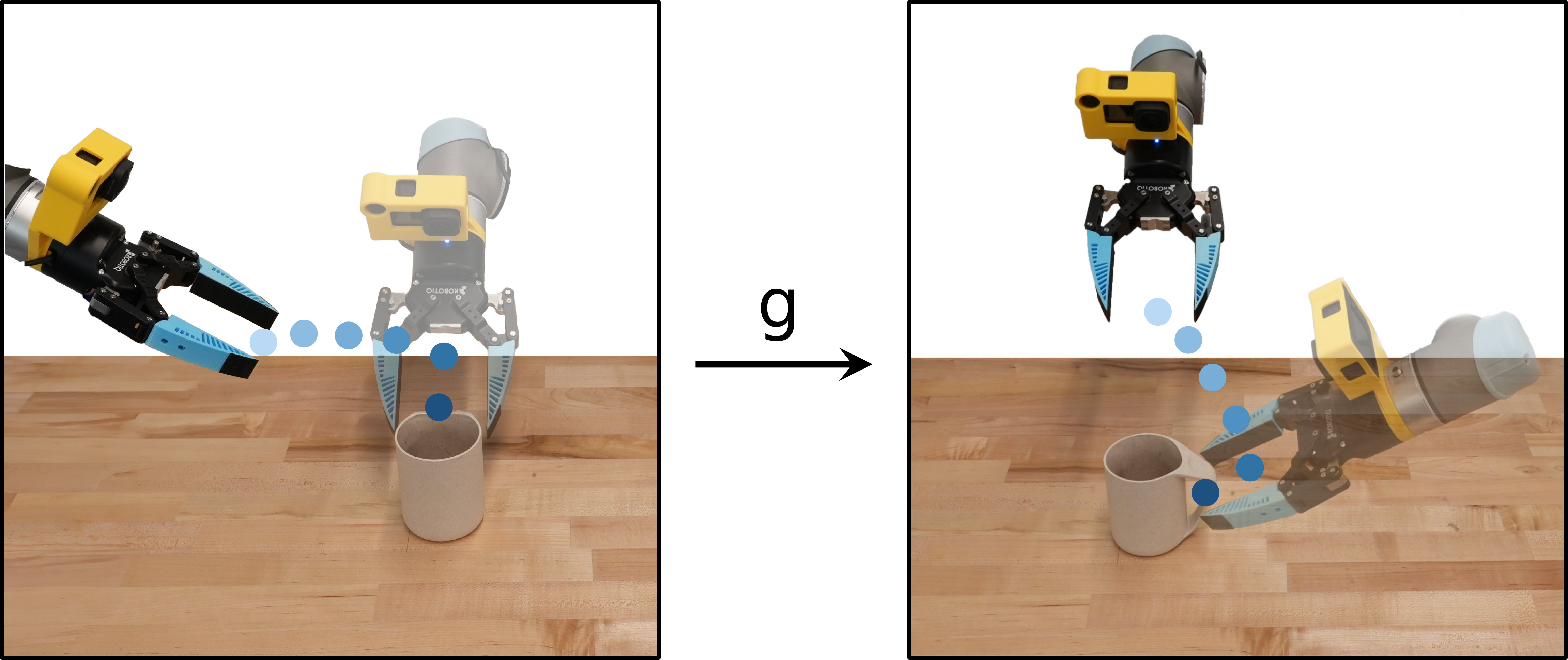}
    \vspace{-1.5em}
    \caption{We propose the first $\SO(3)$-equivariant policy learning framework based on a single eye-in-hand RGB image, where the predicted action sequence transforms equivariantly under the same group action $g \in \SO(3)$ applied to the whole scene.}
    \label{fig:teaser}
\end{wrapfigure}
The\blfootnote{\vspace{-1em}\scriptsize\textsuperscript{\Letter}Corresponding to \texttt{dianwang@stanford.edu}} eye-in-hand configuration, where the primary perception modality is a camera mounted near the wrist of the robot, is an important setting for robotic policy learning. This setup avoids the need for carefully calibrated external camera systems, is easier to integrate onto mobile robot platforms, and provides fine-grained visual details in the regions where the end-effector interacts with the environment. Moreover, it is used in a growing number of large robot datasets~\cite{kalashnikov2018scalable, kalashnikov2021mt, o2024open, khazatsky2024droid, brohan2022rt, qi2025bear}.

Despite recent advances in equivariant learning~\citep{weiler2019general,wang2022mathrm}, there remains a lack of effective network architectures for leveraging equivariant structure in this setting using only RGB input. Equivariant neural networks improve data efficiency and generalization by incorporating prior knowledge of domain symmetries directly into the model~\cite{liao2023equiformer, walters2020trajectory,mondal2020group}, and have recently been shown to enhance the performance of diffusion policy~\cite{brehmer2023edgi,yang2024equibot}. However, existing equivariant diffusion policy frameworks perform best with point cloud data captured from multiple depth cameras~\citep{tie2024seed}. When used with RGB data, current equivariant policies are unable to fully leverage the $\SO(3)$ structure present in the problem and underperform the point cloud version significantly~\cite{EquiDiff}. This naturally raises the question: can $\SO(3)$-equivariance be achieved directly from monocular RGB images to support data-efficient visuomotor learning? Such a capability should also have the potential to serve as a modular, plug-and-play component that generalizes seamlessly to richer sensing setups.

This paper addresses this challenge by introducing a novel diffusion policy framework that incorporates $\SO(3)$-equivariance into eye-in-hand visuomotor learning. Our method first projects features extracted from 2D RGB observations onto a sphere and then rotates the resulting spherical signal to compensate for camera motion. This yields a stable, $\SO(3)$-equivariant representation that is well-suited for downstream equivariant architectures. Unlike prior work that relies on segmented point clouds~\citep{tie2024seed,yang2024equibot} or calibrated multi-camera systems~\citep{EquiDiff}, our approach maintains equivariance throughout the entire policy and supports robust, sample-efficient closed-loop control directly from raw eye-in-hand inputs. To the best of our knowledge, this is the first framework to learn $\SO(3)$-equivariant visuomotor policies from monocular RGB observations in eye-in-hand settings.

\textbf{Our key contributions are summarized as follows:}
\begin{itemize}[itemsep=0.4pt, topsep=0.4pt, leftmargin=1.5em]
  \item We introduce \textbf{Image-to-Sphere Policy} (\textbf{ISP}), the first $\SO(3)$-equivariant policy learning framework that uses spherical projection from 2D RGB inputs to model 3D symmetries.
  
  \item We theoretically prove that our method achieves global $\SO(3)$-equivariance and local $\SO(2)$-invariance, facilitating policy learning.
  \item We validate our method through extensive experiments, achieving an average success rate improvement of 11.6\% over twelve simulation tasks and 42.5\% across four real-world tasks.

\end{itemize}
\vspace{-0.2cm}
\section{Related Work}
\vspace{-0.1cm}
\noindent\textbf{Eye-in-hand Policy Learning}\;
Eye-in-hand policy learning~\citep{intelligence2025pi_,jiang2025behavior,xu2025robopanoptes,wu2023tidybot, ha2024umi,pertsch2025fast} has become a flexible and scalable alternative to traditional systems that rely on multiple fixed, externally mounted cameras with precise calibration~\citep{EquiDiff,jia2022seil, tie2024seed,ze20243d, chi2023diffusion}. By mounting cameras on the robot's wrist, these methods simplify deployment, avoid explicit calibration, and ease demonstration collection~\citep{chi2024universal,liu2024fastumi,seo2024legato,jangir2022look}. However, the constantly shifting viewpoint introduces challenges like partial observability, which motivates the use of closed-loop policies that can handle local, viewpoint-dependent observations~\citep{zhang2018deep, jangir2022look, cheng2018reinforcement, ren2024diffusion,zhao2025hierarchical}. Recent work has explored transformer-based~\citep{vaswani2017attention} or diffusion-based~\citep{ho2020denoising} architectures for eye-in-hand-control~\citep{zhao2023learning, ha2024umi}, showing promising results across diverse manipulation tasks. Despite this progress, existing approaches often require large-scale demonstration data~\citep{lin2024data,liu2024rdt}, and often overlook symmetry structures inherently present in observations. Our method addresses this gap by introducing equivariant representations that encode geometric structure for eye-in-hand settings.

\noindent\textbf{Closed-loop Visuomotor Policy Learning}\;
Early approaches to closed-loop visuomotor policy learning relied on reinforcement learning and CNN-based policies to map visual inputs to single-step actions~\citep{levine2016end, zhang2018deep, kalashnikov2018scalable}. While effective in simple tasks, these methods were sample-inefficient and struggled to capture multimodal behaviors, as each action was predicted independently without considering temporal context. To address this, subsequent work introduced temporal modeling into behavior cloning frameworks, such as BCRNN~\citep{mandlekar2021matters} and BeT~\citep{shafiullah2022behavior}, to improve sequential consistency and planning horizons. Building on this direction, recent advances have adopted generative policy models~\citep{chi2023diffusion, prasad2024consistency, ze20243d, EquiDiff}, which model multi-step action sequences as a denoising process conditioned on observations. These approaches offer stronger expressiveness and improved multimodal behavior modeling. ISP extends this line of work by further integrating structural inductive biases, which enable more generalizable closed-loop control in complex manipulation settings.

\noindent\textbf{Equivariance in Robotic Manipulation}\; Equivariant and invariant representations have been shown to improve performance and sample efficiency~\citep{ heefficient, simeonov2022neural, guan2023d, teng2022lie,li2025affine, 11150764, he2022neural, li2024affine, dong2025learning}. Prior work has incorporated equivariant architectures for open-loop pick-and-place tasks~\citep{zhu2022sample, wang2022equivariant, hu2024orbitgrasp,huang2022equivariant,gao2024riemann,qi2025two, wang2022robot, huang2024leveraging}, showing strong performance with fewer demonstrations. Recently, equivariance has been extended to closed-loop diffusion policies~\citep{EquiDiff, yang2024equibot, tie2024seed}. EquiDiff~\citep{EquiDiff} employs an $\SO(2)$-equivariant architecture to enhance Diffusion Policy~\citep{chi2023diffusion}. EquiBot~\citep{yang2024equibot} adopts an $\mathrm{SIM}(3)$-equivariant structure, and ET-SEED~\citep{tie2024seed} performs trajectory-level $\SE(3)$-equivariant diffusion, both leveraging segmented point cloud inputs to model spatial symmetries, thereby improving policy generalization. 
However, these approaches typically rely on multi-camera setups with fixed viewpoints or preprocessed 3D inputs. These constraints reduce their practicality in eye-in-hand settings, where the continuously shifting viewpoint and monocular RGB input violate the assumptions of existing equivariant models. To fill this gap, ISP models symmetry in the eye-in-hand RGB setting, preserves $\SO(3)$-equivariance, and can integrate with other frameworks to enhance their effectiveness without additional preprocessing.
\section{Background}
\subsection{Representations of \texorpdfstring{$\SO(3)$}{SO(3)}}
\label{sect:representations}

A \emph{group representation} encodes symmetry by mapping elements of a group to linear transformations. In this work, we focus on the special orthogonal group $\SO(3)$ of 3D rotations. A representation of $\SO(3)$ is a homomorphism $\rho: \SO(3) \rightarrow \GL(V)$, where $V$ is a finite-dimensional vector space and $\GL(V)$ denotes the group of invertible linear transformations on $V$. We highlight three commonly used representations in robotics and geometric deep learning:
\begin{itemize}[leftmargin=1em,itemsep=0pt,topsep=0pt]
    \item \textbf{Degree-0 trivial representation} $\rho_0$: Maps every $g \in \SO(3)$ to the identity transformation on $\mathbb{R}$. This is used for rotation-invariant quantities, such as scalar sensor readings or gripper states.
    
    \item \textbf{Degree-1 standard representation} $\rho_1$: Maps $g \in \SO(3)$ to a $3 \times 3$ rotation matrix acting on $v \in \mathbb{R}^3$ via $\rho_1(g)v = g v$, capturing vector features like positions and directions.

    \item \textbf{Higher degree irreducible representations} $\rho_\ell$: For $\ell \in \mathbb{N}$, the representation $\rho_\ell \colon \SO(3) \to \GL(\mathbb{R}^{2\ell+1})$ is given by the Wigner $D$-matrix of degree $\ell$. It is used to describe higher degree features, such as relative poses, and is often used for latent features in equivariant neural networks.
\end{itemize}

\subsection{Spherical Harmonics and Fourier Coefficients}
Spherical harmonics $Y_\ell^m: \mathbb{S}^2 \rightarrow \mathbb{R}$ form an orthonormal basis for square-integrable functions on the 2-sphere and realize the irreducible representations of $\SO(3)$. Any spherical function $\Phi: \mathbb{S}^2 \rightarrow \mathbb{R}^d$ can thus be expanded as:
\begin{equation}
\label{eqn:ifft}
\Phi(\theta, \phi) = \sum_{\ell=0}^{\infty} \sum_{m=-\ell}^{\ell} c_{\ell}^m \, Y_{\ell}^m(\theta, \phi),
\end{equation}
where $c_\ell^m$ are the corresponding Fourier coefficients.  The mapping $\Phi \mapsto \{c_\ell^m\}$ is known as the Spherical Fourier Transform. Under a rotation $R \in \SO(3)$, each coefficient vector $c_\ell \in \mathbb{R}^{2\ell+1}$ transforms linearly via the representation $\rho_\ell$:
\begin{equation}
\label{eq:wigner}
c_\ell' = \rho_\ell(R) \cdot c_\ell.
\end{equation}
This enables efficient rotation-equivariant operations on spherical signals in the spectral domain.

\subsection{Diffusion Policy}
\label{background: dp}
Diffusion-based policy learning~\citep{janner2022planning,reuss2023goal} is a class of imitation learning methods that model distributions over action trajectories using denoising diffusion probabilistic models (DDPMs)~\citep{ho2020denoising}. These methods iteratively denoise sequences of noisy actions, conditioned on observations, to recover expert-like behavior. Formally, given an observation $\OO$ and diffusion timestep $k$, the policy predicts a noise estimate $\epsilon^k$ from a corrupted action sequence $\ac^k = \ac^0 + \epsilon^k$ using a denoiser network $\Gamma$. The model is trained to minimize the denoising objective $
\mathcal{L}_\text{diff} = \mathbb{E}_{\ac^0, k,\epsilon^k}\big[\|\Gamma(\mathcal{O}, \ac^k, k) - \epsilon^k\|^2\big].$ At test time, the policy generates actions by iteratively denoising a randomly initialized sequence from Gaussian noise. Recent extensions~\citep{EquiDiff} incorporate symmetry priors by designing the denoiser to be equivariant with respect to a transformation group $G$. Specifically, for compact groups such as $\SO(3)$, the denoiser $\Gamma$ is required to satisfy the equivariance constraint:
\begin{equation}
    \Gamma(g \cdot \mathcal{O},\, g \cdot \ac^k,\, k) = g \cdot \Gamma(\mathcal{O}, \ac^k, k) \quad \forall g \in G.
\end{equation}
This formulation ensures that the denoising process respects the symmetry of the environment.

\subsection{Problem formulation}
We study closed-loop robotic visuomotor policy learning through behavior cloning, where a policy is trained to imitate expert demonstrations. Given an observation sequence $\OO=\{o_{t-k+1},...,o_{t}\}$ at timestep $t$, the learned policy predicts an action chunk $A = \{\ac_{t+1},...,\ac_{t+n}\}$, where $k$ and $n$ are the observation length and prediction horizon, respectively. Each observation $o = (I, P)$ consists of an RGB image $I$ from the wrist camera, proprioceptive input $P$ describing the end-effector pose. Prior work has shown that higher performance is achieved when using absolute action representations, i.e., actions expressed in the world frame~\cite{chi2023diffusion, EquiDiff}. Following this, we represent each predicted action $a_t \in \mathbb{R}^{10}$ as the absolute end-effector pose including a position in $\mathbb{R}^{3}$, an orientation represented as a 6D rotation vector in $\mathbb{R}^{6}$ (see \cite{6d_repr}), and a gripper open state in $\mathbb{R}^{1}$. As noted by~\citep{EquiDiff}, the absolute action parametrization has a symmetry: 3D transformations of the world frame result in the same 3D transformations to the action. We formalize the equivariance properties of ISP in Section \ref{sec:so3-equiv-obs-encoder}.

\begin{figure*}[!t]
\centering
\includegraphics[width=\linewidth]{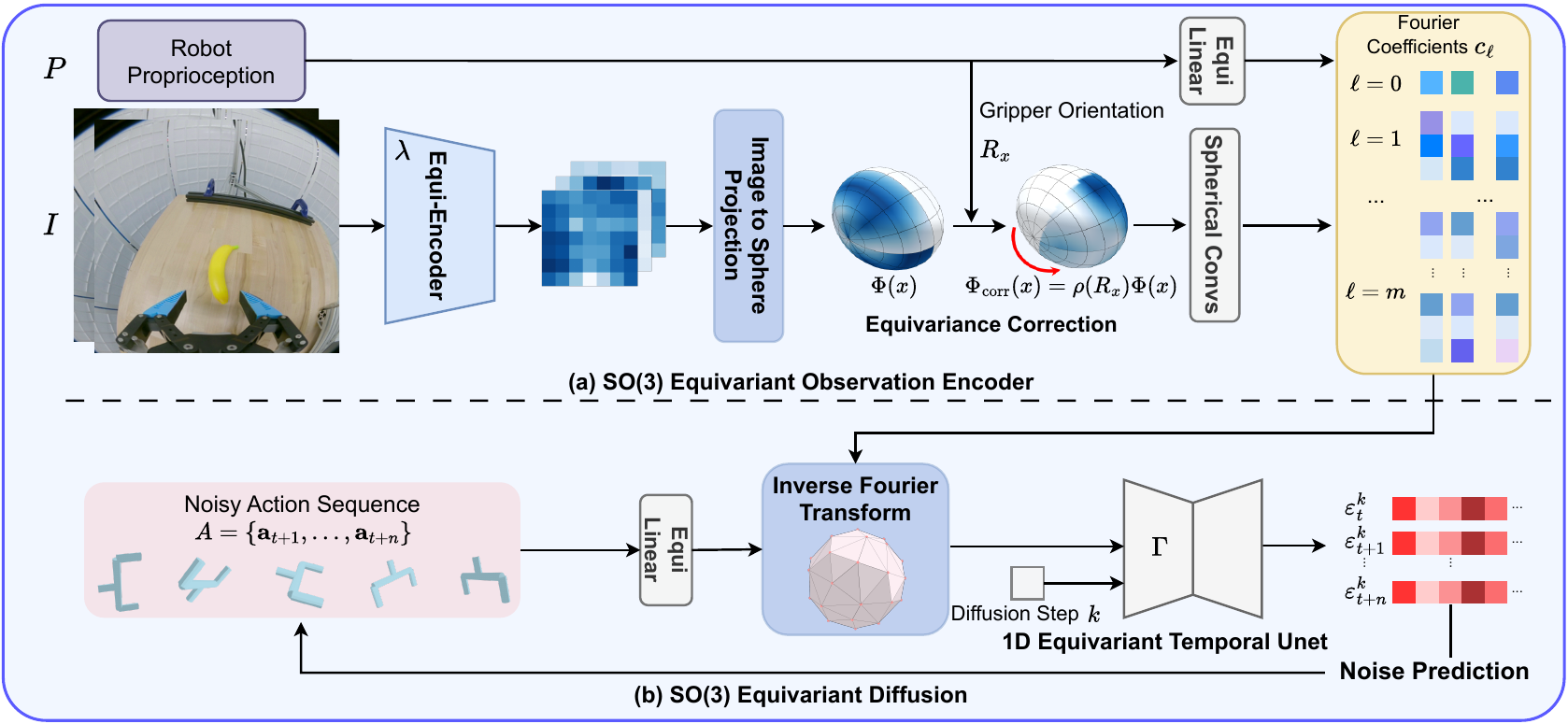}
\vspace{-0.3cm}
\caption{\textbf{Overview of Image-to-Sphere Policy (ISP)} (a) An $\SO(3)$-equivariant observation encoder extracts features from the RGB input, projects them onto the sphere, and applies an equivariance correction using the gripper orientation $R_x$ to account for the camera's dynamic viewpoint (red arrow). The corrected spherical signal $\Phi_{\text{corr}}(x)$ is then processed by spherical convolution layers to extract $\SO(3)$ signals. Proprioceptive inputs are embedded via equivariant linear layers. Both image and proprioceptive features are represented as a set of Fourier coefficients $c_{\ell}$ on $\SO(3)$ and fused (yellow block). (b) The encoded spherical signals are transformed back to the spatial domain via inverse Fourier transform, sampling finite group elements as the conditioning vector for $\SO(3)$-equivariant denoising. The noisy action sequence is processed in the same way, through equivariant linear layers and projected onto the same group elements.}
\label{fig:pipeline}
\vspace{-0.3cm}
\end{figure*}

\section{Method}
Figure~\ref{fig:pipeline} illustrates an overview of our proposed method, which consists of two key components, an $\SO(3)$-equivariant observation encoder (Figure~\ref{fig:pipeline}\kern0.07em a) and an $\SO(3)$-equivariant diffusion module (Figure~\ref{fig:pipeline}\kern0.07em b). The observation encoder uses spherical projection to map image-extracted features onto a hemisphere and applies spherical convolutions to ensure $\SO(3)$-equivariance, producing the conditioning vector for the diffusion process. The diffusion module is designed as an $\SO(3)$-equivariant function of the conditioning vectors and noisy inputs. As a result, the entire policy is end-to-end $\SO(3)$-equivariant.
In the following subsections, we first describe our observation encoder, which extracts $\SO(3)$-equivariant features from 2D images, and then our equivariant diffusion module.

\subsection{\texorpdfstring{$\SO(3)$ Equivariant Observation Encoder}{SO(3) Equivariant Observation Encoder}}\label{sec:so3-equiv-obs-encoder}

This section describes how we construct an $\SO(3)$-equivariant observation encoder that maps 2D images and robot proprioception into a 3D feature representation.
The observation $x\in X$ consists of two parts, an eye-in-hand RGB image $I$, that captures visual information, and proprioceptive data, $P\in\mathbb{R}^{7}$, including the end-effector's 6D pose (position and orientation) and gripper state. Both these signals need to be represented in a way that encodes equivariance. Representing $P$ is relatively easy. Following~\citep{EquiDiff, yang2024equibot, ryu2024diffusion}, we embed end-effector pose in $\SO(3)$ using the standard representation and gripper state using the trivial representation (Section~\ref{sect:representations}). In contrast, encoding the 2D image input $I$ into $\SO(3)$-equivariant features is harder because changes in the pose of the wrist-mounted camera induce out-of-plane viewpoint variations that are hard to model. We address this by projecting a standard 2D encoding of the image onto the sphere, as described below and first proposed in the context of object pose estimation~\citep{i2s,induce2s}. This enables us to reason about $\SO(3)$ action using its irreducible representations encoded as Wigner D-matrices (Section~\ref{sect:representations}).
\begin{figure*}[!t]
\centering
\includegraphics[width=0.9\linewidth]{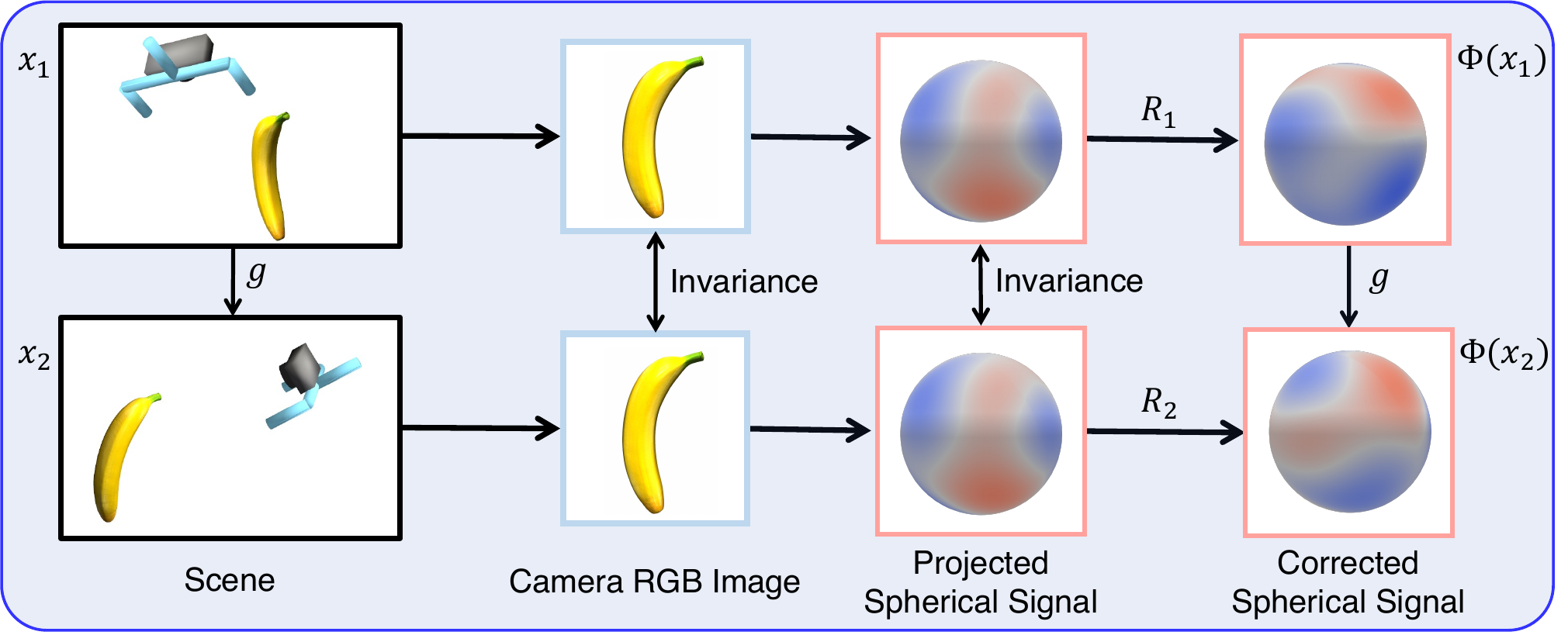}
\vspace{-0.1cm}
\caption{\textbf{Illustration of Equivariance Correction.} The left side shows two identical scenes under different global transformations. Since the wrist-mounted camera captures images in its local frame, the resulting images, and thus the projected spherical signals, remain identical across both scenes. By applying the gripper orientation $R$ as an equivariance correction, we align these spherical signals to a common world frame, ensuring their equivariant transformation under global scene rotations.}
\label{fig:equi_repr}
\vspace{-0.5cm}
\end{figure*}

\noindent\textbf{Image Encoder}\; Our image encoder is detailed in Figure~\ref{fig:pipeline}\kern0.07em a. First, we encode the input image $I$ from the observation $x$ using a standard $\SO(2)$-equivariant image encoder $\lambda$. Next, the resulting feature map $\lambda(I)$ is mapped onto the sphere using a learnable orthographic projection (see Appendix~\ref{app:ortho-proj} for details). This converts a ``flat'' image into a spherical signal $\Phi(x)\!:\mathbb{S}^2\!\to\!\mathbb{R}^d$ that is easier to manipulate in $\SO(3)$. We represent this spherical signal in the spectral domain as truncated Fourier coefficients calculated using the spherical Fourier transform (Equation~\ref{eqn:ifft}). 

\noindent\textbf{Equivariance Correction}\; At this point, the image encoding has been projected onto the sphere and represented using spherical harmonics. However, there is a problem. Since global 3D transformations of the world transform the camera and objects equally, the observed image and the projected signal $\Phi(x)$ would be invariant. This introduces a mismatch in that $\Phi(x)$ stays constant while the world and actions rotate, thereby breaking global $\SO(3)$-equivariance.
We accommodate this by rotating the spherical signal by an amount corresponding to the $\SO(3)$ orientation of the gripper. We call this the \emph{equivariance correction} factor, and it is illustrated in Figure~\ref{fig:equi_repr}. On the left of Figure~\ref{fig:equi_repr}, we see two scenes that are the same except for an $\SO(3)$ rotation of $g$. The eye-in-hand camera image (of the banana) is the same in both situations, even though the scene is rotated. This results in the identical projected spherical signal. It is only by applying the equivariance correction factor to the two respective signals ($R_1$ and $R_2$) that we recover the camera pose in the spherical signal. This ensures that the spherical signals produced in different camera poses are represented in a consistent global frame. We analyze this approach below.

\begin{definition}[Equivariance Correction]
\label{def:ec}
Let $G$ be a group acting on the input space $X$ and output space $Y$.
For a function $f\!:\!X\!\to\! Y$,
an \emph{Equivariance‑Correction map} is any $\mathcal{C}\!:\!X\!\to\!G$ satisfying $\mathcal{C}(g\!\cdot\!x)\,f(g\!\cdot\!x)=g\,\mathcal{C}(x)\,f(x)$
for all $g\!\in\! G,$ and $x\!\in\! X.$ The corrected function $f_{\text{corr}}(x) = \mathcal{C}(x)\,f(x)$ is therefore $G$‑equivariant.
\end{definition}

Notice that Definition~\ref{def:ec} implies $f(x)$ and $f(gx)$ are in the same orbit. Equivariance Correction is similar to a \emph{canonicalization map} $c:X\to G$, where $f_\text{cano}=c(x)f(c(x)^{-1}x)$ transforms the input to a canonical frame, then transforms the output back to the original frame. When $f(x) = f(gx)$, Equivariance Correction is a special case of canonicalization where $f(c(x)^{-1}x)=f(x)$ is invariant, so it only transforms the output to restore equivariance without altering the input.

We now show that Definition~\ref{def:ec} is satisfied when the correction map is chosen to be the end‑effector rotation. Let $x \in X$ denote the robot observation at a given timestep, which includes an eye-in-hand RGB image $I$ and the corresponding camera (end-effector) pose $R_x \in \SO(3)$ in the world frame. Let $\Phi(x)$ denote the spherical signal derived from the image $I$, expressed in the camera frame, and let $\rho$ be a representation of $\SO(3)$ acting on $\Phi(x)$.
\begin{proposition}[Equivariance Correction via End-Effector Pose]
\label{prop:end-eff}
The map $\mathcal{C}\colon (I, R_x) \mapsto R_x$, which assigns each camera image to its corresponding camera pose $R_x\in \SO(3)$ is an equivariance correction. The corrected signal $\Phi_{\text{corr}}(x)=\rho(\mathcal{C}(x))\Phi(x)=\rho(R_x)\Phi(x)$ is in a world‑aligned frame. Thus, the mapping $\Phi_{\text{corr}}$ is $\SO(3)$‑equivariant: for any global rotation $g\in\SO(3)$, we have $\Phi_{\text{corr}}(g\!\cdot\!x)=\rho(g)\,\Phi_{\text{corr}}(x).$
\end{proposition}

\begin{proof}
Let $g\in\SO(3)$ be a global rotation applied simultaneously to the scene and the camera.
Since the image is recorded in the camera frame, the spherical signal is unaffected, i.e.\ $\Phi(g\!\cdot\!x)=\Phi(x)$, while the camera pose updates as $R_x\mapsto R_{gx} = gR_x$.
For the corrected signal, we therefore obtain
\begin{equation}
    \Phi_{\text{corr}}(gx)=\rho(R_{gx})\Phi(gx)
               =\rho(g)\rho(R_x)\Phi(x)
               =\rho(g)\Phi_{\text{corr}}(x),
\end{equation}
where the second equality follows from the homomorphism property $\rho(gR_x)=\rho(g)\rho(R_x)$ of the representation~$\rho$.
Hence the map $\Phi_{\text{corr}}$ is $\SO(3)$‑equivariant.
\end{proof}
A concrete realization of $\rho$ with the spherical–harmonic coefficients and Wigner $D$-matrices is given in Appendix~\ref{app:spectral-corr}, where the proposition reduces to the
rotation of coefficient vectors in Eq.~\ref{eq:wigner}.

\noindent\textbf{Camera-rotation invariance}\; Our model also enforces an additional symmetry, rotations of the camera around its optical axis while the object remains stationary. These rotations form an $\SO(2)$ subgroup. Such rotations transform both the image and the camera pose, but their effects cancel out in the corrected world-frame signal. We now formalize the invariance of the corrected world-frame signal under such transformations.
\begin{proposition}[Invariance to $\SO(2)$ Rotation of the Eye-in-hand Camera]
\label{prop:roll}
    Let $g\!\in\!\SO(2)$ be a rotation about the camera’s optical axis. Then, under the transformation $(I, R_x) \mapsto (g \cdot I, R_x g^{-1})$, the corrected signal defined in Proposition~\ref{prop:end-eff} remains invariant: $\Phi_{\text{corr}}(g\cdot x)=\Phi_{\text{corr}}(x)$.
\end{proposition}    

\begin{proof}
Assume the image encoder $\lambda$ is $\SO(2)$-equivariant, i.e., $\lambda(g \cdot I) = g \cdot \lambda(I)$ for all $g \in \SO(2)$. Because spherical projection and spherical Fourier transform preserve equivariance, the spherical signal satisfies $\Phi(g \cdot x) = g\cdot \Phi(x)$. Meanwhile, the camera pose transforms as $R_x \mapsto R_x g^{-1}$, since applying an $\SO(2)$ rotation $g$ in the camera frame corresponds to right-multiplying its world-frame orientation $R_x$ by $g^{-1}$ (i.e., rotating the camera relative to itself). The corrected signal is: 
\begin{equation}
    \Phi_{\text{corr}}(g\cdot x)=\rho(R_xg^{-1})\,\Phi(g\cdot x)
               =\rho(R_xg^{-1})\,\rho(g)\,\Phi(x)
               =\rho(R_x)\,\Phi(x)
               =\Phi_{\text{corr}}(x).
\end{equation}
Thus, the corrected signal is invariant under any $\SO(2)$ rotations of the eye-in-hand camera.
\end{proof}
By combining Propositions~\ref{prop:end-eff} and~\ref{prop:roll}, we obtain a two-level symmetry in the encoder: the features are globally $\SO(3)$-equivariant and locally $\SO(2)$-invariant to rotations of the camera around its optical axis. These properties are inherently preserved without requiring additional constraints. As shown in Section~\ref{sec:exp}, encoding these properties into the network leads to empirically improved performance.

\subsection{\texorpdfstring{$\SO(3)$ Equivariant Diffusion}{SO(3) Equivariant Diffusion}}
\label{sec:equiv-diff}
As described in Section~\ref{background: dp}, we enforce end-to-end $\SO(3)$-equivariance by requiring the denoising network $\Gamma$ to satisfy:
$\Gamma(g \cdot \OO, g \cdot \ac^k, k) = g \cdot \Gamma(\OO, \ac^k, k)$ for all $g \in \SO(3).$
To achieve this, we extend the 2D denoising model from \emph{EquiDiff}~\citep{EquiDiff} to 3D. EquiDiff applies a shared 1D temporal U-Net~\citep{ronneberger2015u} independently to each group element in $C_n \subset \SO(2)$. This element-wise weight sharing guarantees that the same parameters act on every group element, resulting in a noise embedding in the regular representation. To generalize to 3D, we approximate the continuous symmetry group $\SO(3)$ with a finite subgroup and perform sampling accordingly. Denote $H \subset \SO(3)$ the subgroup that the diffusion process is equivariant to (e.g., the icosahedral group $I_{60}$). Denote $S\subset\SO(3)$ a set that is closed under $H$, i.e., $HS=S$. Intuitively, $S$ could be viewed as copies of the rotations in $H$, each with different offset angles to capture a denser discrete signal. Given a signal $\Psi:\SO(3)\to \mathbb{R}^d$, we first sample $\Psi(S)=\{\Psi(s_i): s_i\in S\}$ and then evaluate the U-Net pointwise on each sample $\Gamma(\Psi(S))=\{\Gamma(\Psi(s_i)): s_i\in S\}$, where both the input and output can be treated as copies of the regular representations of $H$. Since $g\in H$ permutes the order of $\Psi(S)$ and $\Gamma(\Psi(S))$ identically, the entire process is $H$-equivariant. Because the spherical convolution layers output a signal on $\SO(3)$, we can flexibly choose any finite group $H$ and sampling set $S$ for discretization. In our implementation, we use both $C_8\subset \SO(2)$ and $I_{60}\subset \SO(3)$ as choices of $H$. We refer readers to Appendix~\ref{appendix:implement_details} for further details.

\subsection{End-to-End Symmetry Analysis}
In this section, we analyze the equivariant properties of our method. First, due to the $\SO(3)$-equivariant encoder (Proposition~\ref{prop:end-eff}) and the $\SO(3)$-equivariant diffusion model (Section~\ref{sec:equiv-diff}), our policy has end-to-end symmetry to global scene $\SO(3)$ rotations. This significantly improves its sample efficiency and generalizability to world coordinate frame changes. 
\begin{wrapfigure}[11]{r}{0.37\textwidth}
    \centering
    \vspace{-0.1em}
    \includegraphics[width=\linewidth]{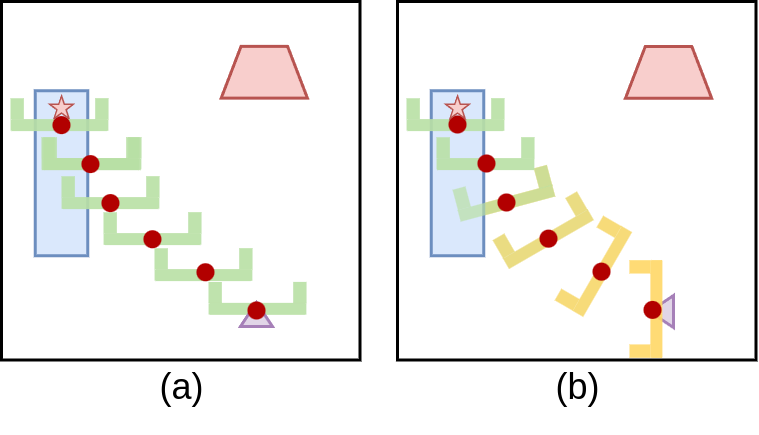}
    \vspace{-2em}
    \caption{Illustration of translation invariance and rotation equivariance-to-invariance transition. }
    \label{fig:camera_rot}
\end{wrapfigure}
The benefit of $\SO(2)$ camera-rotation invariance (Proposition~\ref{prop:roll}) is subtle. Under a rotation of the gripper with respect to the workspace, no \text{a priori} constraint can be placed on how the action trajectory should transform. However, our diffusion model receives a representation from the observation encoder that is equivariant to this rotation because it is constructed from invariant features (from the spherical signals) and equivariant features (from the end-effector rotation), thus providing a structured geometric bias. Figure~\ref{fig:camera_rot} illustrates the benefit of this design. In both states (a) and (b), the gripper (triangle) aims to reach the same goal pose (star), but in (b) it is rotated by 90$^\circ$ around its optical axis. Translationally, the action in (b) should remain invariant (red dots), while rotationally, it should gradually transition from equivariant (yellow) to invariant (green) behavior. The equivariant component in the representation ensures that the model can correctly handle the initial 90$^\circ$ rotation through its symmetry, while the invariant component provides stability and goal alignment. Together, this representation offers a geometric inductive bias for learning such trajectories, whereas non-equivariant models must infer these patterns purely from data. The advantage is empirically validated in Section~\ref{sec:exp}.

\section{Experiments}
\label{sec:exp}
\subsection{Simulation}

\begin{figure*}[!t]
\centering
\includegraphics[width=0.99\linewidth]{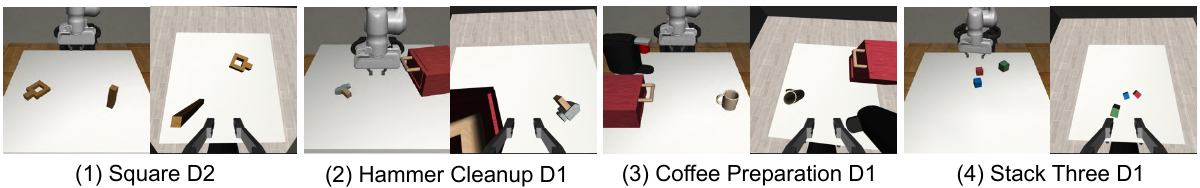}

\caption{A subset of experimental environments from MimicGen.  Left: external view of the task. Right: eye-in-hand observation used in the experiments. The full set of tasks is shown in Appendix~\ref{sec:sim_settings}.}
\label{fig:sim_tasks}
\vspace{-0.3cm}
\end{figure*}

\noindent\textbf{Experiment Setting}\; We evaluate ISP on twelve robotic manipulation tasks from the MimicGen benchmark~\citep{mandlekar2023mimicgen}, which is widely used in previous work on closed-loop policy learning~\citep{funk2024actionflow,EquiDiff}. A representative subset of these simulation tasks is shown in Figure~\ref{fig:sim_tasks} (see Appendix~\ref{sec:sim_settings} for a full description of all twelve MimicGen tasks). Policies are trained and evaluated exclusively using eye-in-hand RGB observations (right image in each subfigure of Figure~\ref{fig:sim_tasks}). To capture sufficient scene context, we enlarge the camera's field of view (FOV) to approximate a typical fisheye camera setup and re-generate the enlarged FOV observations using the original Mimicgen demonstrations for our method and baselines.
For each task, we train three independent models with different random seeds (0, 1, and 2) for each of the 100- and 200-demonstration settings. The models are evaluated 60 times throughout training using 50 fixed rollouts per evaluation. We report the average of the best success rates from the three runs. Task and training details are provided in Appendix~\ref{sec:sim_settings} and Appendix~\ref{sec:train_details}.

\noindent\textbf{Baselines}\; Our experiments aim to validate the benefits of explicitly modeling equivariance in eye-in-hand visuomotor policies. We evaluate two versions of ISP with different symmetry levels, an $\bm{\SO(3)}$\textbf{-equivariant} version and an $\bm{\SO(2)}$\textbf{-equivariant} variant, which is symmetric only about rotations in the plane of the table. Although the $\SO(3)$ version has more symmetry, the $\SO(2)$ version is more lightweight, which may be preferable in some settings. We compare against three strong baselines: \textbf{(1)} \textbf{Diffusion Policy}~\citep{chi2023diffusion}: A diffusion-based policy without any equivariance, serving as the primary reference. 
\textbf{(2)} \textbf{EquiDiff (modified)}~\citep{EquiDiff}: Designed for fixed-camera settings, it achieves $\SO(2)$ equivariance via an equivariant image encoder and an equivariant temporal U-Net. For eye-in-hand control, we replace its image encoder with a standard ResNet~\citep{He_2016_CVPR}, so only proprioception and denoising remain equivariant. 
\textbf{(3)} \textbf{ACT}~\citep{zhao2023learning}: A transformer-based behavior cloning method. To ensure a fair comparison, all experiments in the following sections, including ablations and method variants, consistently \textbf{apply} $\SO(2)$ data augmentation during training by rotating the end-effector pose in both proprioception and actions, equivalent to jointly rotating the gripper and scene.

\begin{table*}[t]
\centering
\scriptsize
\renewcommand{\arraystretch}{1.2}
\caption{Success rates (\%) on MimicGen tasks with 100 and 200 demonstrations, averaged over 3 seeds. We report both overall mean and per-task performance. The best result is highlighted in \textbf{bold}, and the second best is \underline{underlined}. Full results with standard deviations are in Appendix~\ref{appendix:seed_variance}.}
\setlength{\tabcolsep}{2pt}
\rowcolors{4}{white!100}{gray!5}
\begin{tabular}{@{}l P{1.2cm}P{1.2cm} P{0.6cm}P{0.6cm} P{0.8cm}P{0.6cm}  P{0.8cm}P{0.6cm} P{0.8cm}P{0.6cm} P{0.8cm}P{0.7cm} P{0.8cm}P{0.5cm}@{}}

\toprule
&
\multicolumn{2}{c}{\textbf{Mean}} & 
\multicolumn{2}{c}{Stack D1}&
\multicolumn{2}{c}{Stack Three D1} &
\multicolumn{2}{c}{Square D2} &
\multicolumn{2}{c}{Threading D0} &
\multicolumn{2}{c}{Three Pc. D0}&
\multicolumn{2}{c}{Hammer Cl. D1}\\

\cmidrule(lr){2-3} \cmidrule(lr){4-5} \cmidrule(lr){6-7} \cmidrule(lr){8-9} \cmidrule(lr){10-11} \cmidrule(lr){12-13} \cmidrule(lr){14-15}
\textbf{Method} & 100 & 200 & 100 & 200 & 100 & 200 & 100 & 200 & 100 & 200 & 100 & 200 & 100 & 200 \\
\midrule
ISP-$\SO(3)$  & 65.2 ({\color{blue}+11.6}) & 75.0 ({\color{blue}+10.5})& \textbf{99} & \textbf{100} & \underline{70} & \textbf{88} & \textbf{35} & \textbf{51} & \textbf{90} & \textbf{92} & 71 & \underline{79} & \underline{66} & \underline{73}  \\
ISP-$\SO(2)$ & 65.0 ({\color{blue}+11.4}) & 73.1 ({\color{blue}+8.6}) & \underline{98} & \textbf{100} & \textbf{75} & \textbf{88} & \underline{32} & \textbf{51} & 85 & 87 & \textbf{75} & \textbf{80} & \textbf{71} & \underline{73}  \\
DiffPo      & 53.6 & 64.1 & 91 & 96  & 43 & 77 & 12 & 25 & 77 & 87 & 73 & 73 & 59 & 63  \\ 
EquiDiff    & 53.0 & 64.5 & 96 & 99  & 61 & 80 & 9  & 19 & \underline{89} & \textbf{92} & \underline{74} & \underline{79} & 59 & \textbf{74}  \\
ACT         & 23.0 & 40.9 & 45 & 77  & 12 & 37 & 3  & 10 & 36 & 53 & 28 & 50 & 35 & 63  \\
\midrule
&&&

\multicolumn{2}{c}{Mug Cl. D1} &
\multicolumn{2}{c}{Coffee D2}&
\multicolumn{2}{c}{Kitchen D1}&
\multicolumn{2}{c}{Pick Place D0} &
\multicolumn{2}{c}{Coffee Prep. D1} &
\multicolumn{2}{c}{Nut Asse. D0} \\
 \cmidrule(lr){4-5} \cmidrule(lr){6-7} \cmidrule(lr){8-9} \cmidrule(lr){10-11} \cmidrule(lr){12-13} \cmidrule(lr){14-15}
\textbf{Method} &&& 100 & 200 & 100 & 200 & 100 & 200 & 100 & 200 & 100 & 200 & 100 & 200 \\
\midrule
ISP-$\SO(3)$ &&&  \underline{54} & 59 &\textbf{64} & \textbf{69} & \textbf{75} & \textbf{79} & \underline{42} & \textbf{66} & \underline{41} & \textbf{61} & \textbf{75} & \underline{82} \\ 
ISP-$\SO(2)$ &&&  \textbf{56} & \underline{61} & \underline{59} & \underline{63} & \underline{65} & \underline{72} & \textbf{46} & \underline{61} & \textbf{47} & \underline{56} & \underline{74} & \textbf{84} \\
DiffPo      &&& 49 & \underline{61} & 53 & 55 & 61 & 71 & 36 & 48 & 37 & 52 & 51 & 62 \\
EquiDiff    &&& 51 & \textbf{62} & 47 & 61 & 55 & 67 & 28 & 46 & 27 & 39 & 40 & 56 \\
ACT         &&& 25 & 37 & 21 & 35 &21 & 51 & 9  & 14 & 8  & 16 & 37 & 49 \\
\bottomrule
\end{tabular}
\label{tab:main_table}
\vspace{-12pt}
\end{table*}

\noindent\textbf{Results}\; Table~\ref{tab:main_table} reports the maximum success rates across all methods and configurations. In terms of performance, ISP-$\SO(3)$ achieves the best results in 21 out of 24 task settings, consistently outperforming the baselines. The remaining three settings show only marginal differences (within 1-2\%), all within the standard error margins. Similarly, ISP-$\SO(2)$ outperforms baselines in 20 settings, which further validates the effectiveness of our design. With only 100 demonstrations, our model exceeds the best-performing baseline by an average of 11.6\%. With 200 demonstrations, the advantage remains similar at 10.5\%. Importantly, our model trained with 100 demonstrations surpasses all baselines trained with 200 demonstrations and additional data augmentation, clearly demonstrating superior data efficiency. These results collectively highlight that the explicit modeling of equivariance is the key factor driving both the improved performance and enhanced sample efficiency of our method. Appendix~\ref{appendix:seed_variance} provides the full experimental results with standard deviations across three random seeds.

\begin{wraptable}{r}{0.56\textwidth}
\vspace{-13pt}
\centering
\caption{Ablation study results. A red cross indicates that the corresponding module is removed in that variant.}
\small
\renewcommand{\arraystretch}{1.2}
\setlength{\tabcolsep}{2pt}
\vspace{-6pt}
\begin{tabular}{ccc|cccc|c}
\toprule

Sphere & EquiEnc & EquiU& Sta. & Cof. & Nut. & Squ. & Mean\\

\midrule
\textcolor{gred}{\XSolidBrush} & \textcolor{ggreen}{\Checkmark} & \textcolor{ggreen}{\Checkmark} & 63.3 & 61.3 & 59.0 & 23.3 & 51.8 ({\color{red}-9.2}) \\
\textcolor{ggreen}{\Checkmark} & \textcolor{gred}{\XSolidBrush} & \textcolor{ggreen}{\Checkmark} & 66.0 & 57.3 & 61.3 & 32.0 & 54.2 ({\color{red}-6.8})\\
\textcolor{ggreen}{\Checkmark} & \textcolor{ggreen}{\Checkmark} & \textcolor{gred}{\XSolidBrush} & 68.7 & 58.7 & 58.0 & 32.0 & 54.3 ({\color{red}-6.7})\\
\textcolor{ggreen}{\Checkmark} & \textcolor{ggreen}{\Checkmark} & \textcolor{ggreen}{\Checkmark} & \textbf{70.0} & \textbf{64.0} & \textbf{75.3} & \textbf{34.7} & \textbf{61.0}\\
\bottomrule
\end{tabular}
\label{tab:ablation}
\vspace{-9pt}
\end{wraptable}
\noindent\textbf{Ablation Study}\; To assess the contribution of each component of our method, we conduct an ablation study on four representative tasks with 100 demonstrations: Stack Three D1, Square D2, Coffee D2, and Nut assembly D0. We evaluate the following variants of ISP-$\SO(3)$, each corresponding to a core module in our design: (1) \textbf{Sphere}: With or without the spherical projection and spherical convolutions for extracting $\SO(3)$-equivariant features from images. (2) \textbf{EquiEnc}: With or without the proposed equivariant image encoder that captures $\SO(2)$-invariant features (Proposition~\ref{prop:roll}). (3) \textbf{EquiU}: With or without an equivariant temporal denoising U-Net in the diffusion module. The results are summarized in Table~\ref{tab:ablation}. Removing the spherical projection leads to the largest performance drop of 9.2\%, highlighting its critical role in capturing symmetries, despite the use of data augmentation. Disabling the equivariant image encoder and the equivariant U-Net results in drops of 6.8\% and 6.7\%, respectively. These results demonstrate that all three components: spherical lifting, invariant encoding, and equivariant denoising, are essential for the overall effectiveness of our method. In addition, we further investigate the role of Equivariance Correction (Proposition~\ref{prop:end-eff}) by comparing delta and absolute control strategies in Appendix~\ref{app:delta}.

\begin{table*}[!t]
\scriptsize
\setlength\tabcolsep{2pt}
\centering
\caption{Success rates (\%) on MimicGen tasks with 100 demonstrations, comparing pretrained and scratch initialization of the equivariant image encoder. Results are averaged over three seeds. Values in parentheses indicate the performance difference between the two settings.}
\vspace{-0.2cm}
\begin{tabular}{l @{\hspace{8pt}} c @{\hspace{12pt}} cccccc}
\toprule
Method & Mean & Stack D1 & Stack Three D1 & Square D2 & Threading D0 & Three Pc. D0 & Hammer Cl. D1 \\
\midrule
ISP-$\SO(2)$ (Pretraining) & 72.1(\color{blue}+7.1) & 98.0 ({\color{blue}=}) & 81.3 ({\color{blue}+6.6})& 56.0 ({\color{blue}+24.0})& 91.3 ({\color{blue}+6.6})& 76.7 ({\color{blue}+2.0}) & 72.7 ({\color{blue}+2.0}) \\
ISP-$\SO(2)$ (Scratch) & 65.0 &98.0 & 74.7 & 32.0 & 84.7 & 74.7 & 70.7 \\
\midrule
 &  & Mug Cl. D1 & Coffee D2 & Kitchen D1 & Pick Place D0  & Coffee Pre. D1 & Nut Assembly D0 \\
\midrule
ISP-$\SO(2)$ (Pretraining) &  & 54.0 ({\color{red}-2.0}) & 66.7 ({\color{blue}+8.0})& 64.0 ({\color{red}-0.7})& 56.3 ({\color{blue}+10.6})& 63.3 ({\color{blue}+16.6}) & 85.0 ({\color{blue}+11.3})\\
ISP-$\SO(2)$ (Scratch) & & 56.0 & 58.7 & 64.7 & 45.7 & 46.7 &73.7\\
\bottomrule
\end{tabular}
\vspace{-0.1cm}
\label{tab:sim_pretrain_all}
\end{table*}

\noindent\textbf{The Benefits of Pretraining}\;
While our method already benefits from explicit equivariance, we further explore whether incorporating a pretrained image encoder can provide additional performance gains. Intuitively, pretraining can introduce stronger geometric priors and yield higher-quality visual features, especially beneficial in data-limited regimes. To evaluate this effect, we conduct experiments on the MimicGen using 100 demonstrations and the same evaluation protocol described above. We compare the ISP-$\SO(2)$ with two variants: \emph{Pretraining}, which initializes the image encoder with an ImageNet-1k~\citep{russakovsky2015imagenet}-pretrained equivariant ResNet-18, and \emph{Scratch}, which trains the entire model from random initialization. Table~\ref{tab:sim_pretrain_all} reports the maximum evaluation success rates.

Results show that ISP-$\SO(2)$ (Pretraining) surpasses ISP-$\SO(2)$ (Scratch) by 7.1\%, indicating consistently improved final performance across most tasks. Moreover, the pretrained version with only 100 demonstrations achieves comparable performance to training from scratch with 200 demonstrations, further highlighting its data efficiency. These findings demonstrate the effectiveness of pretraining in providing richer and more stable visuomotor representations. Although the performance gains are marginal or absent in a few tasks, this suggests that naive pretraining may not always align perfectly with the downstream visuomotor learning objective. Developing pretraining strategies that are tailored to equivariant visuomotor policy representations is, therefore, a promising future direction.
\begin{figure*}[t]
\centering
\includegraphics[width=0.99\linewidth]{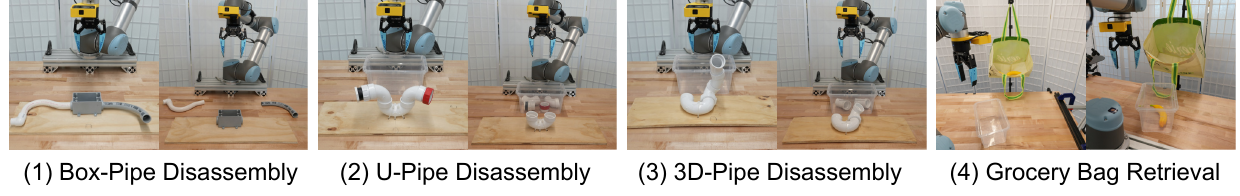}
\vspace{-0.2cm}
\caption{Real-world environments for evaluation. A GoPro camera is mounted on the robot's wrist to capture eye-in-hand observations. In each subfigure, the left image shows the initial state, while the right image shows the goal state. See Appendix~\ref{sec:real_world_details} for detailed task descriptions.}
\label{fig:real_tasks}
\vspace{-0.3cm}
\end{figure*}
\subsection{Real World}
\noindent\textbf{Physical Setups}\;
Our real robot experiments use a Universal Robot UR5 equipped with a Robotiq-85 Gripper and custom-designed soft fingers. A GoPro camera is mounted on the wrist, following prior setups~\citep{chi2024universal,ha2024umi,lin2024data}. Demonstrations are collected via the Gello teleoperation interface~\citep{wu2024gello}, with observations and actions recorded at 5~Hz. Following~\cite{chi2023diffusion, EquiDiff}, we employ DDIM~\citep{song2020denoising} to reduce the number of denoising steps to 16. Figure~\ref{fig:real_tasks} illustrates the four real-world manipulation tasks. The first three tasks involve pipe disassembly, each focusing on different challenges in closed-loop control: background-object segmentation (Box-Pipe), long-horizon control (U-Pipe), and handling complex 3D geometries (3D-Pipe). The fourth task involves retrieving objects from a deformable grocery bag, for which wrist-mounted camera observations are the only reliable source of visual information due to severe occlusions and limited external visibility. We compare ISP-$\SO(3)$ against the Diffusion Policy~\citep{chi2023diffusion}. Further details on the physical setup, task visualization, goal specification, and practical guidelines for data collection are provided in Appendix~\ref{sec:real_world_details} and Appendix~\ref{app:data_collection}.

\noindent\textbf{Results}\;Table~\ref{tab:real_world} reports success rates over 20 trials per task. Our method consistently outperforms the Diffusion Policy~\citep{chi2023diffusion} baseline, with significant improvements on Box-Pipe (80\% vs. 10\%) and 3D-Pipe (75\% vs. 15\%). The former benefits from more precise visual representations that distinguish the gray pipe from the gray box background, while the latter showcases the advantage of $\SO(3)$-equivariant features for reasoning over complex 3D geometries. The U-Pipe task also shows a notable gain (85\% vs. 65\%), demonstrating the sustained and stable performance of our equivariant method in the long-horizon task.
On the Grocery Bag task, which heavily relies on eye-in-hand perception, our method achieves a 95\% success rate. This shows its high stability and robustness. These results confirm the effectiveness of our equivariant design in addressing diverse manipulation challenges in the real world. See Appendix~\ref{sec:failure} for a detailed failure analysis. We further evaluate the computational efficiency of ISP in real-world settings, with a comprehensive discussion provided in Appendix~\ref{app:efficiency}. In addition, we discuss potential limitations and practical considerations of equivariance in Appendix~\ref{app:limitation_of_equi}.

\begin{wraptable}{r}{0.62\textwidth}
\setlength\tabcolsep{2.6pt}
\small
\centering
\caption{Real-world task performance over 20 trials. The number of demonstrations used for training each task is shown in the second row.}
\vspace{-2pt}
\begin{tabular}{lcccc}
\toprule
 & Box-Pipe & U-Pipe & 3D-Pipe & Grocery Bag \\
\midrule
\# Demos & 65 & 65 & 65 & 60\\
\midrule
ISP-$\SO(3)$  & 80\%(16/20) & 85\%(17/20) & 75\%(15/20) & 95\%(19/20) \\
DiffPo \cite{chi2023diffusion} & 10\%(2/20) & 65\%(13/20) & 15\%(3/20) & 75\%(15/20)\\
\bottomrule
\end{tabular}
\label{tab:real_world}
\vspace{-6pt}
\end{wraptable}

\begin{figure*}[!t]
    \centering
    \begin{subfigure}[t]{0.32\linewidth}
    \captionsetup{aboveskip=1.5pt}
        \centering
        \includegraphics[width=1\linewidth]{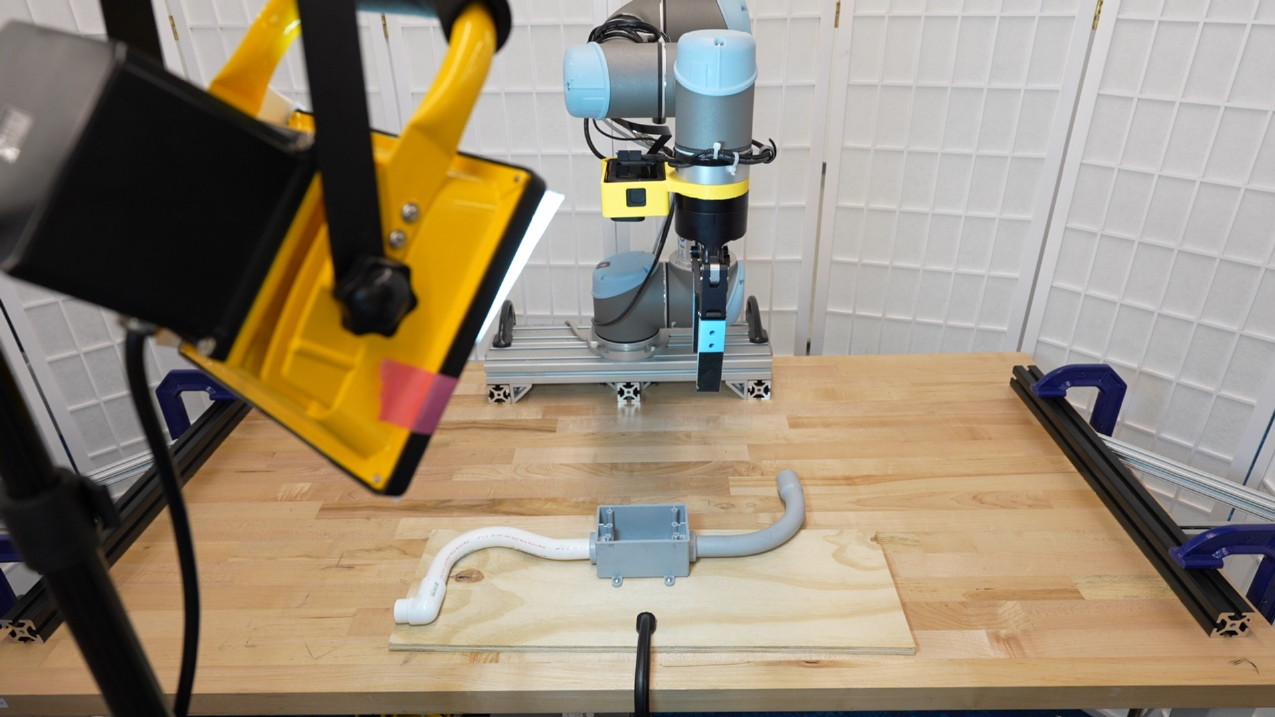}
        \caption{Lighting Change}
    \end{subfigure}
    \begin{subfigure}[t]{0.32\linewidth}
    \captionsetup{aboveskip=1.5pt}
        \centering
        \includegraphics[width=1\linewidth]{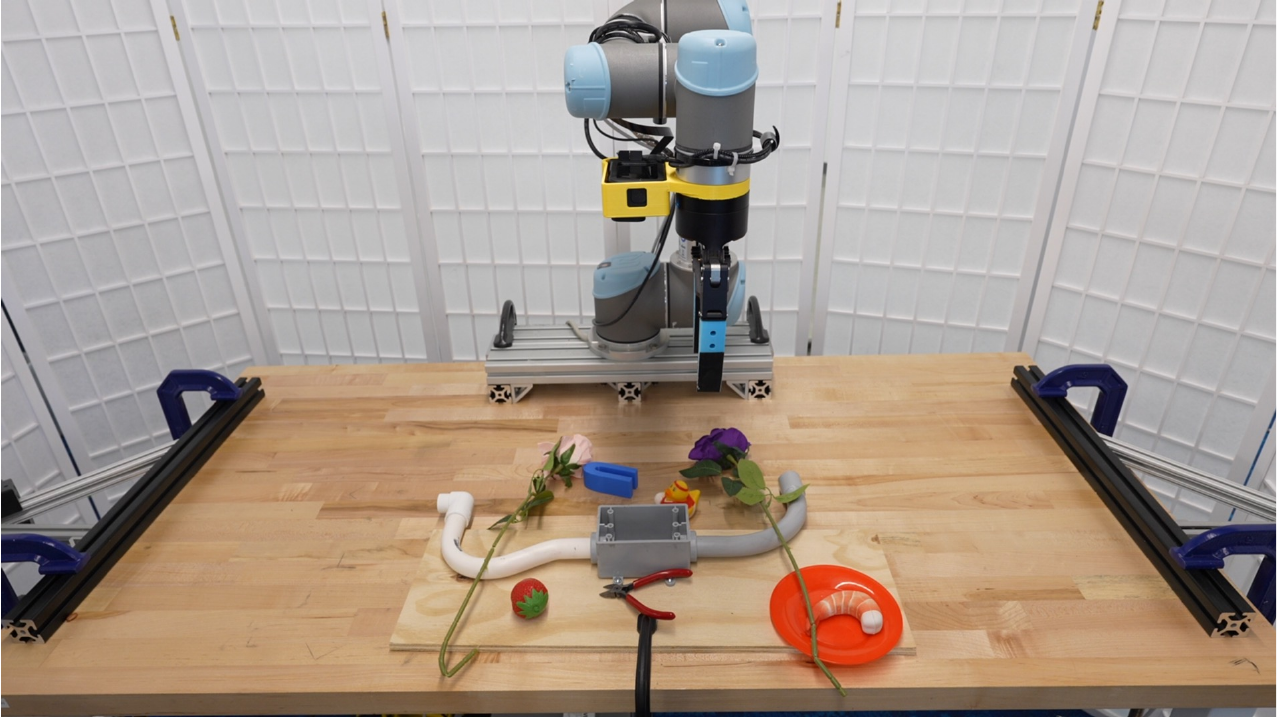}
        \caption{Background Clutter}
    \end{subfigure}
    \begin{subfigure}[t]{0.32\linewidth}
    \captionsetup{aboveskip=1.5pt}
        \centering
        \includegraphics[width=1\linewidth]{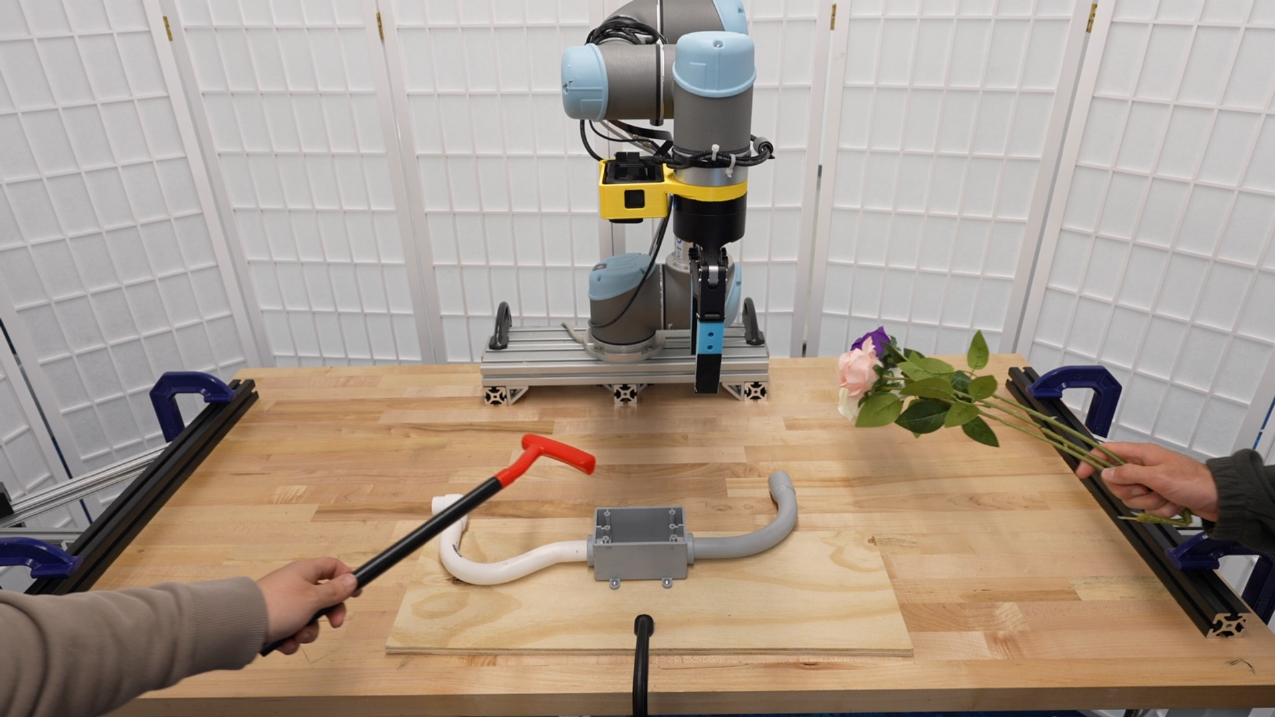}
        \caption{Partial Camera Occlusion}
    \end{subfigure}
    \caption{Real-world perturbation scenarios used to evaluate the robustness and generalization of our method on the Box-Pipe Disassembly task.}
    \vspace{-0.1cm}
    \label{fig:exp_perturb}
\end{figure*}

\noindent\textbf{Robustness to Real-World Perturbations}\;
To further evaluate the robustness and generalization ability of our policy, we conducted additional real-world experiments on the Box-Pipe Disassembly task under various domain shifts. First, we altered the lighting conditions by introducing a strong white point light source near the workspace, which substantially changed the shadows and color temperature of the scene (Figure~\ref{fig:exp_perturb}\kern0.07em a). Second, we perturbed the background by placing multiple household objects on the table to create clutter (Figure~\ref{fig:exp_perturb}\kern0.07em b). Finally, to test robustness against partial occlusion, we repeatedly and briefly blocked the eye-in-hand camera by rapidly waving different objects (e.g., a toy golf club and a flower) in front of it during policy rollout (Figure~\ref{fig:exp_perturb}\kern0.07em c). Using the same initial states and 20 rollouts per condition as in the previous real-world experiments, ISP-$\SO(2)$ achieves success rates of 85\% under lighting changes, 75\% with background clutter, and 85\% under partial camera occlusion. For reference, the performance of ISP-$\SO(3)$ without perturbations is 80\%. These results demonstrate that the proposed method generalizes well to real-world disturbances and maintains strong task performance under challenging visual conditions.

\section{Conclusion}
In this paper, we propose \textbf{Image-to-Sphere Policy (ISP)}, the first $\SO(3)$-equivariant policy learning framework for eye-in-hand visuomotor control using only monocular RGB inputs. By lifting 2D image features onto the sphere and introducing an equivariance correction mechanism to compensate for dynamic camera viewpoints, our method achieves global $\SO(3)$-equivariance and local $\SO(2)$-invariance without relying on depth sensors or multi-camera setups. This design enables robust and sample-efficient policy learning in dynamic, real-world settings. Extensive experiments in both simulation and real-world tasks demonstrate that ISP consistently outperforms strong baselines, achieving higher success rates with fewer demonstrations. Our work provides a general and effective algorithmic solution that is both deployable and scalable for eye-in-hand visuomotor learning.

\noindent\textbf{Limitations}\;
Our method has several limitations for future investigation. First, we only consider a single wrist-mounted RGB camera. While this view provides fine-grained local information, it lacks the global scene context that an agent-view camera could offer. Effectively combining these complementary perspectives remains an important challenge. Second, our approach models rotational equivariance but does not address translational equivariance. This limits the model's ability to generalize to object translations within the scene. Extending the equivariance correction to handle camera translations is a promising direction for future work. Third, the use of equivariant networks increases training time. Although inference remains efficient, reducing training overhead through more lightweight architectures would further enhance practicality. Fourth, our current method focuses on single-arm manipulation. Extending the framework to bimanual systems, where coordination between two arms is required, is a natural next step. Finally, our method does not yet leverage vision-language models. Integrating high-level semantic understanding through vision language models could further improve generalization and task understanding in more diverse environments.

\newpage
\begin{ack}
We would like to thank Rachel Lim and Andrew Cole for their assistance with the real-world experiments as well as all members of the Helping Hands Lab for their valuable discussions and feedback on the manuscript. This work was supported in part by NSF grants 2107256, 2134178, 2314182, 2409351, 2442658, and NASA grant 80NSSC19K1474.

\end{ack}
\bibliographystyle{plainnat}
\bibliography{arxiv}
\newpage
\appendix
\section{Orthographic Projection Details}
\label{app:ortho-proj}
The orthographic projection in our method follows the approach of~\citep{i2s}, which lifts 2D feature maps onto the unit sphere in the camera frame through a signal remapping operation. Unlike traditional geometric projections that rely on explicit camera calibration or depth information, our projection is entirely learned and does not depend on a predefined 3D center or physical camera parameters.

In practice, the spherical signal is sampled on a HEALPix grid, which provides an equal-area, hierarchical discretization of the sphere. For each point on the sphere, we apply a learnable weighted aggregation over the entire 2D feature map to compute its corresponding signal value. This design allows the network to flexibly determine how spatial features are mapped onto the sphere, rather than relying on fixed projection kernels. This is particularly useful for wide FOV images, where classical orthographic lifting can introduce distortions near image boundaries. By learning the mapping, the model can implicitly compensate for such distortions. However, this also means that robustness to changes in camera intrinsics (e.g., different FOVs or lens distortions) is not explicitly enforced. A promising future direction is to train the projection module under diverse intrinsic settings to support the model to learn a more general and transferable projection function.

\section{Spectral Realization of the Equivariance Correction}
\label{app:spectral-corr}
In this section, we provide a concrete spectral realization of the equivariance correction introduced in Proposition~\ref{prop:end-eff}, using the spherical-harmonic coefficients and Wigner $D$-matrices.
\begin{proof}
Let $x$ be the observation with camera pose $R_x \in \SO(3)$ and let $c_\ell(x) \in \mathbb{R}^{2\ell+1}$ denotes spherical harmonic coefficients. Under a global rotation $g \in \SO(3)$ applied to both the scene and the camera, the camera pose transforms as $R_x \mapsto R_{g x} = g R_x$. Since the signal $\Phi(x)$ is expressed in the local (camera) frame, the spherical coefficients remain unchanged under the global transformation, so $c_\ell(g x) = c_\ell(x)$. Applying Equation~\ref{eq:wigner} with the updated camera pose, the corrected coefficients at $g x$ are:
\begin{equation}
c_{\ell,\text{corr}}(g x) = D^\ell(R_{g x})\, c_\ell(g x)
= D^\ell(g R_x)\, c_\ell(x).
\end{equation}
Since the Wigner $D$-matrices $D^\ell$ form a group representation of $\SO(3)$, they satisfy the homomorphism property: $D^\ell(gR_x)=D^\ell(g)\,D^\ell(R_x)$.
Substituting this, we obtain:
\begin{equation}
c_{\ell,\text{corr}}(g x) = D^\ell(g)\, D^\ell(R_x)\, c_\ell(x)
\end{equation}
Recognizing that $c_{\ell,\text{corr}}(x) = D^\ell(R_x)\, c_\ell(x)$ by Proposition~\ref{prop:end-eff}, we conclude:
\begin{equation}
c_{\ell,\text{corr}}(g x) =  D^\ell(g)\, c_{\ell,\text{corr}}(x)
\end{equation}
Thus, the corrected coefficients $c_{\ell,\text{corr}}(x)$ transform equivariantly under the group action $g\in \SO(3)$.
\end{proof}
This result shows that equivariance correction can be implemented spectrally by left-multiplying the spherical harmonic coefficients with Wigner $D$-matrices according to the camera orientation. This aligns the signal, originally expressed in the camera frame, to a common world frame for consistent and equivariant downstream processing across varying viewpoints.

\section{Implementation of Our Policy}
\label{appendix:implement_details}
Our model consists of an $\SO(3)$-equivariant observation encoder followed by an $\SO(3)$-equivariant diffusion module, both implemented using \texttt{escnn}~\citep{escnn} and \texttt{e3nn} ~\citep{e3nn}.

Given an observation $x\in X$, the $\SO(2)$-equivariant image encoder $\lambda$ first maps the RGB image $I$ into a regular representation, which is then mapped to a trivial representation $\lambda(I)\in\mathbb{R}^{n\times h\times w}$, where $n$, $h$, and $w$ denote the number of channels, height, and width, respectively. These 2D features are lifted to the sphere via orthographic projection, producing a signal $\Phi(x)$ on $\Ss$. To account for varying viewpoints, we use the gripper orientation $R_x$ as an \emph{equivariance correction} factor to align the spherical signal into a common reference frame. In our setup, the wrist-mounted camera is rigidly attached to the gripper, so the gripper orientation provides a fixed proxy for the camera pose. This approximation is sufficient for aligning the image features with the proprioceptive signals, and any minor misalignments can be further handled by the equivariant convolution layers. The corrected signal is then processed by a sequence of $\Ss\rightarrow\SO(3)$ and $\SO(3)\rightarrow\SO(3)$ spherical convolution layers to generate the signal $\Psi(x)$ on $\SO(3)$. The proprioceptive state is encoded using the irreps $\rho_{0}$ and $\rho_{1}$, and passed through $\SO(3)$-equivariant linear layers to yield Fourier coefficients of the same type as $\Psi(x)$. These are then concatenated with the image signal to form the global conditioning vector \(e_{\ob}\in\mathbb{R}^{u\times d_{\ob}}\), where $u$ is the number of channels and $d_{\ob}$ is the feature dimension. Similarly, the noisy action chunk $\ac^{k}$ is embedded into $e_{\ac}\in\mathbb{R}^{u\times d_{\ac}\times n}$, where $d_{\ac}$ denotes the number of action feature channels and $n$ the number of time steps. An inverse FFT is applied to sample both $e_{\ob}$ and $e_{\ac}$ onto discrete subgroups, either the icosahedral group $I_{60}\subset\SO(3)$ or the cyclic group $C_8\subset\SO(2)$, producing $e_{\ob}\in\mathbb{R}^{p\times d_{\ob}},\;
e_{\ac}\in\mathbb{R}^{p\times d_{\ac}\times n},$
where $p=60$ or $8$ is the number of group elements. For each group element $g\in I_{60}$ or $g\in C_8$, a shared $\SO(3)$- or $\SO(2)$-equivariant 1-D temporal U-Net processes the action sequence $e_{\ac}^{g}$, conditioned on the observation $e_{\ob}^{g}$ and diffusion step $k$. This design follows the point-wise equivariant processing strategy proposed in~\citep{EquiDiff}, ensuring equivariance across group elements. Finally, an equivariant decoder maps the denoised representation to the noise estimate $\epsilon^{k}$.

\section{Simulation Settings}
\label{sec:sim_settings}
\begin{figure*}[!h]
\centering
\includegraphics[width=0.99\linewidth]{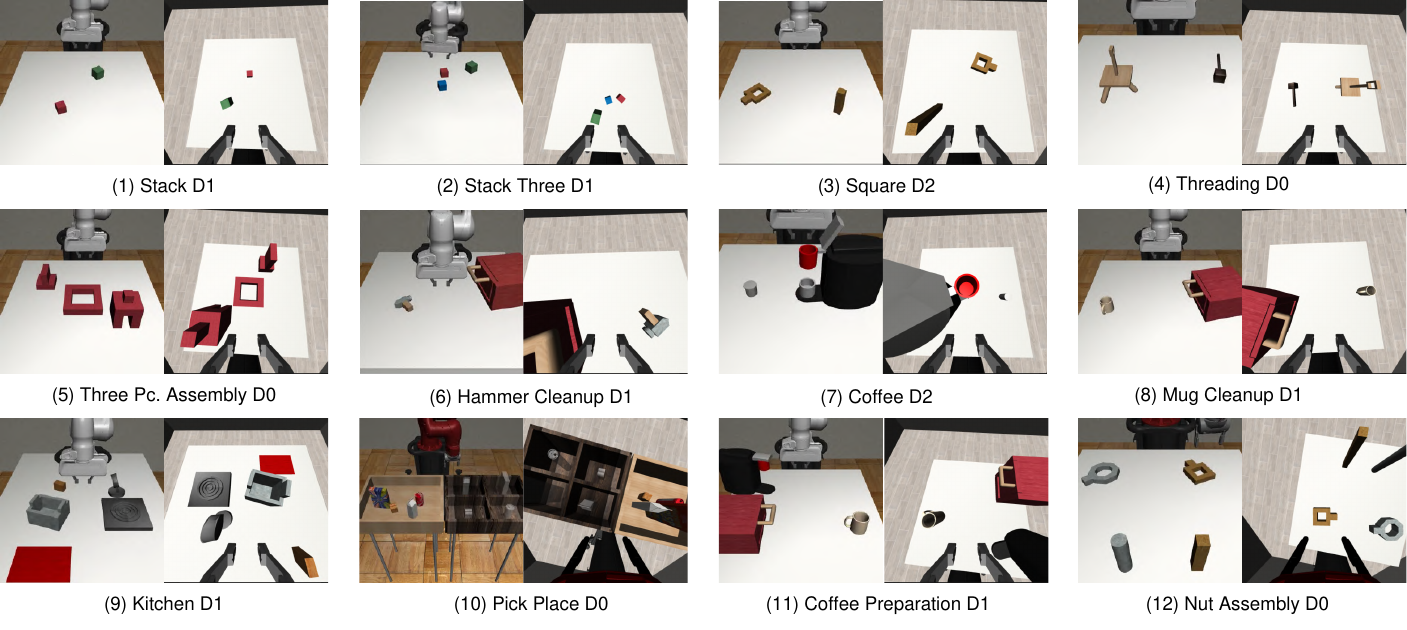}
\caption{The twelve simulation tasks from the MimicGen~\citep{mandlekar2023mimicgen} simulator. In each subfigure, the left image shows the task scene, while the right image shows the corresponding eye-in-hand view.} 
\label{fig:all_sim_tasks}
\end{figure*}

Figure~\ref{fig:all_sim_tasks} illustrates the twelve tasks in the MimicGen simulation. In each subfigure, the left image shows the full environment scene from the agent's view, while the right image is the eye-in-hand RGB observation used by the model. Following prior work~\citep{mandlekar2023mimicgen,EquiDiff}, we set the resolution of the eye-in-hand image to $3 \times 84 \times 84$ and adopt the same maximum episode length. To enable the wrist-mounted camera to capture more contextual information, we increase its FOV from 75 to 130 degrees, similar to that of a typical wide-angle camera.

\section{Training Details}
\label{sec:train_details}
For the simulation experiments, we follow the hyperparameter settings from prior work~\citep{EquiDiff,chi2023diffusion}. In detail, we use an observation window of two history steps for ISP-$\SO(3)$ and one step for ISP-$\SO(2)$. In both cases, the denoising network outputs a sequence of 16 action steps, which are used for optimization during training, while only the first 8 steps are executed during evaluation. During training, input images are randomly cropped to a resolution of $76 \times 76$, while a center crop is applied at evaluation time. We train all models using the AdamW~\citep{loshchilov2017decoupled} optimizer with Exponential Moving Average, and adopt the DDPM~\citep{ho2020denoising} framework with 100 denoising steps for both training and evaluation. For all baselines, we retain their original hyperparameter settings for evaluation and only adjust the number of training steps to ensure consistency across methods. All methods are trained on the same dataset and evaluated using three random seeds.

For the real-world experiments, we use the same hyperparameters as in the simulation, except that we replace DDPM with DDIM~\citep{song2020denoising} for both training and evaluation, and reduce the number of denoising steps to 16 at evaluation time. However, we find that using a resolution of $76 \times 76$ is insufficient for fine-grained manipulation in the real world, as the extremely wide FOV from the GoPro camera causes each pixel to correspond to a relatively large spatial region in the original setting. To address this, we increase the input resolution to $224\times224$. Specifically, starting from the original $720 \times 720$ RGB image captured using a GoPro with the Max Lens Mod, we apply a center crop of size $480 \times 480$, followed by resizing to $224 \times 224$. In addition, we apply standard data augmentations, including random cropping, rotation, and color jitter, to improve the robustness of both our method and the baselines.

All models are trained on single GPUs using compute clusters and workstations equipped with multiple high-performance consumer-grade GPUs.

\section{Full Simulation Experiment Results with Standard Deviations}
\label{appendix:seed_variance}
Table~\ref{tab:main_table_std} presents the same results as Table~\ref{tab:main_table}, with standard deviations included.

\begin{table*}[h]
\centering
\scriptsize
\renewcommand{\arraystretch}{1.2}
\caption{Maximum success rates (\%) on MimicGen tasks with 100 and 200 demonstrations across different methods, averaged over three random seeds. The $\pm$ indicates standard deviation.}
\setlength{\tabcolsep}{2pt}
\begin{tabular}{@{}l P{1.6cm}P{1.2cm} P{1.5cm}P{1.3cm}  P{1.5cm}P{1.2cm} P{1.5cm}P{1.2cm} @{}}
\toprule
& 
\multicolumn{2}{c}{Stack D1} &
\multicolumn{2}{c}{Stack Three D1} &
\multicolumn{2}{c}{Square D2} &
\multicolumn{2}{c}{Threading D0} \\

\cmidrule(lr){2-3} \cmidrule(lr){4-5} \cmidrule(lr){6-7} \cmidrule(lr){8-9}
\textbf{Method} & 100 & 200 & 100 & 200 & 100 & 200 & 100 & 200  \\
\midrule
ISP-$\SO(3)$ & 99.3 $\pm$ 1.2 & 100.0 $\pm$ 0.0 & 70.0 $\pm$ 2.0 & 88.0 $\pm$ 2.0 & 34.7 $\pm$ 4.2 & 51.3 $\pm$ 2.3 & 90.0 $\pm$ 2.0  & 92.0 $\pm$ 0.0\\ \rowcolor{gray!10}
ISP-$\SO(2)$ & 98.0 $\pm$ 2.0 & 100.0 $\pm$ 0.0 & 74.7 $\pm$ 7.6 & 88.0 $\pm$ 2.0 & 32.0 $\pm$ 0.0 & 50.7 $\pm$ 5.0 & 84.7 $\pm$ 1.2  & 87.3 $\pm$ 3.1\\
DiffPo      & 90.7 $\pm$ 4.2 &  96.0 $\pm$ 2.0 & 43.3 $\pm$ 4.2 & 76.7 $\pm$ 4.2 & 12.0 $\pm$ 2.0 & 25.3 $\pm$ 3.1 & 77.3 $\pm$ 10.3 & 86.7 $\pm$ 7.0\\
\rowcolor{gray!10}
EquiDiff    & 96.0 $\pm$ 0.0 &  98.7 $\pm$ 1.2 & 61.3 $\pm$ 5.0 & 80.0 $\pm$ 2.0 &  8.7 $\pm$ 1.2 & 19.3 $\pm$ 1.2 & 88.7 $\pm$ 5.8  & 92.0 $\pm$ 2.0\\
ACT         & 45.3 $\pm$ 7.6 &  77.3 $\pm$ 2.3 & 12.0 $\pm$ 2.0 & 36.7 $\pm$ 9.9 &  2.7 $\pm$ 1.2 & 10.0 $\pm$ 2.0 & 36.0 $\pm$ 6.9  & 53.3 $\pm$ 6.1\\

\midrule
&
\multicolumn{2}{c}{Three Pc. Assembly D0} &
\multicolumn{2}{c}{Hammer Cleanup D1} & 
\multicolumn{2}{c}{Mug Cleanup D1} &
\multicolumn{2}{c}{Coffee D2} 
 \\
\cmidrule(lr){2-3} \cmidrule(lr){4-5} \cmidrule(lr){6-7} \cmidrule(lr){8-9}
\textbf{Method} & 100 & 200 & 100 & 200 & 100 & 200 & 100 & 200 \\
\midrule
ISP-$\SO(3)$ & 70.7 $\pm$ 1.2 & 79.3 $\pm$ 1.2	& 66.0 $\pm$ 0.0 & 73.3 $\pm$ 1.2  & 54.0 $\pm$ 8.7 & 58.7 $\pm$ 2.3 & 64.0 $\pm$ 0.0 & 68.7 $\pm$ 3.1\\
\rowcolor{gray!10}
ISP-$\SO(2)$ & 74.7 $\pm$ 1.2 & 80.0 $\pm$ 2.0 & 70.7 $\pm$ 1.2 & 73.3 $\pm$ 2.3  & 56.0 $\pm$ 2.0 & 60.7 $\pm$ 1.2 & 58.7 $\pm$ 3.1 & 63.3 $\pm$ 4.2\\
DiffPo      & 72.7 $\pm$ 3.1 & 73.3 $\pm$ 2.3 & 58.7 $\pm$ 7.6 & 63.3 $\pm$ 11.7 & 49.3 $\pm$ 8.3 & 61.0 $\pm$ 1.7 & 53.3 $\pm$ 3.1 & 54.7 $\pm$ 4.2 \\
\rowcolor{gray!10}
EquiDiff    & 74.0 $\pm$ 5.3 & 78.7 $\pm$ 1.2 & 59.3 $\pm$ 4.2 & 74.0 $\pm$ 2.0  & 50.7 $\pm$ 2.3 & 62.0 $\pm$ 0.0 & 47.3 $\pm$ 3.1 & 61.3 $\pm$ 2.3\\
ACT         & 28.0 $\pm$ 4.0 & 50.0 $\pm$ 5.3 & 34.7 $\pm$ 2.3 & 62.7 $\pm$ 5.8  & 24.7 $\pm$ 3.1 & 37.3 $\pm$ 5.8 & 20.7 $\pm$ 3.1 & 34.7 $\pm$ 2.3\\

\midrule &
\multicolumn{2}{c}{Kitchen D1} &
\multicolumn{2}{c}{Pick Place D0} &
\multicolumn{2}{c}{Coffee Preparation D1} &
\multicolumn{2}{c}{Nut Assembly D0} \\
\cmidrule(lr){2-3} \cmidrule(lr){4-5} \cmidrule(lr){6-7} \cmidrule(lr){8-9} 
\textbf{Method} & 100 & 200 & 100 & 200 & 100 & 200 & 100 & 200 \\
\midrule
ISP-$\SO(3)$ & 75.3 $\pm$ 3.1 & 79.3 $\pm$ 4.2 & 42.0 $\pm$ 4.4 & 65.7 $\pm$ 5.5 & 40.7 $\pm$ 2.3 & 61.3 $\pm$ 2.3 & 75.3 $\pm$ 2.5 & 82.0 $\pm$ 7.5\\
\rowcolor{gray!10}
ISP-$\SO(2)$ & 64.7 $\pm$ 2.3 & 72.0 $\pm$ 2.0 & 45.7 $\pm$ 8.0 & 61.0 $\pm$ 5.6 & 46.7 $\pm$ 4.6 & 56.0 $\pm$ 0.0 & 73.7 $\pm$ 7.6 & 84.3 $\pm$ 1.5\\
DiffPo      & 60.7 $\pm$ 8.1 & 70.7 $\pm$ 3.1 & 36.3 $\pm$ 2.1 & 47.7 $\pm$ 1.5 & 37.3 $\pm$ 1.2 & 52.0 $\pm$ 5.3 & 51.3 $\pm$ 3.8 & 62.3 $\pm$ 1.5\\
\rowcolor{gray!10}
EquiDiff    & 55.3 $\pm$ 1.2 & 66.7 $\pm$ 2.3 & 27.7 $\pm$ 2.9 & 46.3 $\pm$ 3.5 & 27.3 $\pm$ 1.2 & 38.7 $\pm$ 2.3 & 40.0 $\pm$ 4.0 & 56.3 $\pm$ 3.1\\
ACT         & 21.3 $\pm$ 1.2 & 50.7 $\pm$ 3.1 &  8.7 $\pm$ 1.5 & 13.7 $\pm$ 2.5 & 7.3 $\pm$ 2.3  & 16.0 $\pm$ 2.0 & 36.7 $\pm$ 1.2 & 49.0 $\pm$ 2.0\\
\bottomrule
\end{tabular}
\label{tab:main_table_std}
\end{table*}

\section{Invariance via Delta Control vs. Equivariance via Rotation Correction}
\label{app:delta}
One of the core components of our method is the rotation correction step, which aligns the spherical signals to a common reference frame to preserve $\SO(3)$-equivariance throughout the policy pipeline. A natural alternative is to remove this step and instead express actions in the moving gripper frame, referred to as \emph{delta actions} in~\citep{chi2024universal}, which can also be interpreted as a sequence of incremental transforms. This formulation leads to an $\SE(3)$-invariant system, as both perception and action are expressed relative to the gripper’s local frame. This raises an important question: \emph{Is rotation correction necessary if delta actions can achieve similar symmetry properties through invariance?}

\noindent\textbf{Empirical Evidence}\;
To investigate this, we conducted additional experiments comparing absolute and delta action control on two MimicGen tasks: Square D2 and Nut Assembly D0. Specifically, we evaluated (a) a variant of our method without rotation correction that uses delta control, and (b) the original Diffusion Policy with delta control. Table~\ref{tab:del_vs_abs} summarizes the results with 100 demonstrations. We observe that absolute action consistently outperforms delta action. Similar trends have been reported in prior work, such as EquiDiff~\citep{EquiDiff} and the Diffusion Policy~\citep{chi2023diffusion}, where delta or velocity control often results in inferior performance.

\begin{table*}[t]
\centering
\caption{Comparison of absolute and delta control on two MimicGen tasks with 100 demonstrations. Values in parentheses indicate performance differences relative to ISP-$\SO(2)$ (Absolute).}
\begin{tabular}{lcc}
\toprule
Method & Square D2 & Nut Assembly D0 \\
\midrule
ISP-$\SO(2)$ (Absolute) & 32 & 74 \\
ISP-$\SO(2)$ (Delta, No Rotation Correction) & 22 \,({\color{red}-10}) & 57 \,({\color{red}-17}) \\
DiffPo (Absolute) & 12 \,({\color{red}-20}) & 51 \,({\color{red}-23}) \\
DiffPo (Delta) & 14 \,({\color{red}-18}) & 22 \,({\color{red}-52}) \\
\bottomrule
\end{tabular}
\label{tab:del_vs_abs}
\end{table*}

\noindent\textbf{Generalization Perspective}\;
Although gripper-relative control guarantees invariance under single-camera setups, it does not generalize seamlessly to multi-camera or hybrid sensing configurations, where additional viewpoints can break this invariance assumption. In contrast, aligning both observations and actions to a shared world frame establishes a consistent global reference across all sensors. This property supports more flexible sensor integration and improved generalization in complex environments with multiple or moving viewpoints.

\begin{figure*}[!t]
\centering
\includegraphics[width=0.8\linewidth]{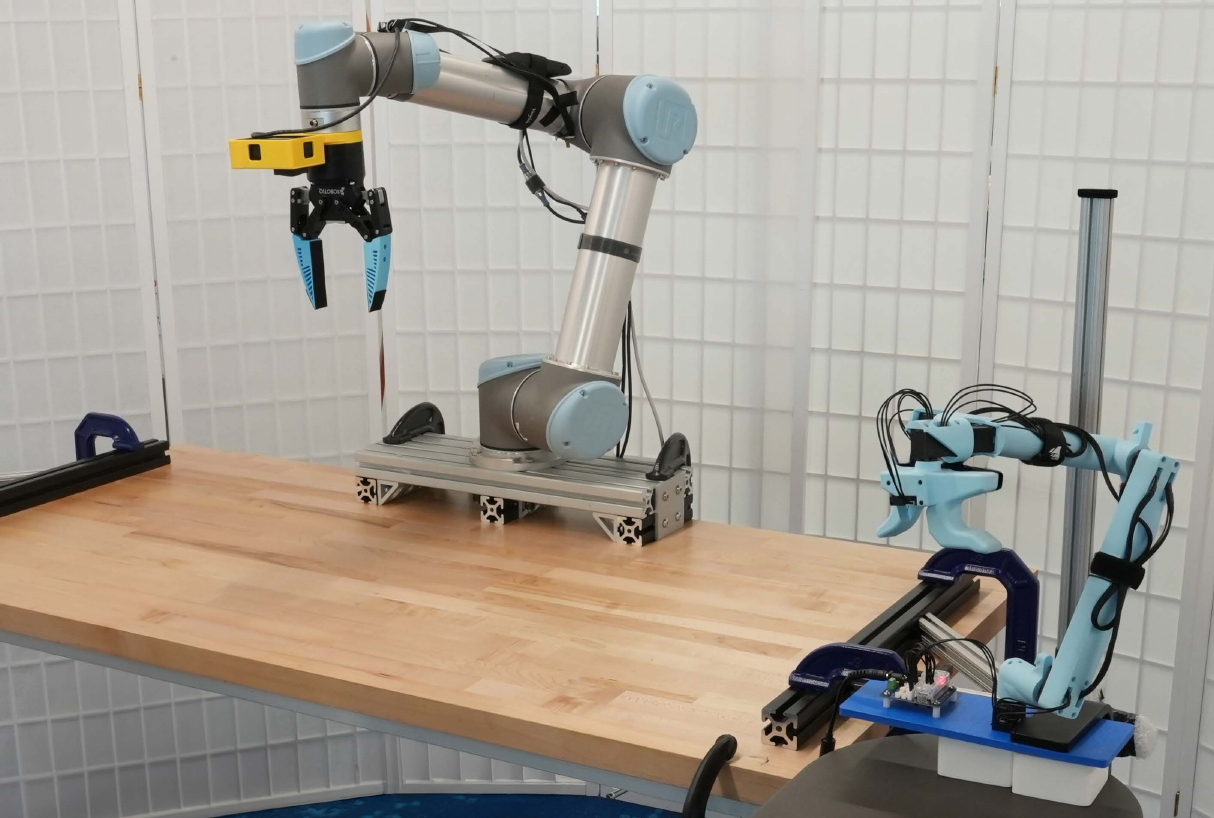}
\caption{Real-world experimental setup. We use a UR5 robot equipped with a Robotiq-85 gripper and custom-designed soft fingers. A GoPro camera is mounted on the wrist to capture visual observations. Demonstrations are collected using the Gello teleoperation interface (bottom right).}

\label{fig:exp_setup}
\end{figure*}
\section{Details of the Real-World Experiment}
\label{sec:real_world_details}
Figure~\ref{fig:exp_setup} shows our real-world experimental setup. Demonstrations are collected using the Gello teleoperation interface~\citep{wu2024gello}. While the robot is teleoperated in joint space, we record end-effector actions, including position, rotation, and gripper state. Visual observations and actions are recorded synchronously at each timestep.

\begin{figure*}[!t]
    \centering
    \begin{subfigure}[t]{0.24\linewidth}
        \centering
        \includegraphics[width=1\linewidth]{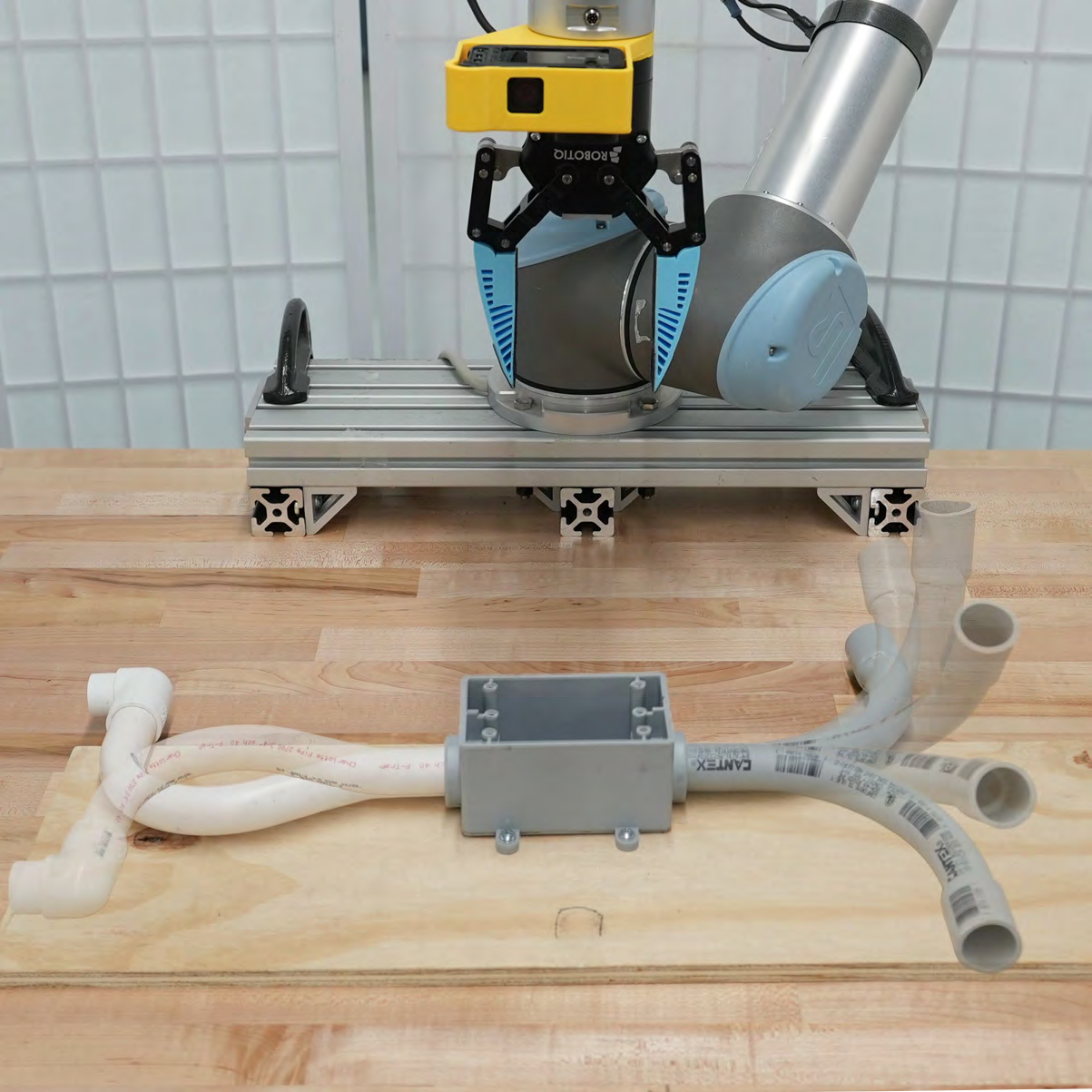}
        \caption{Box-Pipe Disassembly}
    \end{subfigure}
    \begin{subfigure}[t]{0.24\linewidth}
        \centering
        \includegraphics[width=1\linewidth]{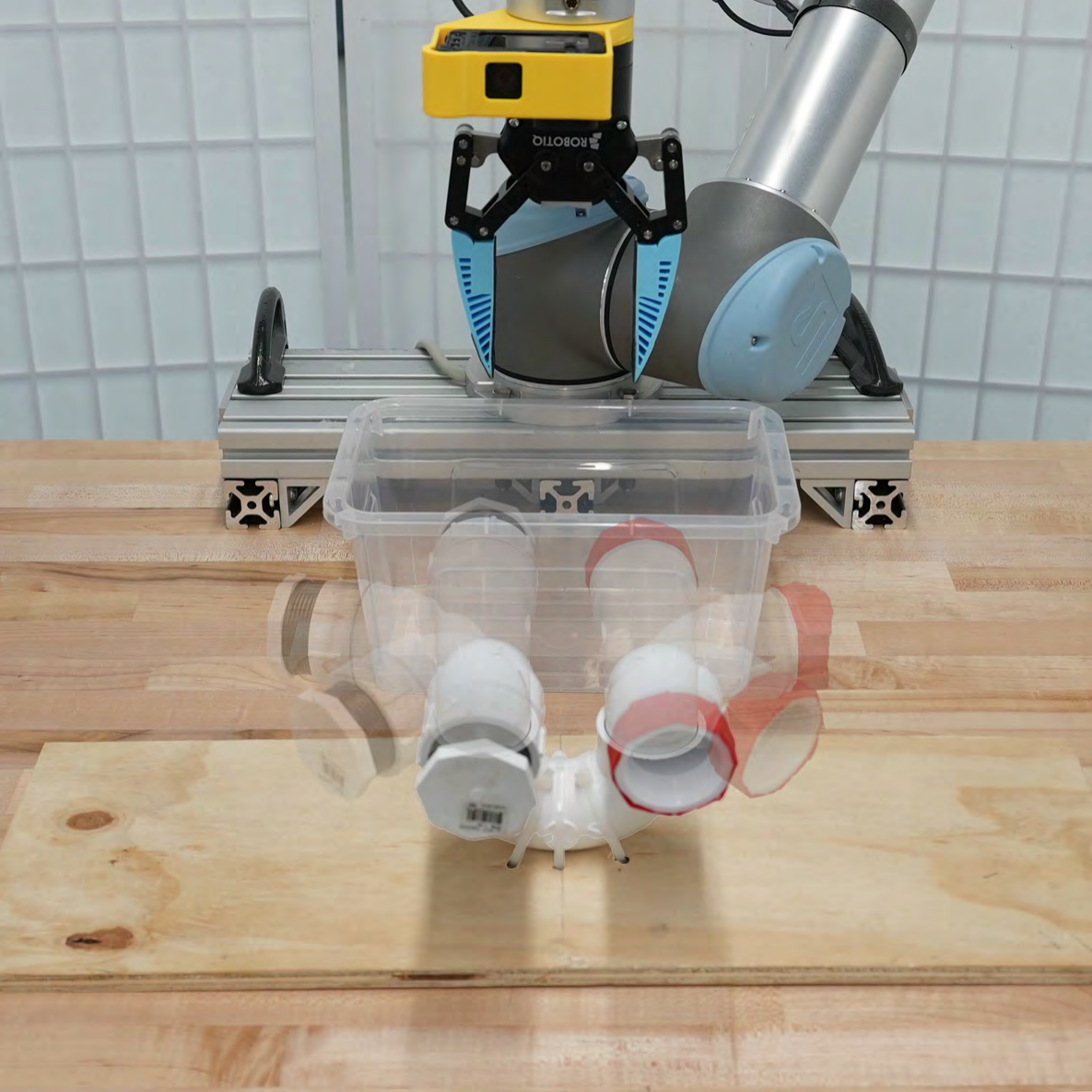}
        \caption{U-Pipe Disassembly}
    \end{subfigure}
    \begin{subfigure}[t]{0.24\linewidth}
        \centering
        \includegraphics[width=1\linewidth]{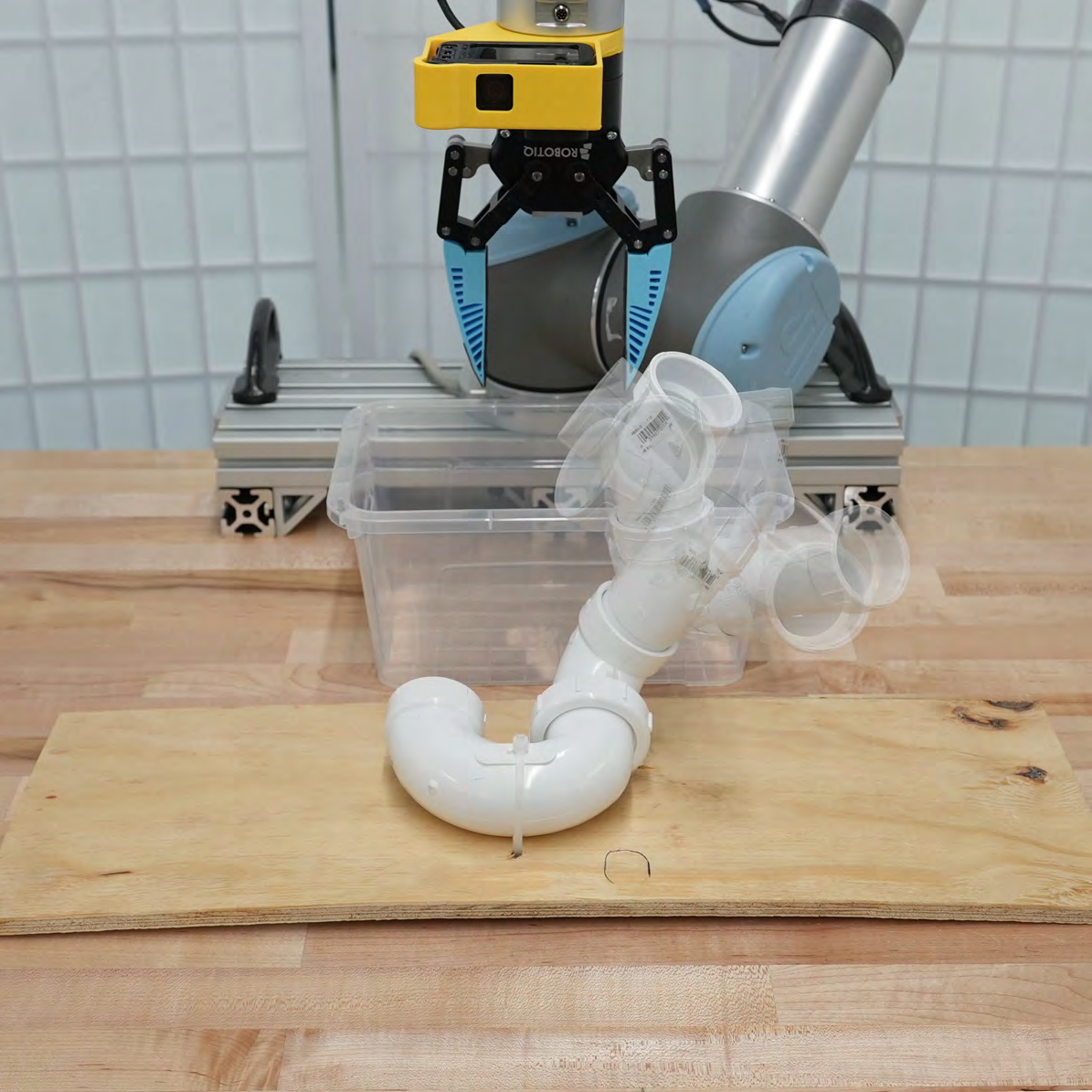}
        \caption{3D-Pipe Disassembly}
    \end{subfigure}
    \begin{subfigure}[t]{0.245\linewidth}
        \centering
        \includegraphics[width=1\linewidth]{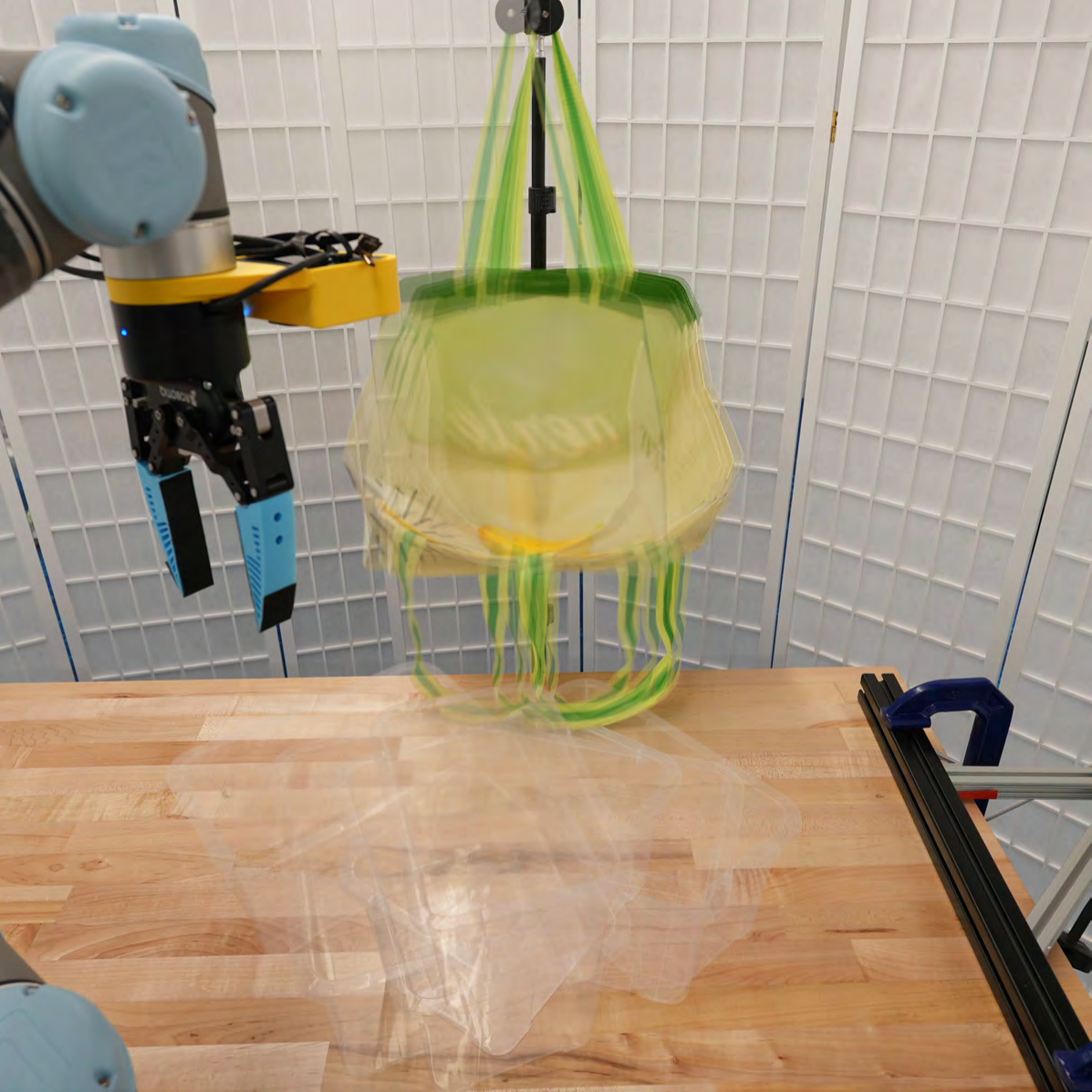}
        \caption{Grocery Bag Retrieval}
    \end{subfigure}
    \caption{Distribution of random initial states used in the real-world experiments.}
    \label{fig:exp_dis}
\end{figure*}

Figure~\ref{fig:exp_dis} illustrates the initial state distributions for each task. In \textit{Box-Pipe Disassembly}, two pipes with different colors are connected to a junction box, where one pipe shares the same color as the box may confuse the policy. The orientations of the pipes are randomized. In \textit{U-Pipe Disassembly}, four pipe fittings are arranged in a U-shape and initialized with random rotations. In \textit{3D-Pipe Disassembly}, two pipes are connected with independently randomized 3D orientations. In \textit{Grocery Bag Retrieval}, a toy banana is randomly placed inside a deformable plastic bag. The robot must reach into the bag, identify and retrieve the banana, and place it into a transparent container with minor positional variation. All subfigures in Figure~\ref{fig:exp_dis} show averaged visualizations across multiple randomized initializations.

We visualize one episode for each task in Figure~\ref{fig:all_rollout}. These tasks emphasize different aspects. The pipe disassembly tasks require precise, closed-loop control to smoothly extract the pipes. This makes them particularly challenging for open-loop policies. The Grocery Bag Retrieval task highlights the importance of the eye-in-hand camera, as the target object is difficult to perceive and localize using only external views.

\section{Practical Guidelines for Data Collection}
\label{app:data_collection}
Before starting data collection on the real robot, it is critical to establish a \textbf{predefined task execution strategy} to ensure motion simplicity, efficiency, and cross-operator consistency. Such a strategy typically involves defining consistent action sequences, execution ranges, and task progression patterns, helping to avoid ambiguous or poorly structured scenarios that may lead to robotic indecision or an undesirably large amount of multimodal behavior during training.

Based on our experience, during data collection, demonstrations should:
\begin{enumerate}[itemsep=3pt, topsep=0.8pt, leftmargin=2em]
    \item Uniformly cover as many task-relevant initial states as possible.
    \item Maintain a consistent end-effector speed within and between trajectories without interruption.
    \item Avoid unnecessary stops, pauses, or other irregular motion patterns.
    \item Synchronize sensing and control to minimize latency-induced artifacts.
    \item Regularly verify alignment between the robot and sensors to prevent drift and maintain data consistency.
\end{enumerate}

After data collection, all trajectories should be automatically or visually inspected to detect potential issues. In particular, segments exhibiting robotic hesitation or stalling, most commonly near the beginning and end of each demonstration, as well as episodes containing negative or low-quality behavior, should be identified and removed. Consistent inspection and pruning of low-quality data can significantly improve the stability and performance of policy learning.

These practical steps help ensure that the collected data are clean, diverse, and informative, which can ultimately enhance the robustness and generalization of learned visuomotor policies.

\section{Real World Experiment Failure Analysis}
In the \textit{Box-Pipe Disassembly} task, one of the primary failure cases arises from the inability to distinguish between the gray junction box and the gray pipe. For the original Diffusion Policy, the policy consistently misidentifies the box as the pipe to be disassembled, which triggers the robot's emergency stop. While our method occasionally encounters the same issue, the failure rate is significantly lower. This suggests that our method is more data-efficient and better at learning robust visual distinctions from limited demonstrations. 

In the \textit{U-Pipe Disassembly} task, a common failure mode for our method and the baseline occurs when pulling the red pipe inadvertently causes a connected pipe to be extracted as well. In such cases, the robot grasps both pipes simultaneously. We consider this a partial success. However, the baseline additionally suffers from incorrect orientation predictions under certain initial states, leading to more frequent failures. 

In the \textit{3D-Pipe Disassembly} task, our method occasionally fails to identify the correct grasp orientation. In contrast, the baseline struggles consistently with this issue and rarely completes the task successfully. One major contributing factor is the multimodality of the task. During data collection, it is difficult to maintain consistency in demonstration strategies because pipes can be grasped in multiple orientations. Nevertheless, by incorporating 3D symmetries, our method is more robust to such variations and generalizes better across diverse configurations.

In the \textit{Grocery Bag Retrieval} task, failure cases primarily result from unsuccessful grasp attempts or inaccuracies during the placement phase. The deformable nature of the bag and the partial occlusion of the banana present additional challenges, especially under limited visual feedback.

\label{sec:failure}

\begin{table*}[t]
\centering
\caption{Comparison of training and inference efficiency on a single RTX 4090 GPU.}
\label{tab:efficiency}
\begin{tabular}{lccc}
\toprule
\textbf{Method} & \textbf{Training Speed (rel.)} & \textbf{Inference Time (ms)} & \textbf{Real-Time Capable} \\
\midrule
DiffPo & $1\times$ & $68$  & Yes \\
ISP-$\SO(2)$    & $2.6\times$ slower & $63$  & Yes \\
ISP-$\SO(3)$      & $5.4\times$ slower & $75$ & Yes \\
\bottomrule
\end{tabular}
\end{table*}

\section{Computational Efficiency Analysis}
\label{app:efficiency}

In this section, we provide quantitative comparisons of the computational efficiency of our method during both training and inference. Our results are measured on a single RTX 4090 GPU.

\textbf{Training Efficiency} 
Compared to the original Diffusion Policy~\citep{chi2023diffusion}, ISP-$\SO(2)$ is approximately $2.6\times$ slower, and ISP-$\SO(3)$ is approximately $5.4\times$ slower during training. This increase is expected due to the added computational complexity of the equivariant layers. Nevertheless, the training speed remains practical for large-scale policy learning.

\textbf{Inference Efficiency}
Despite the higher training cost, our method maintains high efficiency during inference, making it well-suited for real-time deployment. Table~\ref{tab:efficiency} summarizes the average inference time of each method in the real-world settings with 16-step DDIM sampling~\citep{song2020denoising}. All methods exhibit comparable inference speeds. The $\SO(2)$ variant is slightly faster than the baseline, primarily due to its lighter-weight diffusion U-Net and the use of a smaller history observation window. Although the $\SO(3)$ variant is marginally slower, its inference time ($\sim$75\,ms) remains close to DiffPo ($\sim$68\,ms), well within real-time control requirements (e.g., 10\,Hz).

\section{On Limitations and Practical Considerations of Equivariance}
\label{app:limitation_of_equi}
\noindent\textbf{Interaction with Real-World Asymmetries}\;
Equivariance may face challenges in manipulation scenarios where asymmetries in the physical world are important. A representative example is tasks involving asymmetric robot kinematics, such as left-right differences in reachable workspace. Although equivariance allows the model to generalize across rotated scenes, joint limits are not preserved under rotation, which may lead to infeasible or suboptimal actions. Another example is manipulation involving heavy objects, where gravity breaks rotational symmetry in practice. An object that is easy to manipulate in one orientation may become unstable or infeasible to lift when rotated. Despite these challenges, prior work has shown that equivariant models can remain robust in the presence of symmetry-breaking factors such as visual appearance, camera pose, and shadows~\citep{wang2022surprising}. These asymmetries are already encoded in the input, allowing the model to learn appropriate behaviors without violating the equivariant structure. While cases where equivariance leads to performance degradation are relatively uncommon, they do highlight scenarios where symmetry-breaking mechanisms may be beneficial. A promising future direction is to augment equivariant architectures with non-equivariant components or task-specific inductive biases (e.g., gravity-aware priors or joint-limit encodings) to better capture real-world asymmetries.

\noindent\textbf{Discretization of Continuous Symmetry Groups}\;
In our framework, equivariance is enforced by sampling finite subgroups of $\SO(3)$ (e.g., $I_{60}$ or $C_8$) and applying a shared U-Net across the sampled group elements. This approximation may, in principle, introduce discrepancies for rotations outside the sampled set. Empirically, however, no significant performance degradation was observed. In fact, discrete subgroups often lead to superior performance compared to continuous irreducible representations, consistent with prior findings in equivariant learning~\citep{escnn, weiler2019general}. While this approach reduces the theoretical degree of symmetry, it provides a more scalable and expressive modeling strategy by avoiding the computational overhead and activation function constraints associated with continuous irreps. Similar strategies have also demonstrated strong empirical effectiveness in other robotics applications~\citep{EquiDiff, huang2024fourier}. To further mitigate potential limitations and improve generalization, we apply random $\SO(2)$ rotations as a data augmentation strategy to the end-effector pose in both proprioceptive inputs and actions during training. Additionally, subgroup sampling is not restricted to a single set: multiple sets can be employed in practice to increase angular coverage when necessary. Finally, while equivariant architectures introduce structural inductive biases, they do not inherently limit the model’s ability to generalize beyond the sampled rotations. With sufficient data diversity and augmentation, the network is able to interpolate smoothly across $\SO(3)$, thereby alleviating the potential impact of discretization.

\section{Broader Impact}
This work has several potential social impacts, both positive and negative. On the positive side, our proposed method enables more data-efficient and generalizable robot policy learning in 3D environments. This can facilitate the development of more robust and capable household robots, particularly in settings where labeled demonstrations are limited. Moreover, by leveraging geometric symmetries and closed-loop visuomotor control from wrist-mounted cameras, our method could lower the barrier for deploying autonomous robots in unstructured real-world environments, thereby expanding accessibility and utility.

However, as with many data-driven learning methods, our approach inherits limitations tied to the quality and intent of the training data. Since the robot policy is learned entirely through imitation, any unsafe, biased, or suboptimal behavior demonstrated during data collection may be reflected in the final policy. Furthermore, the increased autonomy enabled by our method underscores the importance of safety monitoring and responsible deployment, especially in applications involving human interaction.

\begin{figure*}[!h]
    \centering
    \vspace{2em}
    \begin{subfigure}[t]{1\textwidth}
        \centering
    \begin{subfigure}[t]{0.32\linewidth}
        \centering
        \includegraphics[width=1\linewidth]{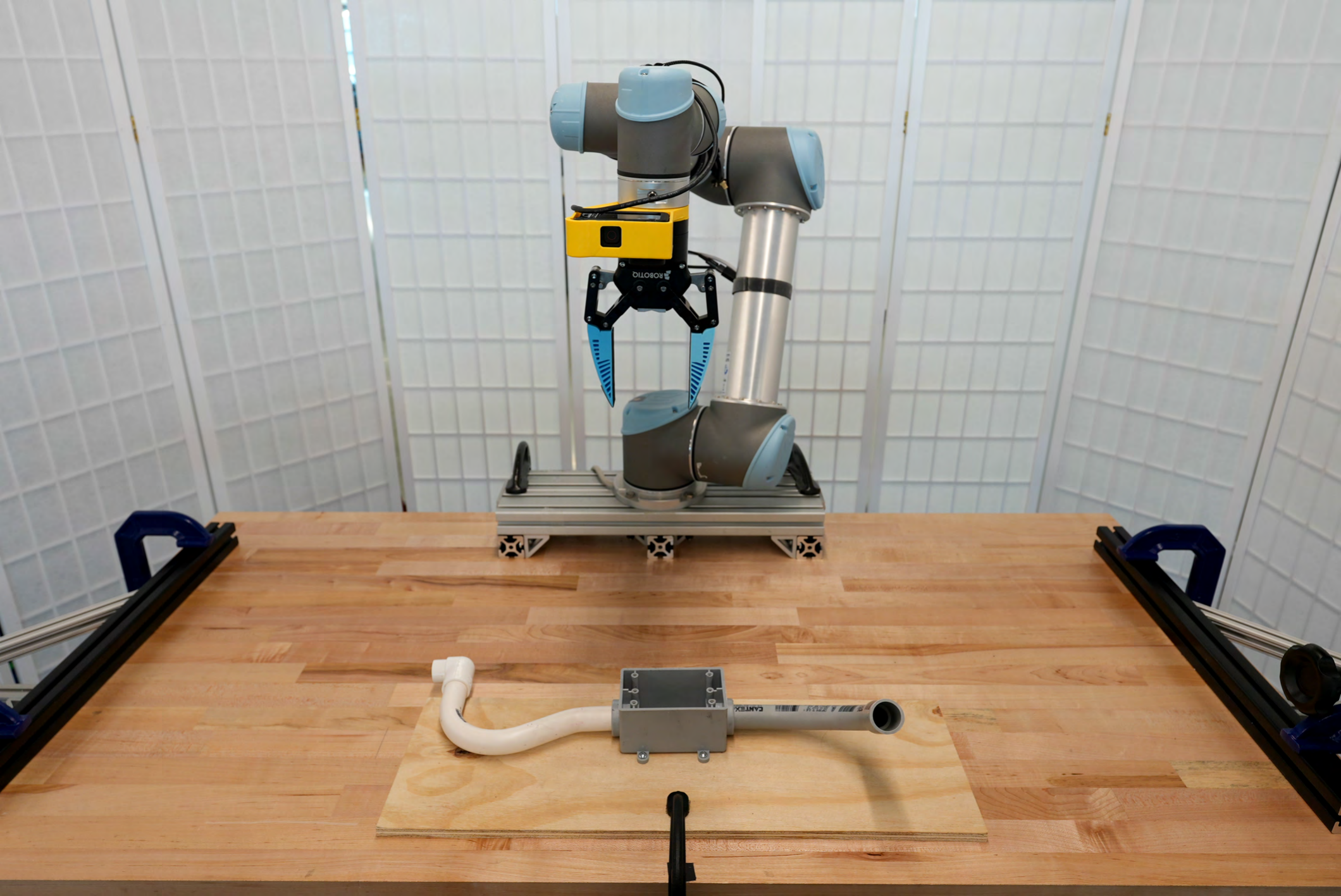}
    \end{subfigure}
    \begin{subfigure}[t]{0.32\linewidth}
        \centering
        \includegraphics[width=1\linewidth]{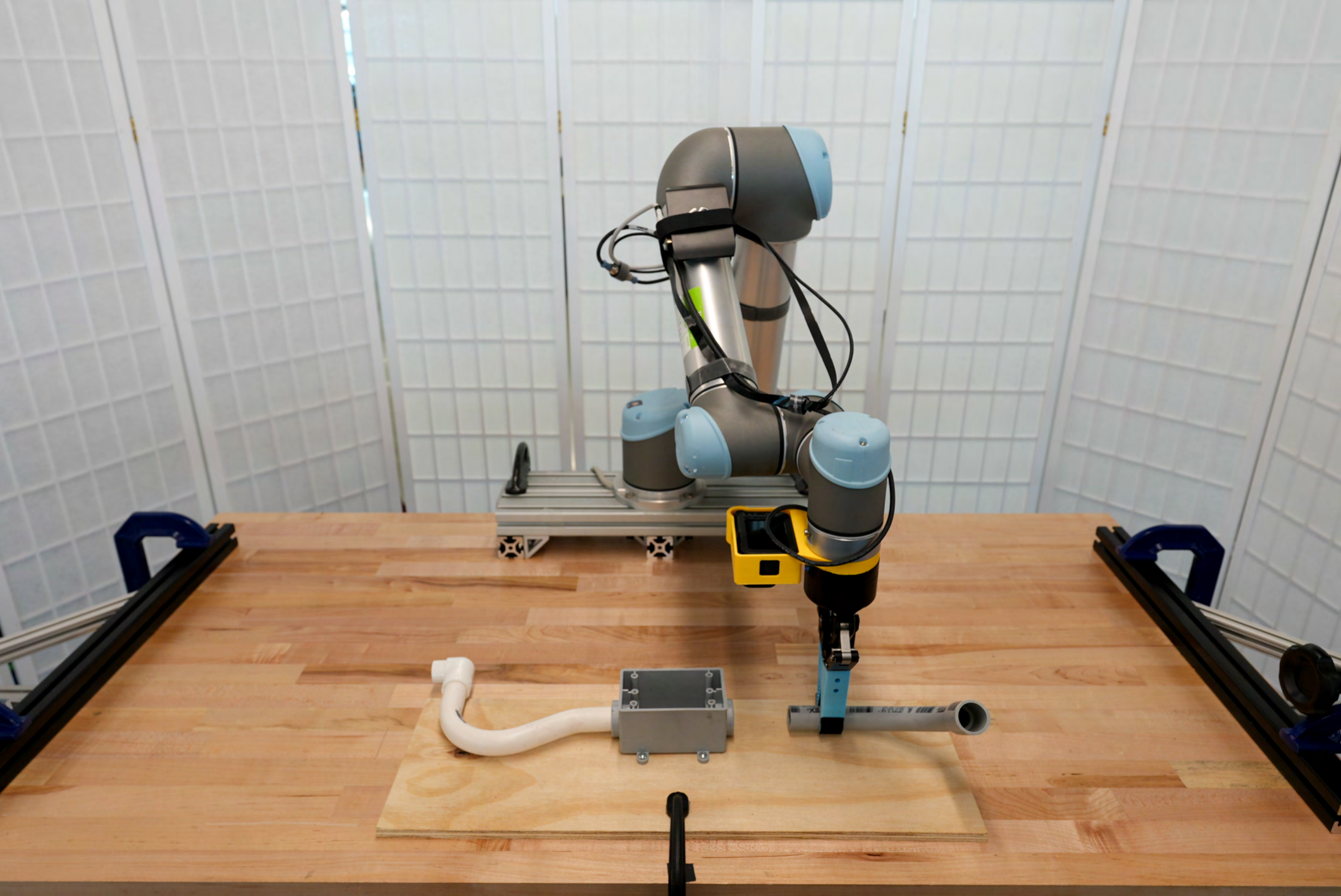}
    \end{subfigure}
    \begin{subfigure}[t]{0.32\linewidth}
        \centering
        \includegraphics[width=1\linewidth]{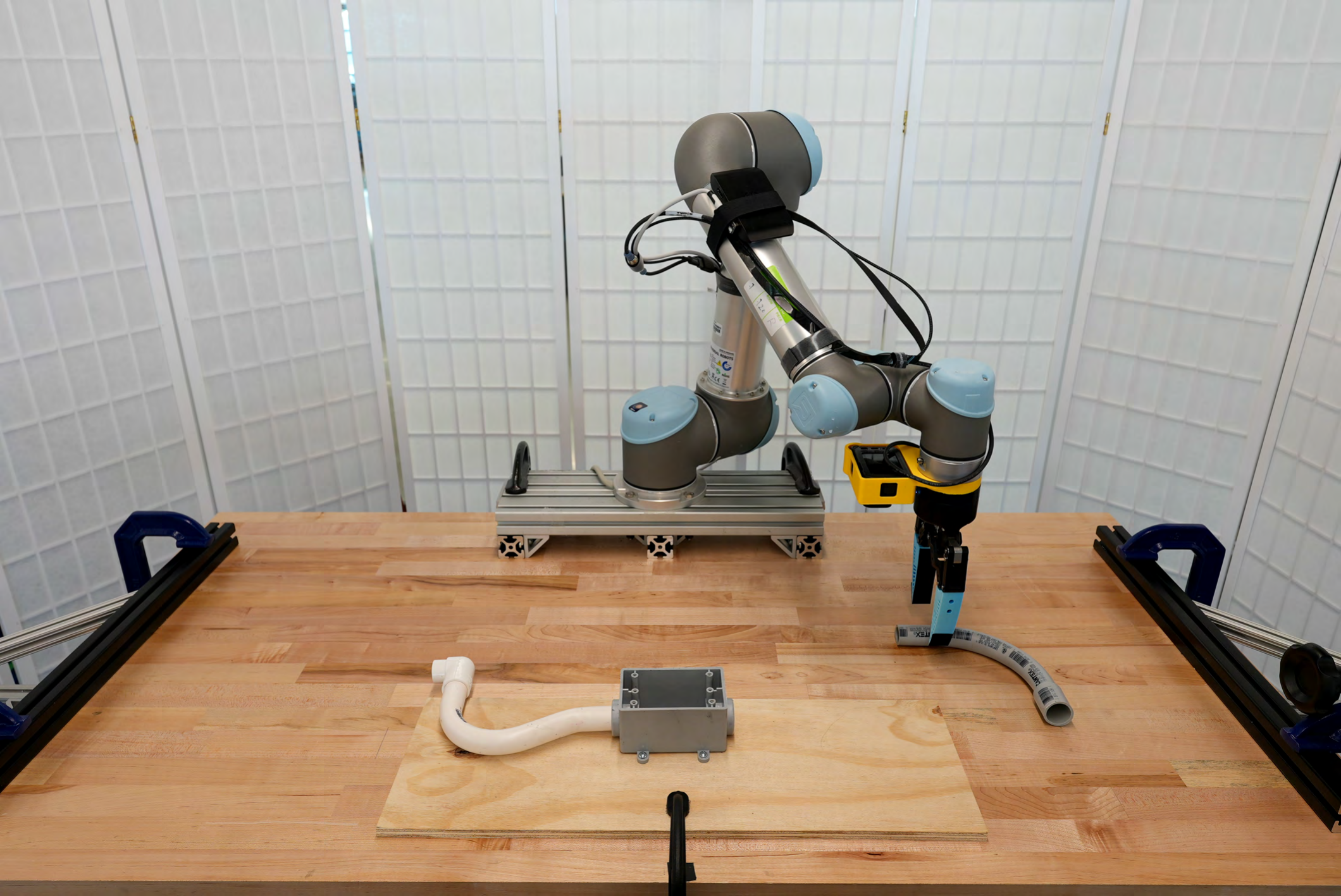}
    \end{subfigure}
    \begin{subfigure}[t]{0.32\linewidth}
        \centering
        \includegraphics[width=1\linewidth]{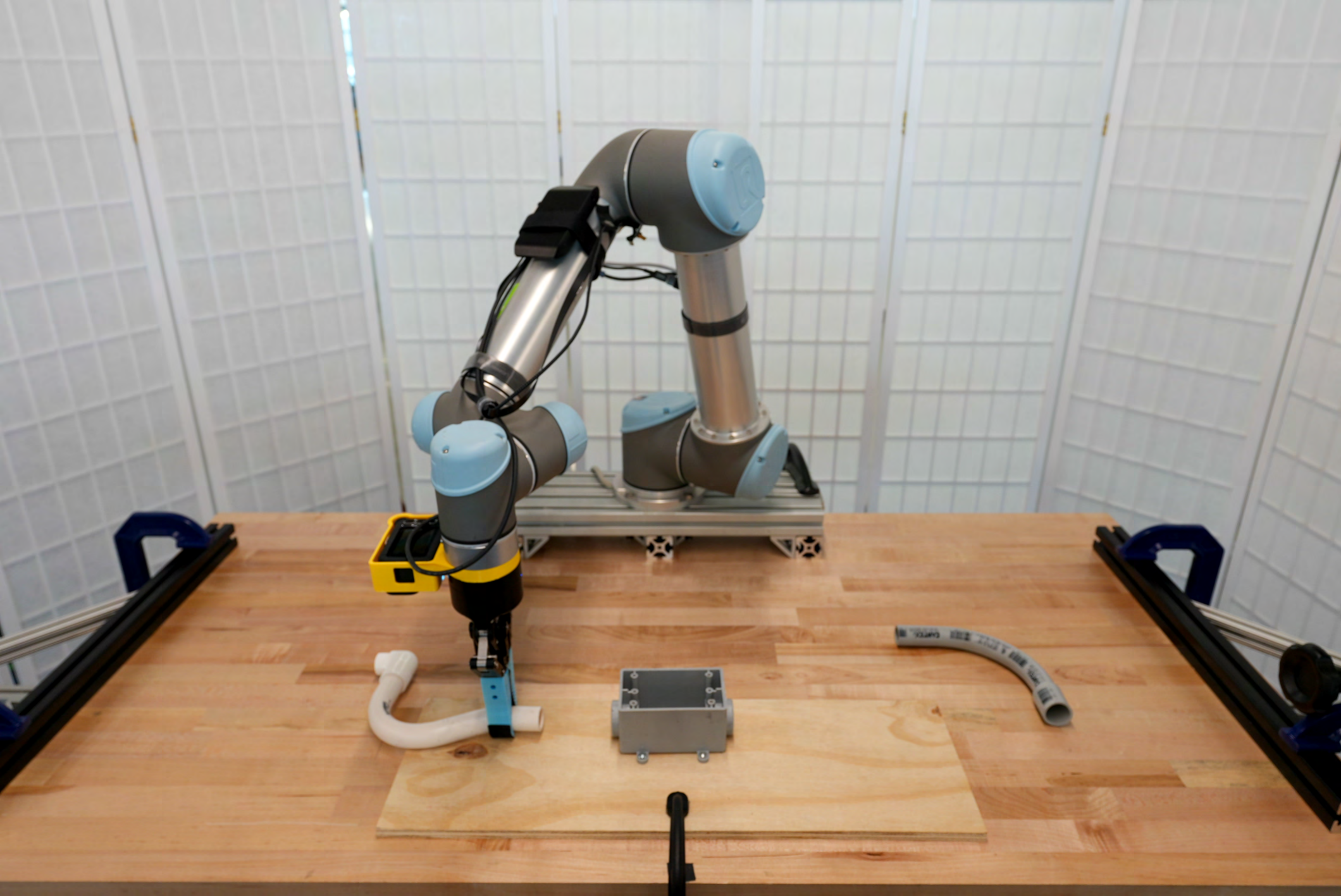}
    \end{subfigure}
    \begin{subfigure}[t]{0.32\linewidth}
        \centering
        \includegraphics[width=1\linewidth]{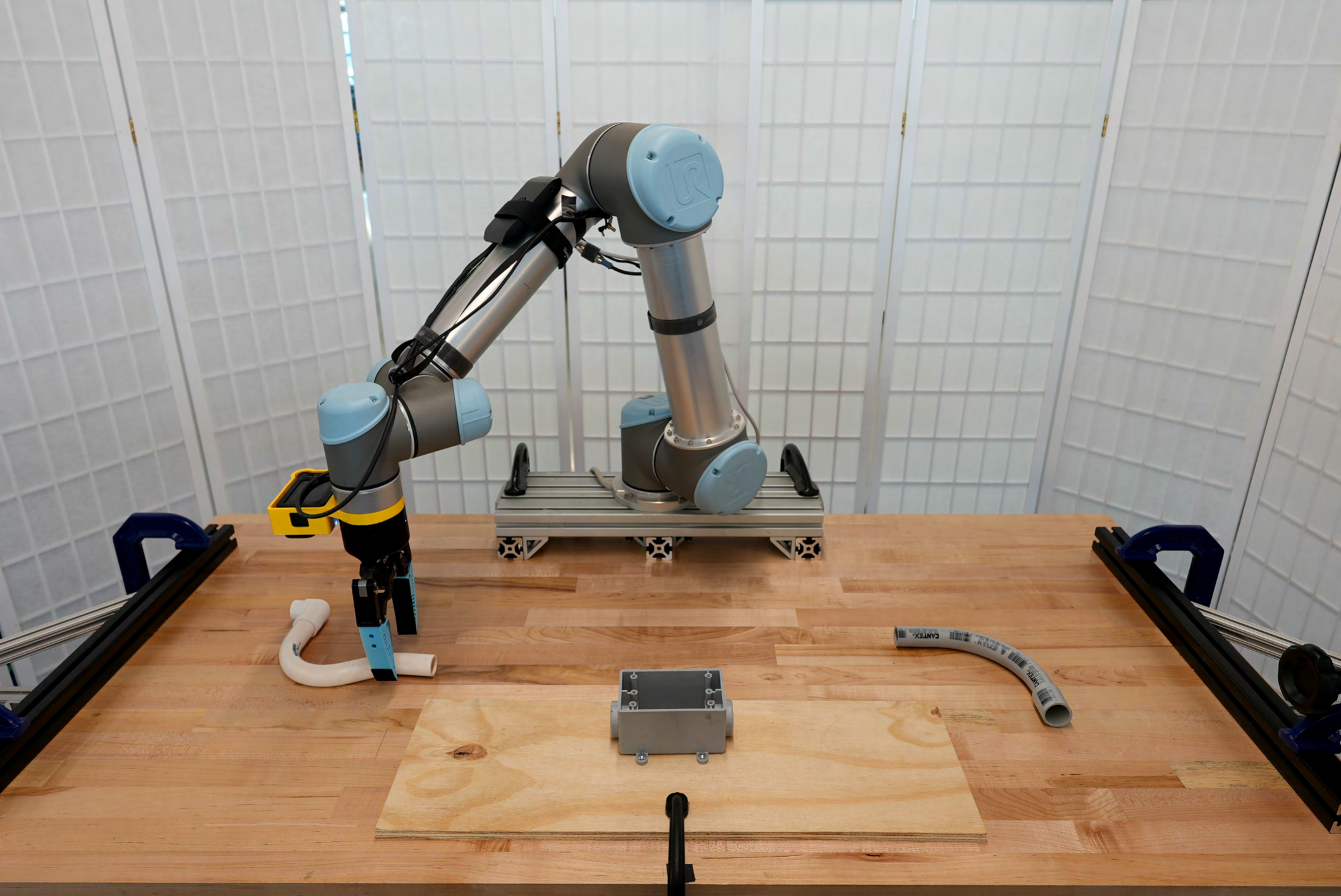}
    \end{subfigure}
    \begin{subfigure}[t]{0.32\linewidth}
        \centering
        \includegraphics[width=1\linewidth]{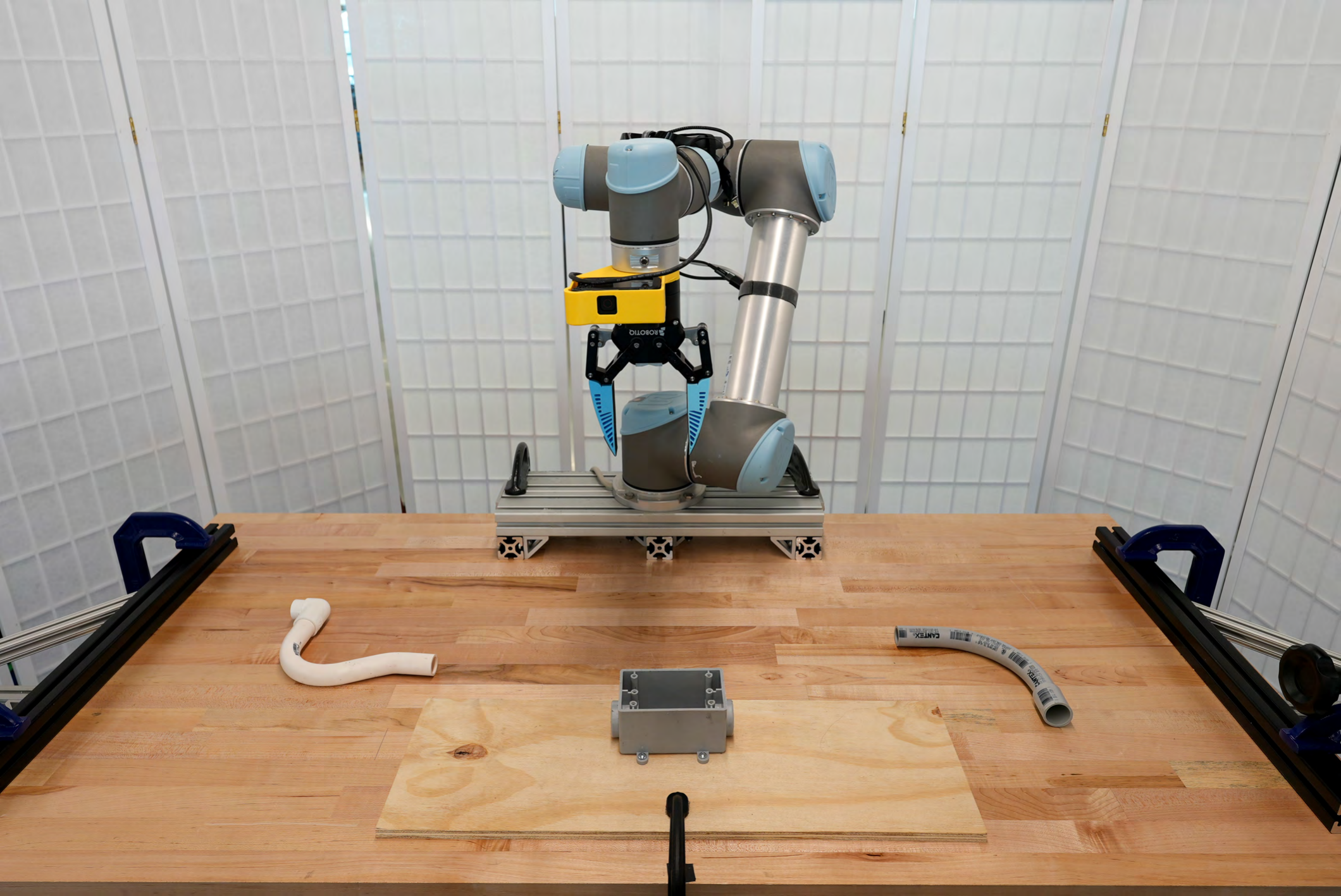}
    \end{subfigure}
    \caption{Box-Pipe Disassembly}
    \label{fig:box_pipe_rollout}
\end{subfigure}

     \begin{subfigure}[t]{1\textwidth}
        \centering
    \begin{subfigure}[t]{0.32\linewidth}
        \centering
        \includegraphics[width=1\linewidth]{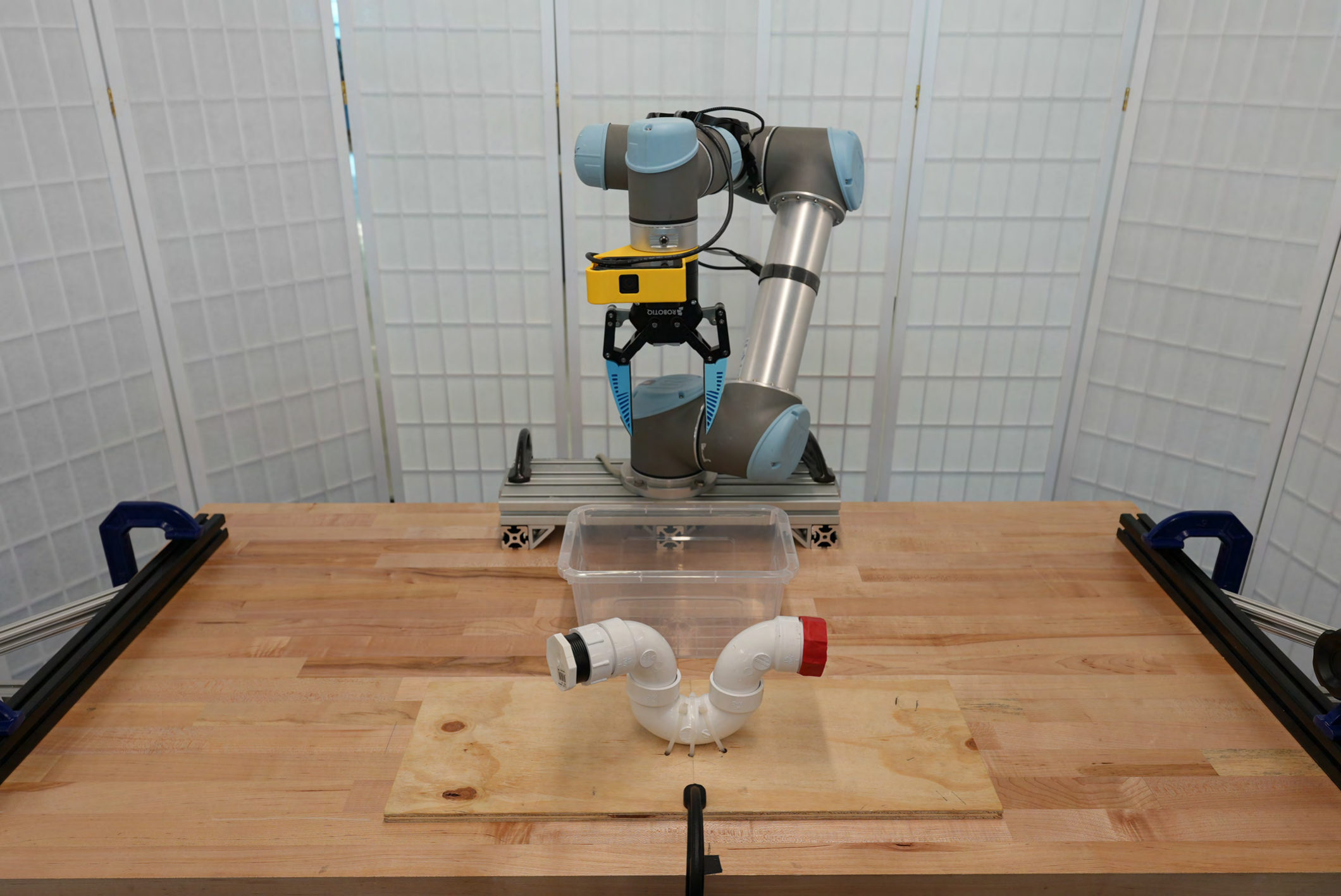}
    \end{subfigure}
    \begin{subfigure}[t]{0.32\linewidth}
        \centering
        \includegraphics[width=1\linewidth]{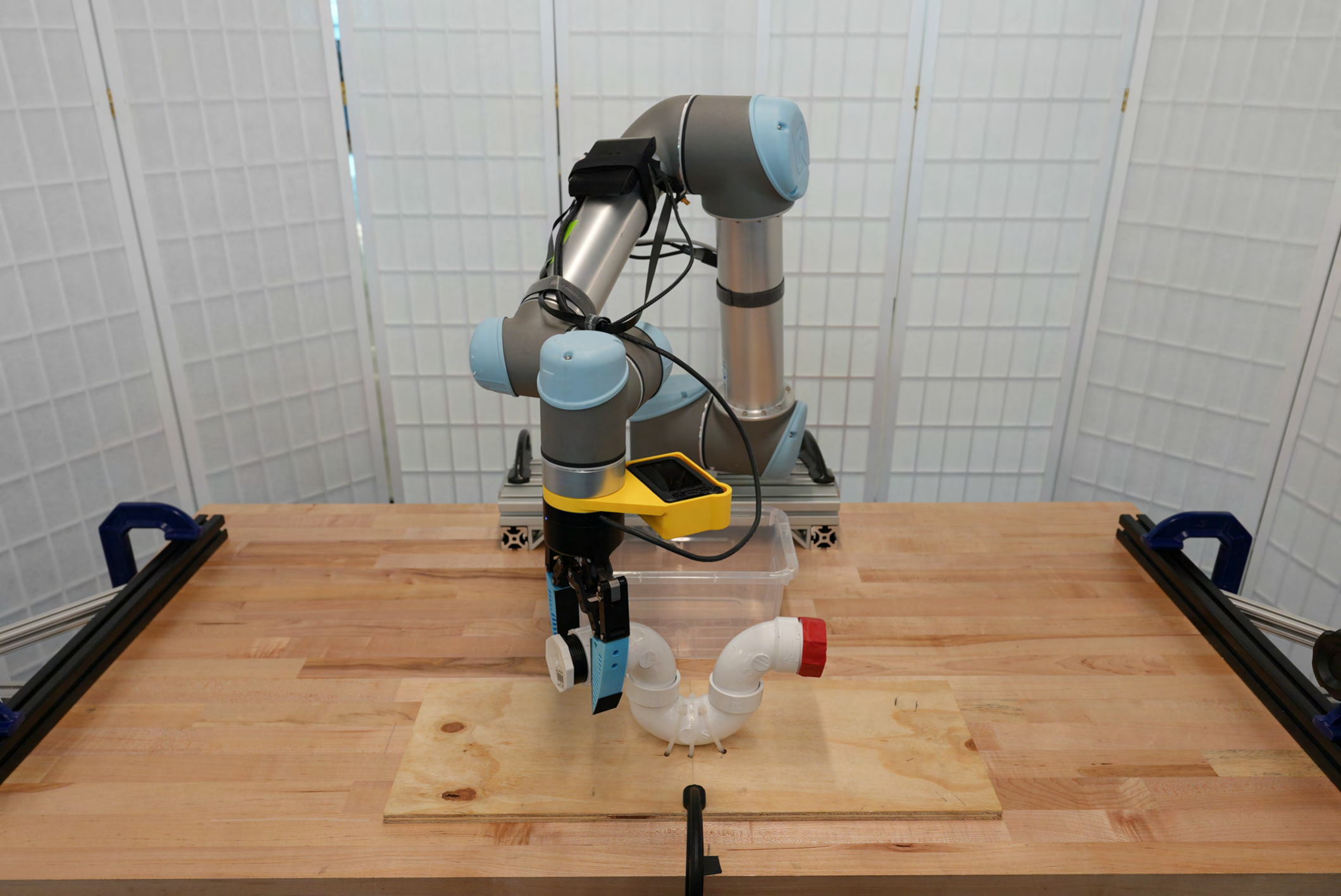}
    \end{subfigure}
    \begin{subfigure}[t]{0.32\linewidth}
        \centering
        \includegraphics[width=1\linewidth]{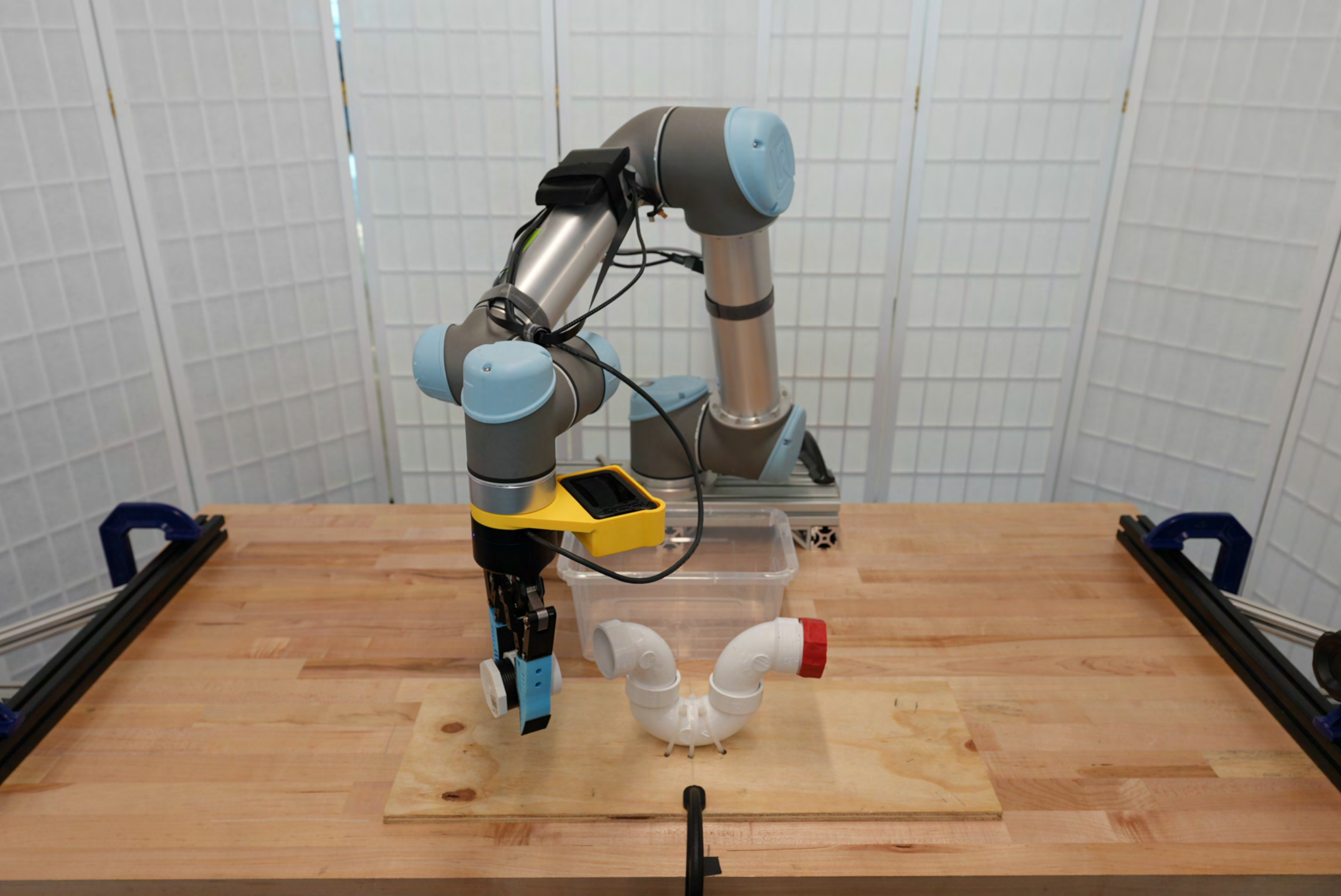}
    \end{subfigure}
    \begin{subfigure}[t]{0.32\linewidth}
        \centering
        \includegraphics[width=1\linewidth]{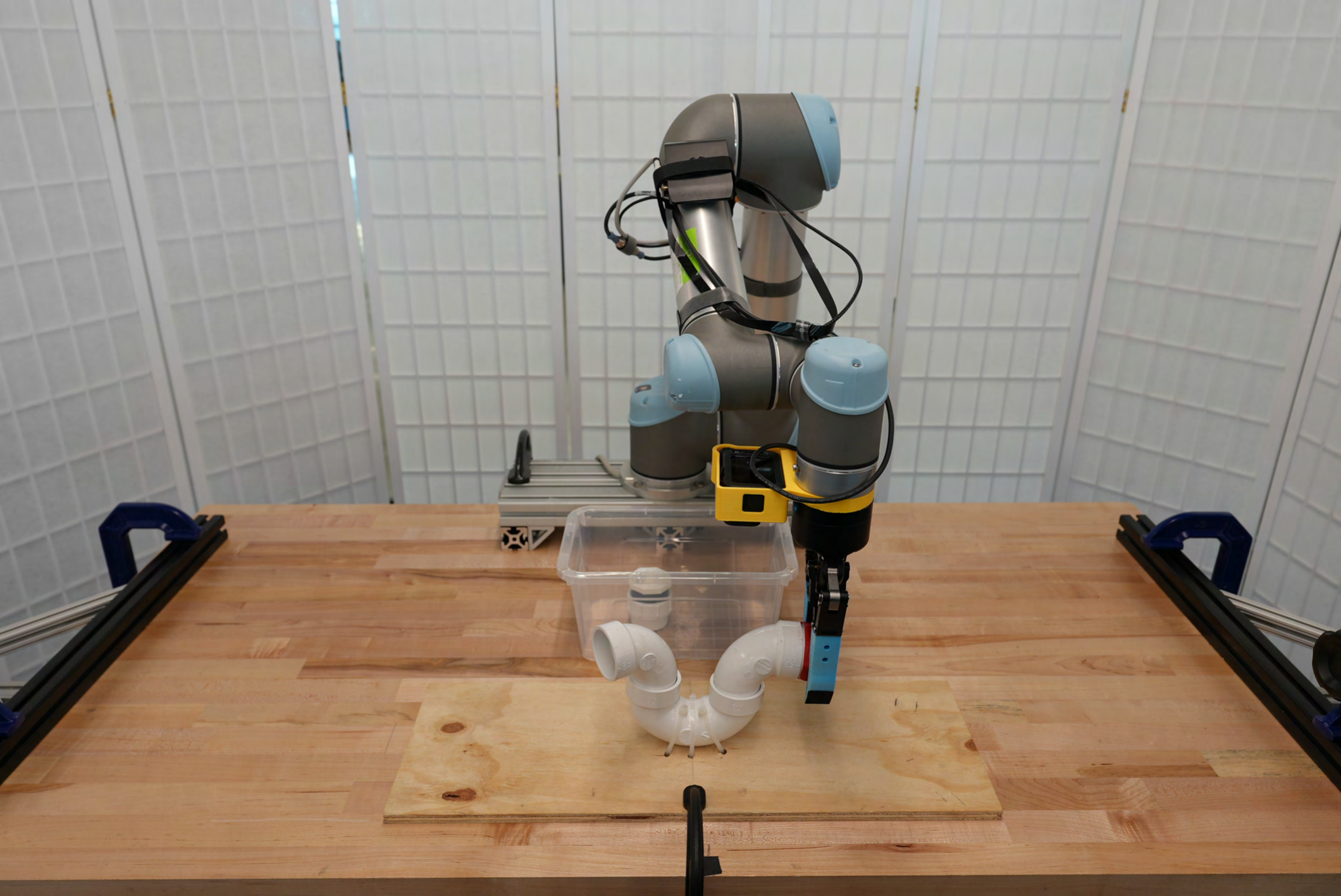}
    \end{subfigure}
    \begin{subfigure}[t]{0.32\linewidth}
        \centering
        \includegraphics[width=1\linewidth]{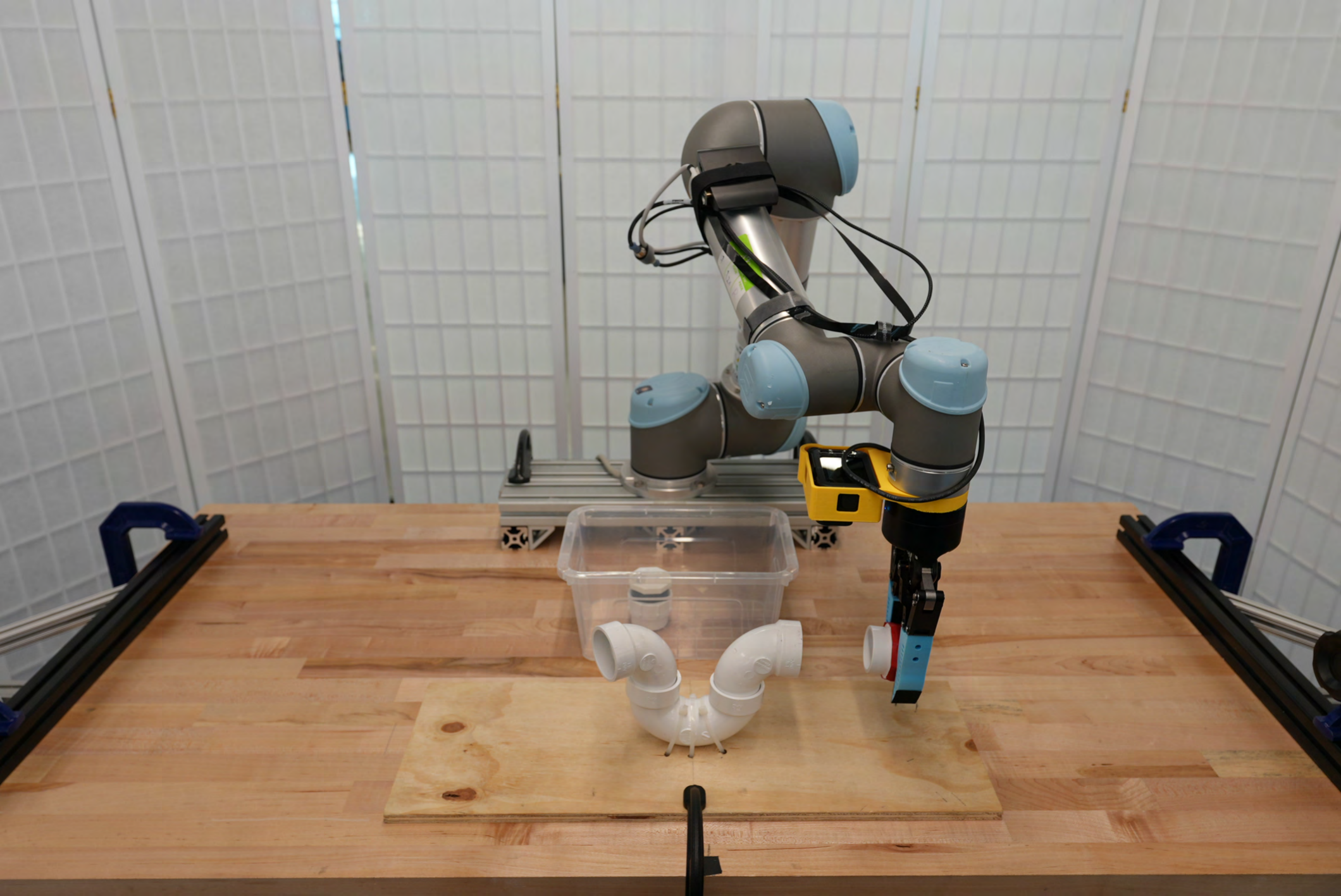}
    \end{subfigure}
    \begin{subfigure}[t]{0.32\linewidth}
        \centering
        \includegraphics[width=1\linewidth]{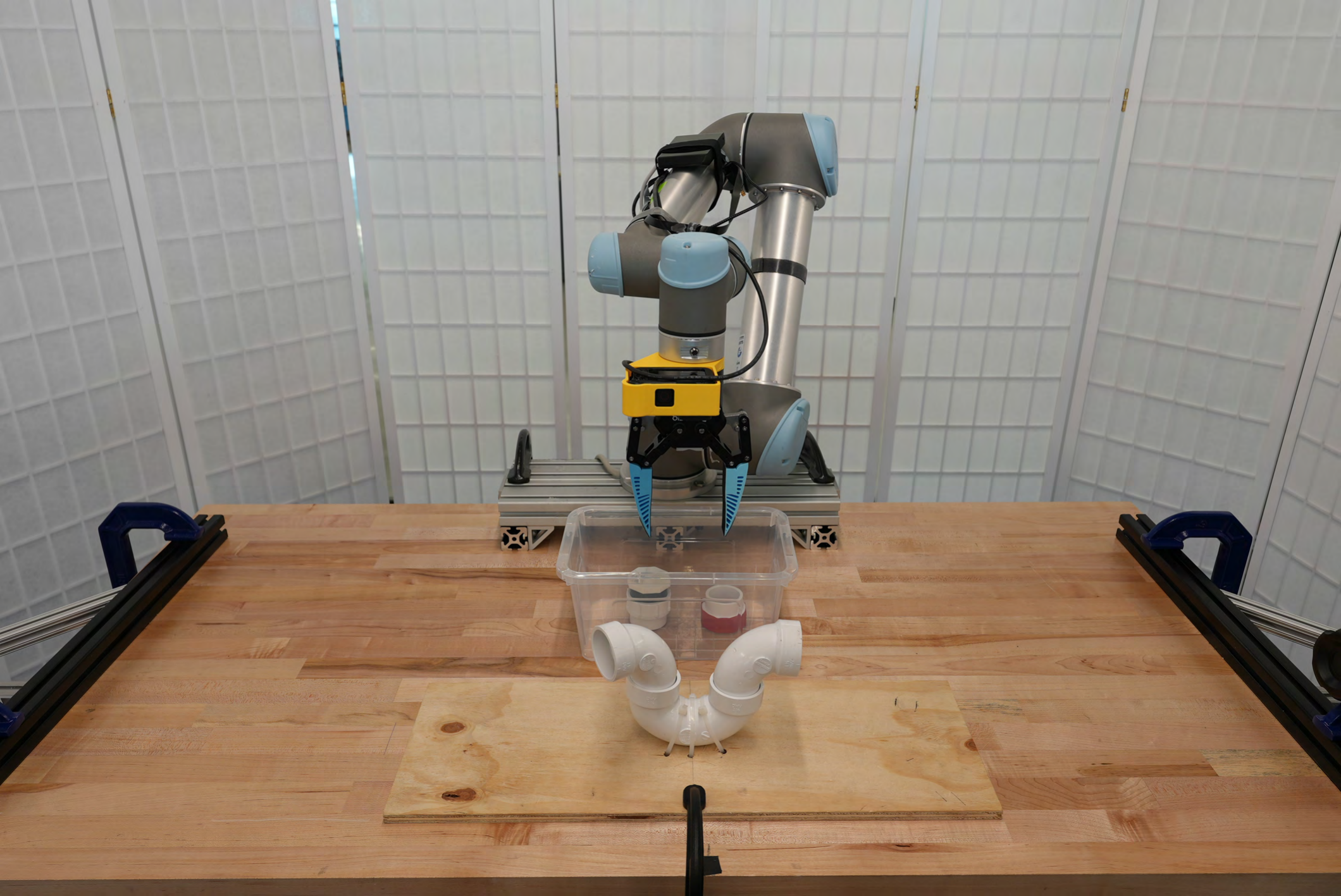}
    \end{subfigure}
    \caption{U-pipe Disassembly}
    \label{fig:u_pipe_rollout}
    \end{subfigure}
\end{figure*}

\begin{figure*}[!h]
\centering
\ContinuedFloat
    \begin{subfigure}[t]{1\textwidth}
    \centering
    \begin{subfigure}[t]{0.32\linewidth}
        \centering
        \includegraphics[width=1\linewidth]{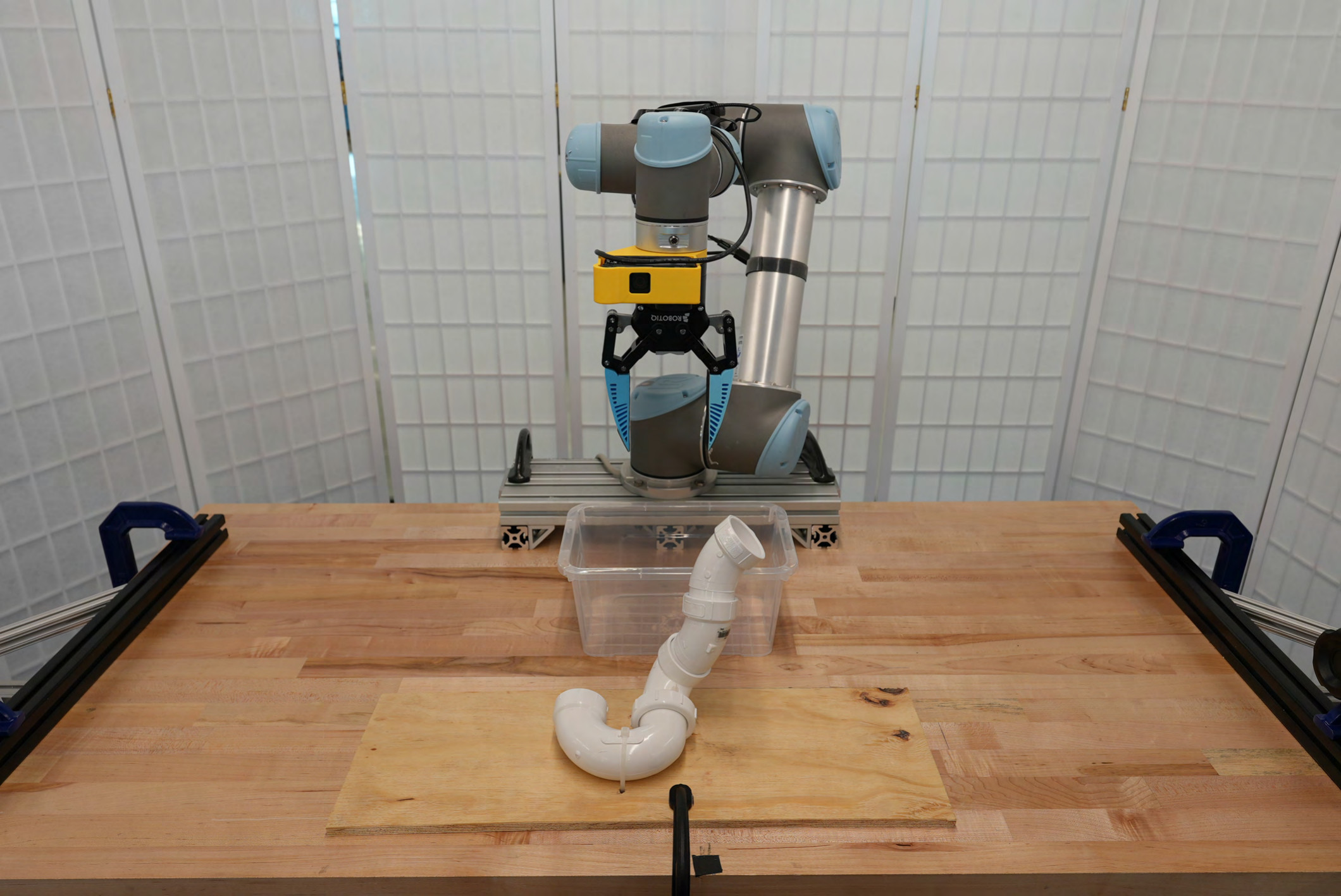}
    \end{subfigure}
    \begin{subfigure}[t]{0.32\linewidth}
        \centering
        \includegraphics[width=1\linewidth]{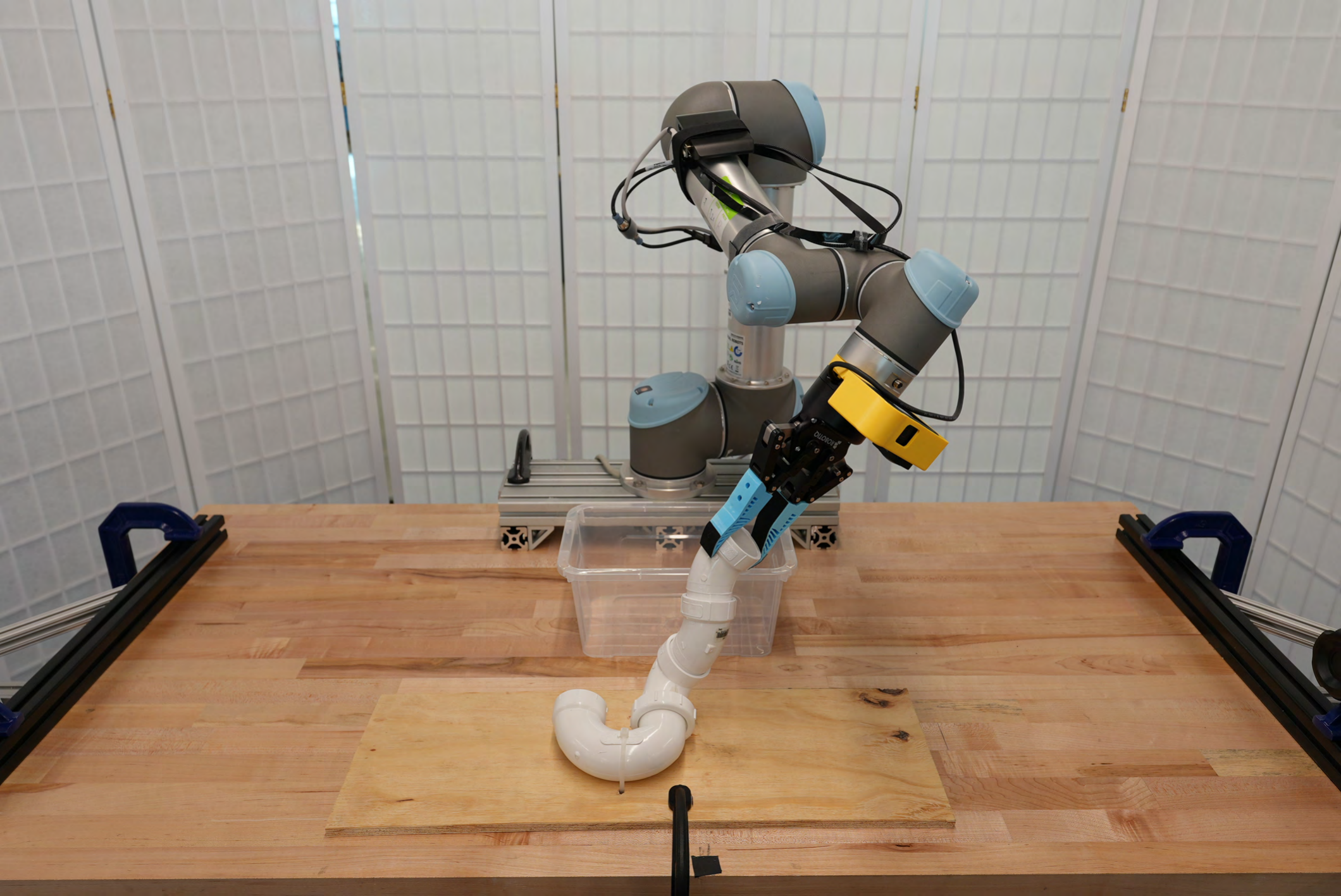}
    \end{subfigure}
    \begin{subfigure}[t]{0.32\linewidth}
        \centering
        \includegraphics[width=1\linewidth]{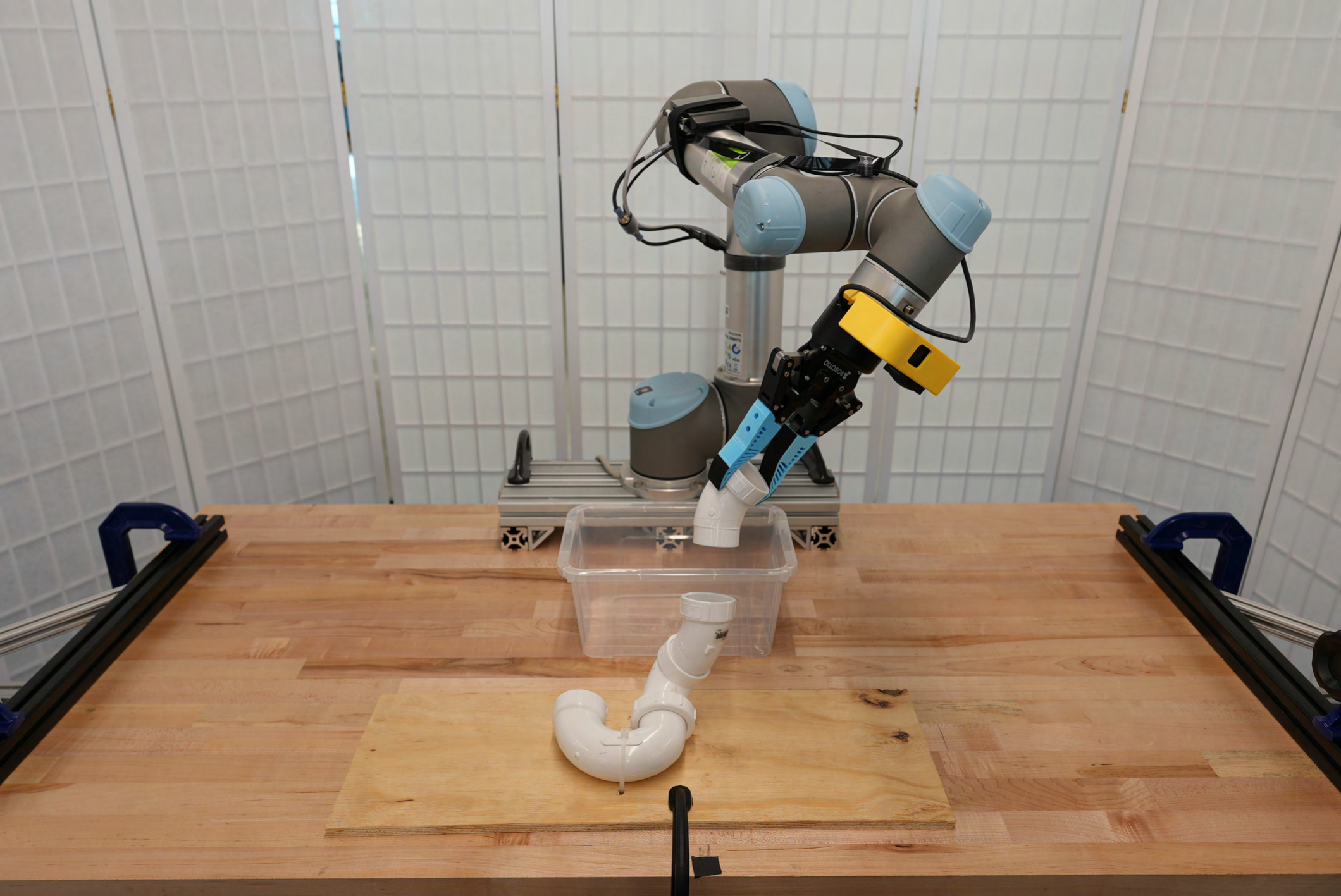}
    \end{subfigure}
    \begin{subfigure}[t]{0.32\linewidth}
        \centering
        \includegraphics[width=1\linewidth]{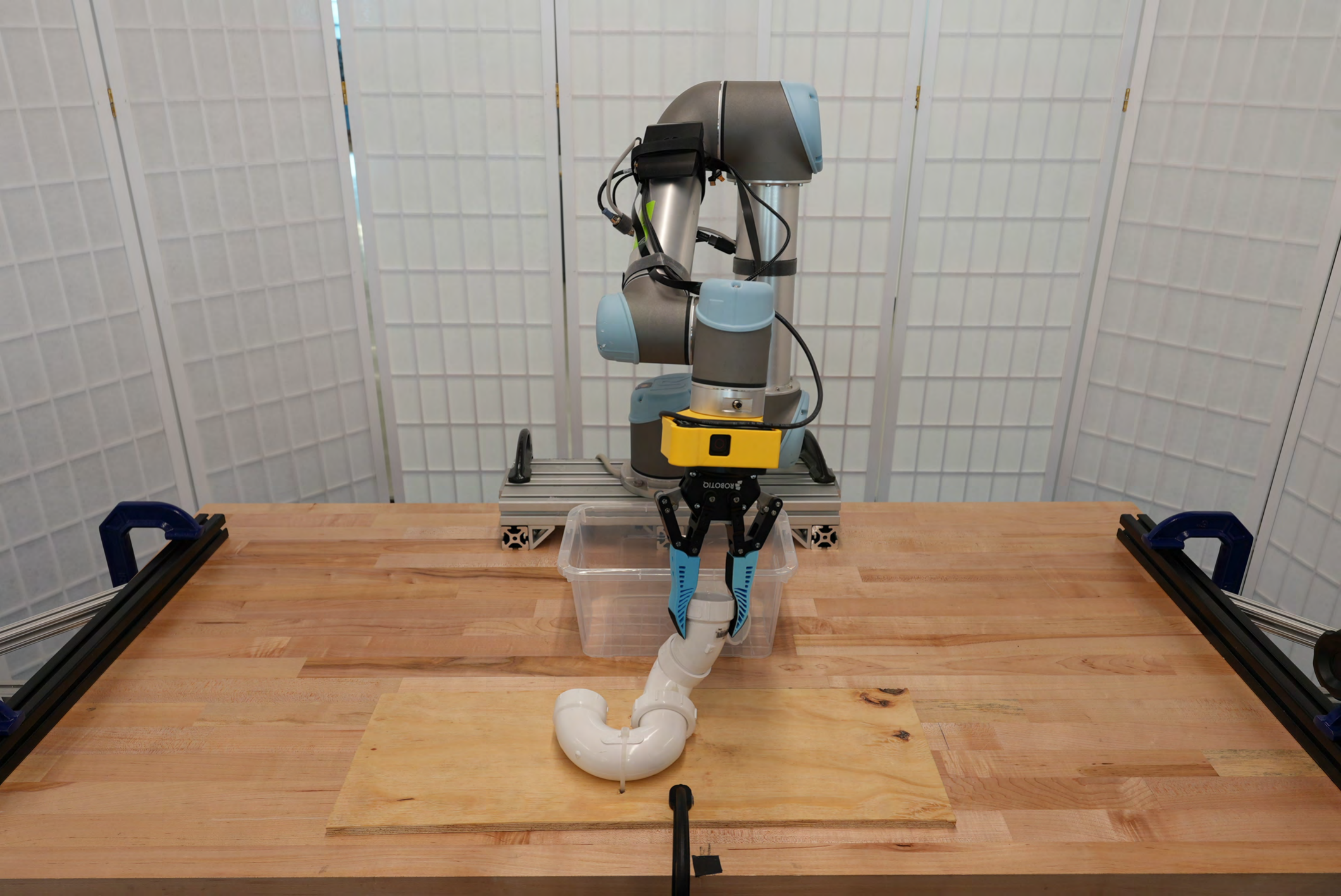}
    \end{subfigure}
    \begin{subfigure}[t]{0.32\linewidth}
        \centering
        \includegraphics[width=1\linewidth]{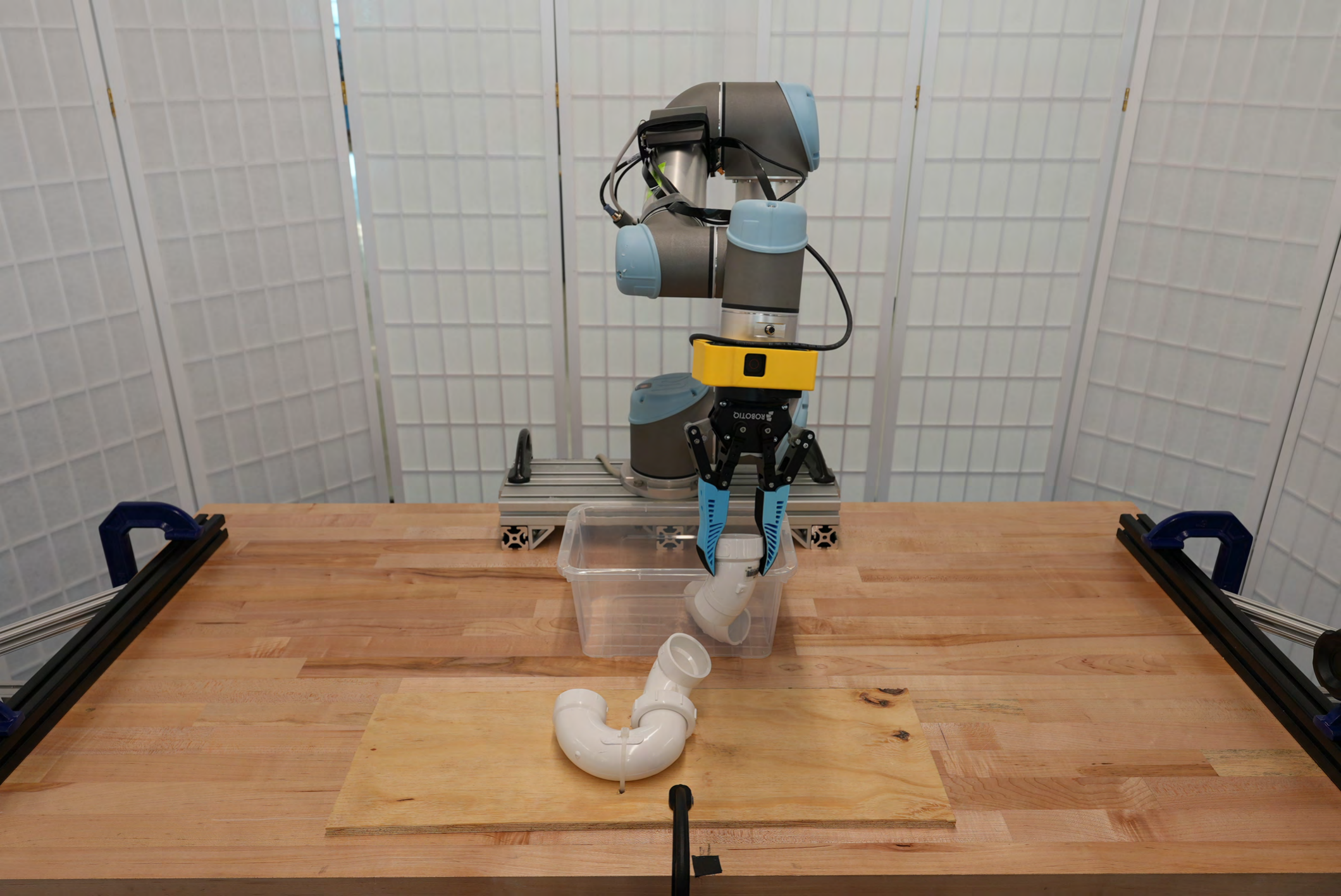}
    \end{subfigure}
    \begin{subfigure}[t]{0.32\linewidth}
        \centering
        \includegraphics[width=1\linewidth]{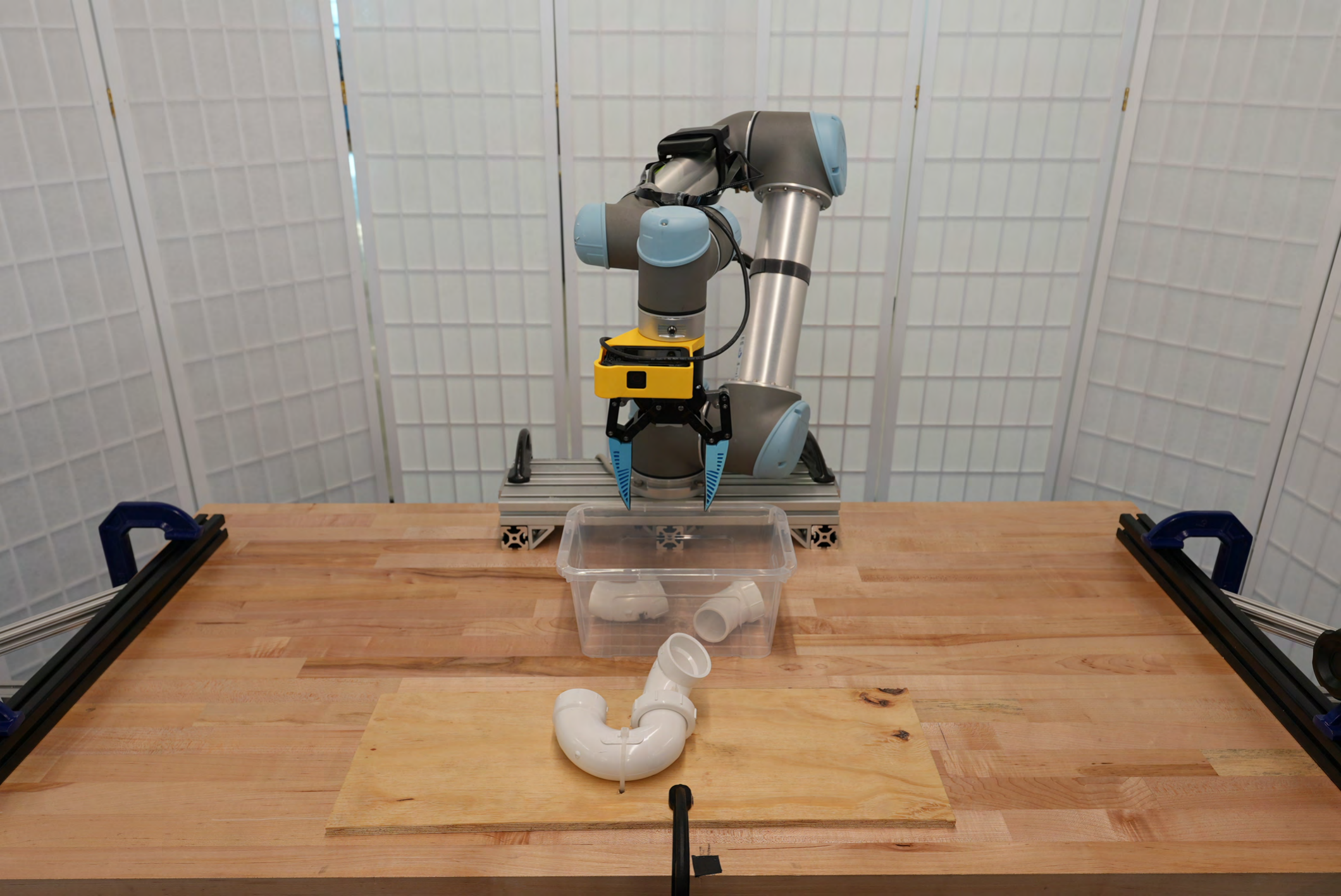}
    \end{subfigure}
    \caption{3D-Pipe Disassembly}
    \label{fig:3d_pipe_rollout}
    \end{subfigure}

    \begin{subfigure}[t]{1\textwidth}
    \centering
    \begin{subfigure}[t]{0.32\linewidth}
        \centering
        \includegraphics[width=1\linewidth]{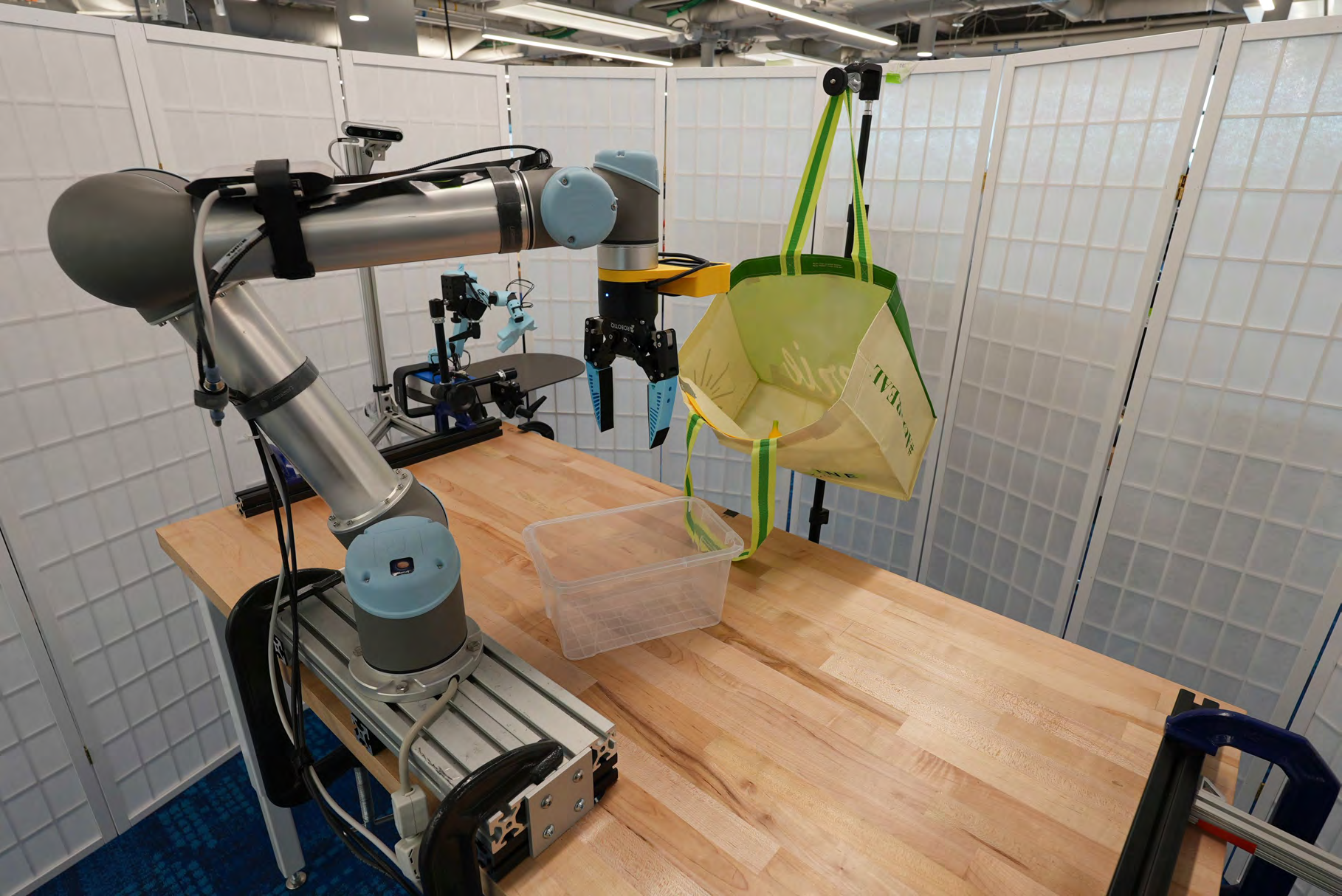}
    \end{subfigure}
    \begin{subfigure}[t]{0.32\linewidth}
        \centering
        \includegraphics[width=1\linewidth]{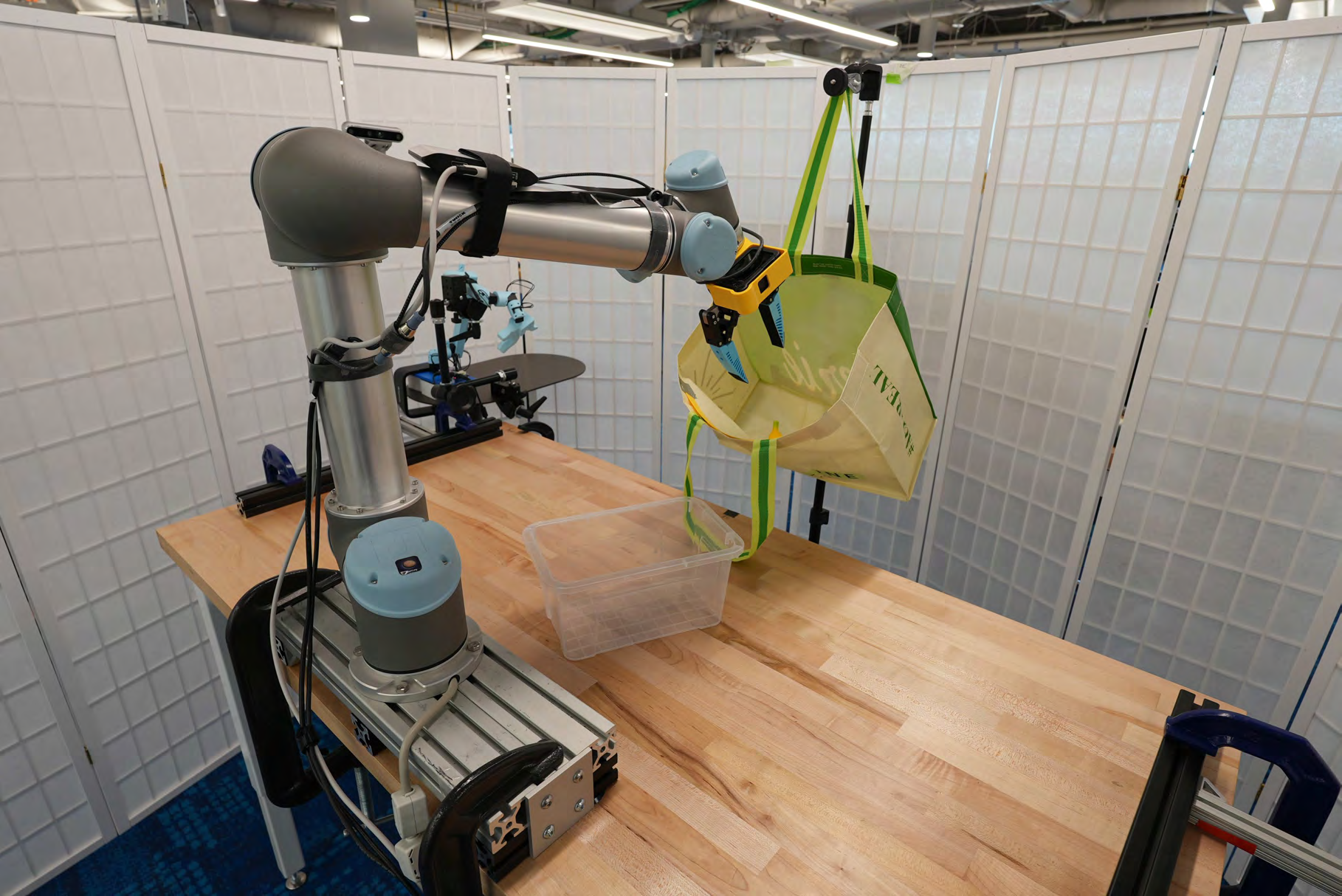}
    \end{subfigure}
    \begin{subfigure}[t]{0.32\linewidth}
        \centering
        \includegraphics[width=1\linewidth]{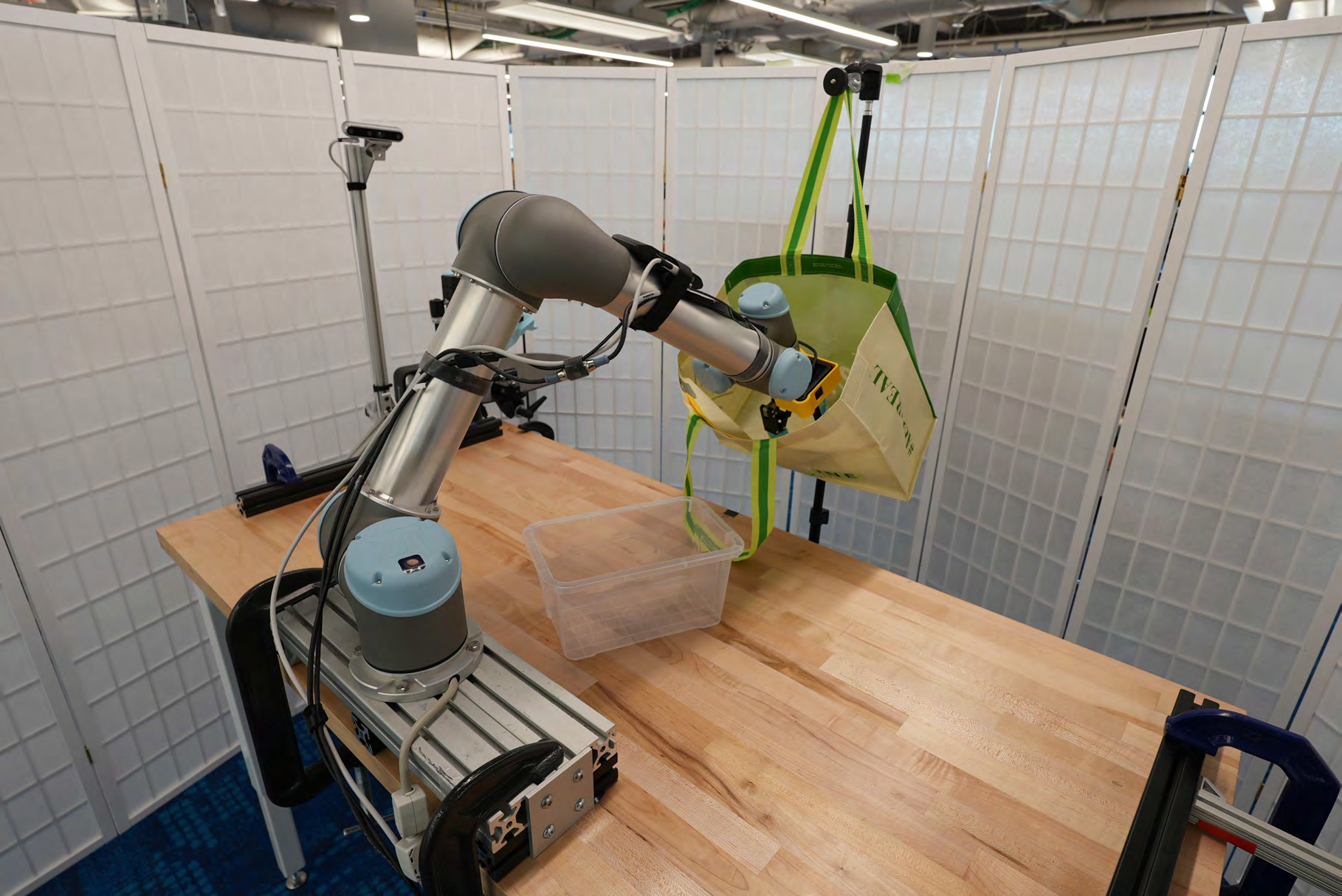}
    \end{subfigure}
    \begin{subfigure}[t]{0.32\linewidth}
        \centering
        \includegraphics[width=1\linewidth]{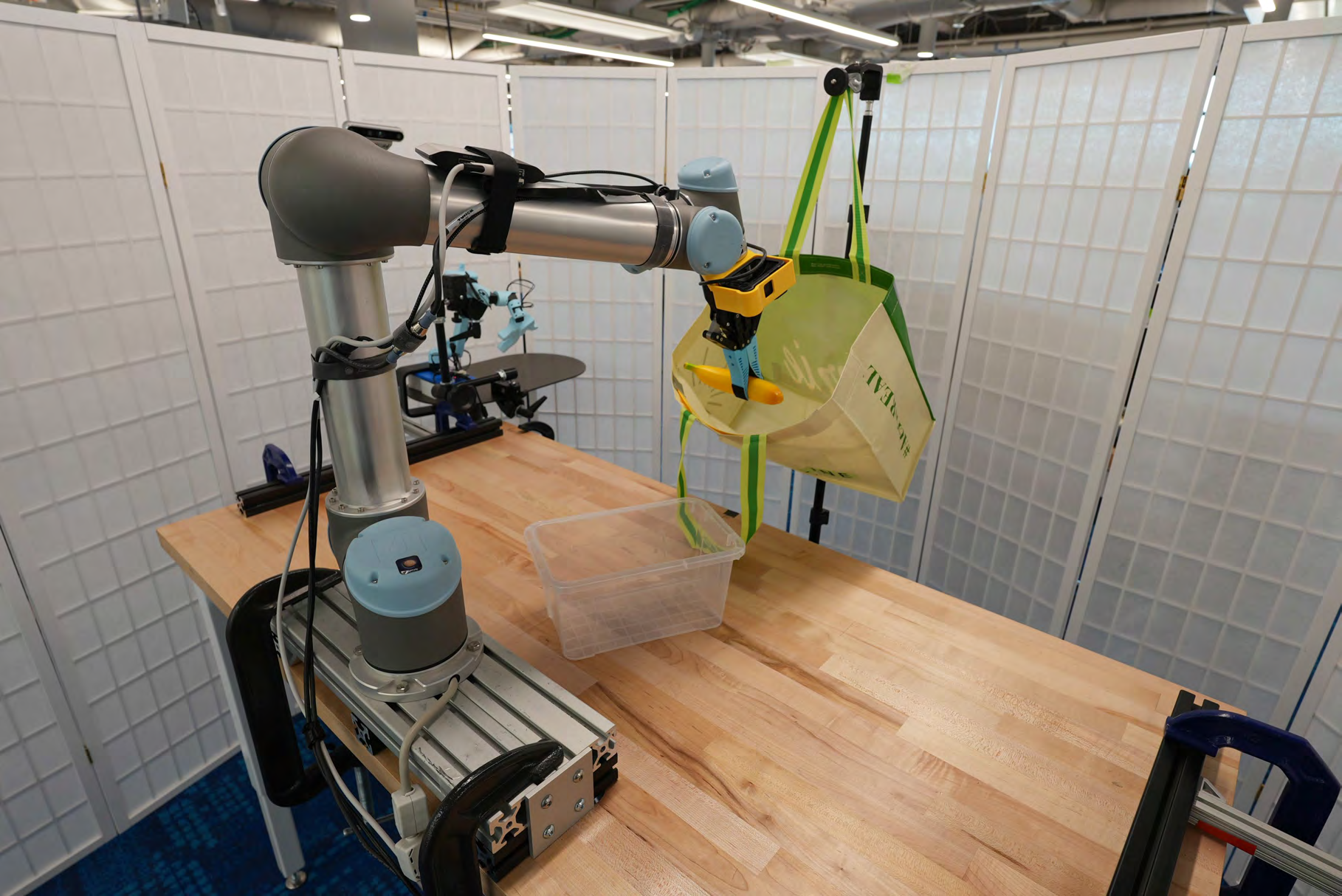}
    \end{subfigure}
    \begin{subfigure}[t]{0.32\linewidth}
        \centering
        \includegraphics[width=1\linewidth]{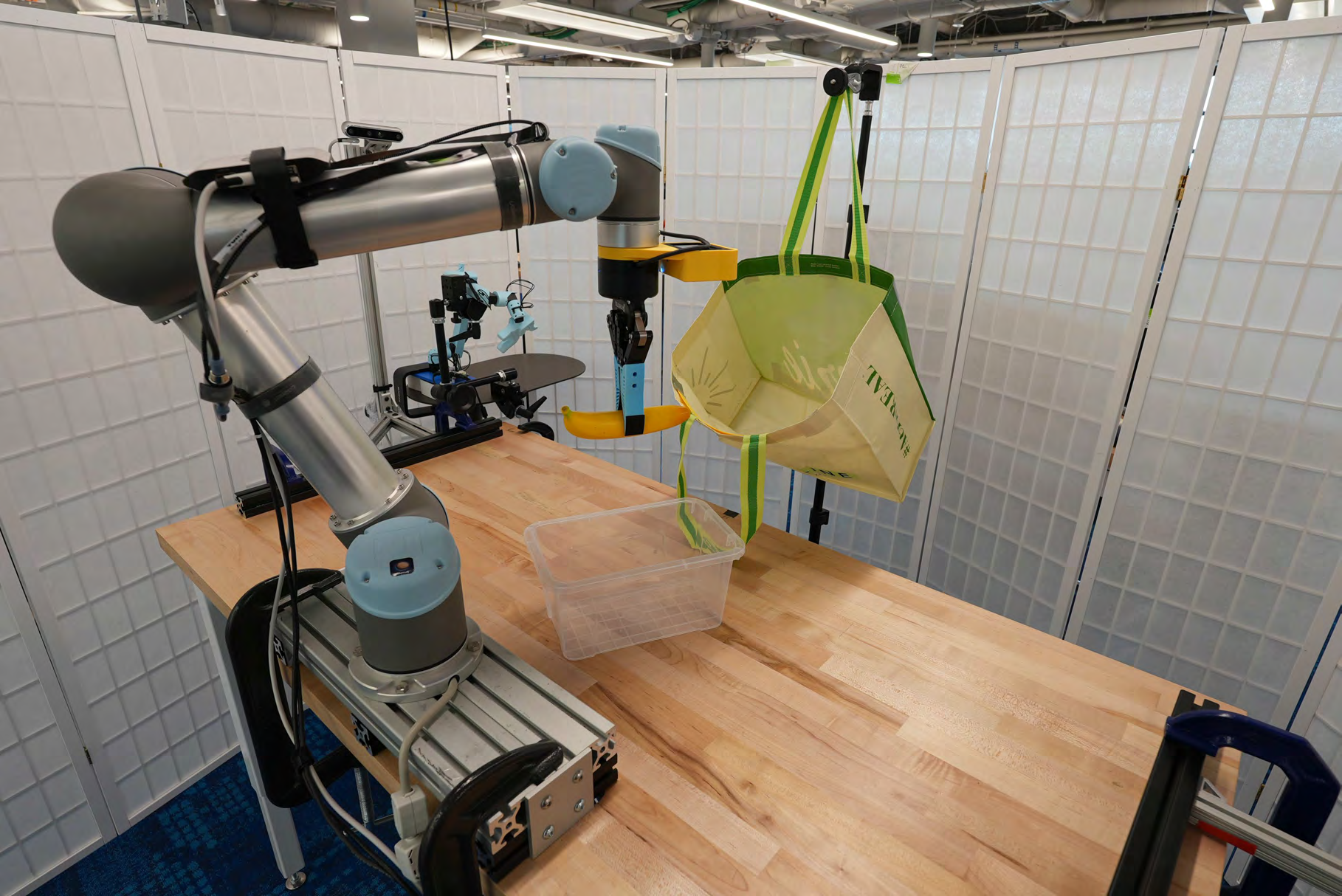}
    \end{subfigure}
    \begin{subfigure}[t]{0.32\linewidth}
        \centering
        \includegraphics[width=1\linewidth]{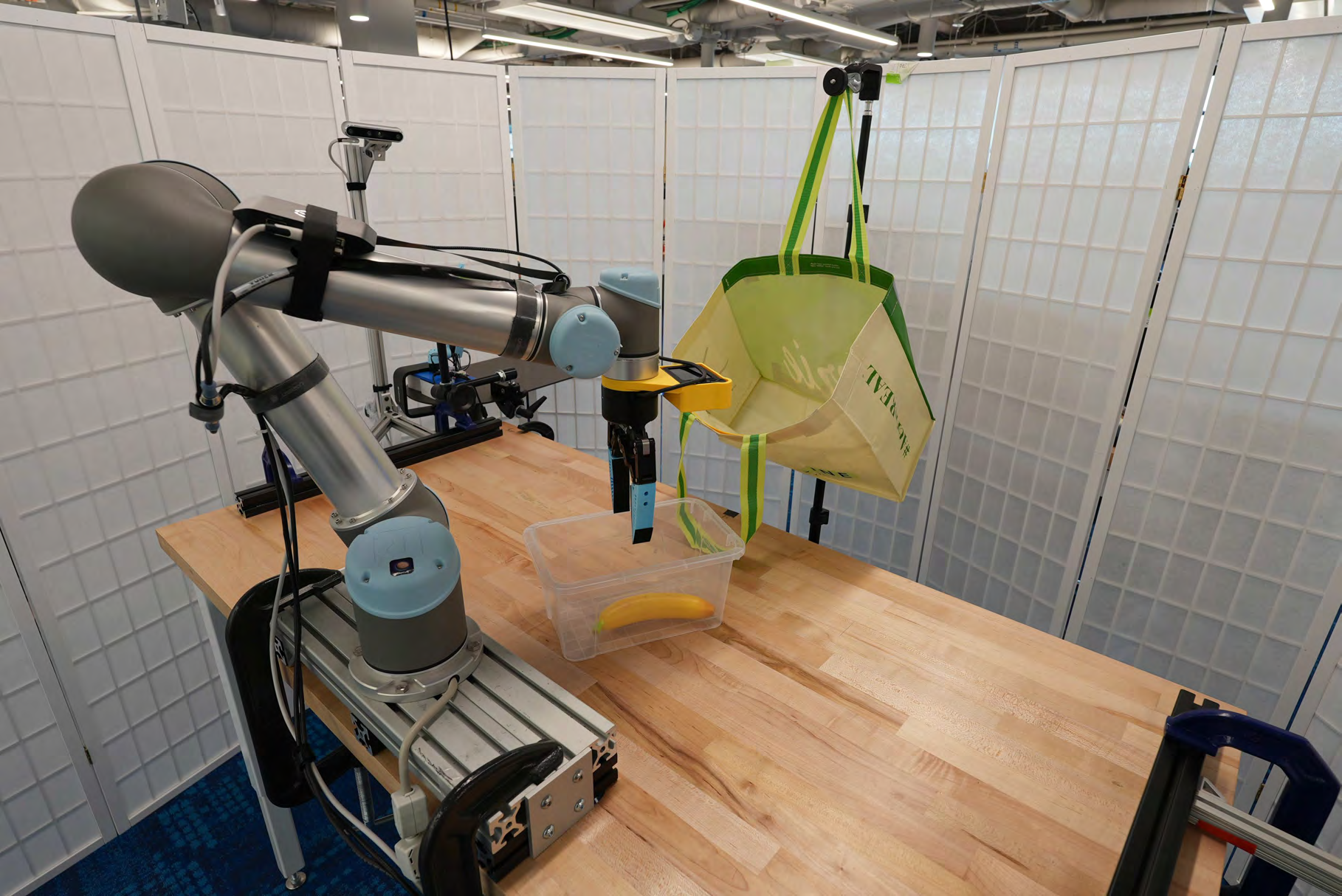}
    \end{subfigure}
    \caption{Grocery Bag Retrieval}
    \label{fig:GB_rollout}
    \end{subfigure}
\caption{Visualization of one episode for each task. Each subfigure illustrates a key action step in the trajectory.}
\label{fig:all_rollout}
\end{figure*}

\end{document}